\pdfoutput=1  
\documentclass[default]{sn-jnl}

\usepackage{graphicx}%
\usepackage{multirow}%
\usepackage{placeins}
\usepackage{amsmath,amssymb,amsfonts}%
\usepackage{amsthm}%
\usepackage{mathrsfs}%
\usepackage[title]{appendix}%
\usepackage{xcolor}%
\usepackage{textcomp}%
\usepackage{manyfoot}%
\usepackage{booktabs}%
\usepackage{algorithm}%
\usepackage{algorithmicx}%
\usepackage{algpseudocode}%
\usepackage{listings}%
\usepackage{adjustbox}%
\usepackage{cleveref}%
\usepackage{tabularx}%
\usepackage{tikz}
\usepackage{comment}
\usepackage{subcaption} 
\usepackage{caption}
\usepackage[shortlabels]{enumitem}
\usepackage{fancyhdr}
\DeclareCaptionFont{custom}{\fontsize{13.5}{14}\selectfont}

\usepackage{array}
\usepackage{arydshln}
\definecolor{olive}{rgb}{0,0.6,0}

\theoremstyle{plain}

\theoremstyle{remark}

\theoremstyle{definition}

\definecolor{cOurs}   {HTML}{006400}  
\definecolor{cZeroth} {HTML}{89CFF0}  
\definecolor{cFirst}  {HTML}{8C564B}  
\definecolor{cTPWL}   {HTML}{17BECF}  
\definecolor{cKoop}   {HTML}{FF0000}  
\definecolor{cDMDc}   {HTML}{9467BD}  
\definecolor{cMZ}     {HTML}{FF7F0E}  
\definecolor{cRL}     {HTML}{1F77B4}  

 \definecolor{cAssmRL} {HTML}{55A868}  
\definecolor{cRnnRL}  {HTML}{E1B12C}  
\definecolor{cTransRL}{HTML}{D62728}  

\usepackage{babel}
\providecommand{\definitionname}{Definition}
\providecommand{\remarkname}{Remark}
\providecommand{\theoremname}{Theorem}
\usepackage{mathtools}
\DeclareMathAlphabet\mathbfcal{OMS}{cmsy}{b}{n}
\usepackage{bm}
\usepackage{amsthm}

\theoremstyle{thmstyleone}%
\theoremstyle{thmstyletwo}%
\newtheorem{remark}{Remark}%

\theoremstyle{thmstylethree}%
\newcommand{\ja}[1]{{\color{black}{#1}}}

\begin{document}

\title[Article Title]{Data-Driven Precision Control of Soft Helicoid Robots}

\subtitle{{\fontsize{12bp}{15bp}\selectfont (submitted to \textit{npj Robotics})}}  


\author[1]{\fnm{Roshan} \sur{S. Kaundinya}}

\author[2]{\fnm{Mark} \sur{Leone}}

\author[3]{\fnm{Roberto} \sur{Sabella}}

\author[2]{\fnm{Hugo} \sur{Buurmeijer}}

\author[2]{\fnm{Luis} \sur{Pabon}}

\author[2]{\fnm{John Irvin} \sur{Alora}}

\author*[2]{\fnm{Marco} \sur{Pavone}}\email{pavone@stanford.edu}

\author*[1]{\fnm{George} \sur{Haller}}\email{georgehaller@ethz.ch}

\affil[1]{\orgdiv{Institute for Mechanical Systems}, \orgname{ETH Z\"urich}, \orgaddress{\street{Leonhardstrasse 21, 8092}, \city{Z\"urich}, \country{Switzerland}}}

\affil[2]{\orgdiv{Department of Aeronautics and Astronautics}, \orgname{Stanford University}, \orgaddress{\street{496 Lomita Mall}, \city{Stanford}, \postcode{94305}, \state{CA}, \country{USA}}}

\affil[3]{\orgdiv{Department of Industrial Engineering}, \orgname{University of Naples Federico II}, \orgaddress{\street{P.le V. Tecchio, 80, 80125}, \city{Naples}, \country{Italy}}}



\abstract{A purely data-driven method grounded in the theory of adiabatic spectral submanifolds (aSSMs) was recently used to model and control nonlinear soft robotic simulators. Here, we extend the methodology to learn aSSM-reduced models from data collected from hardware soft robotic platforms. Specifically, we learn a five dimensional (5D) aSSM-reduced nonlinear ordinary differential equation (ODE) that accurately models the dynamics of a soft-helicoid trunk robot. We then use this ODE in a model predictive control scheme to demonstrate real-time precision control across various challenging closed-loop trajectory-tracking tasks, ranging from 2D to 3D trajectories with static and dynamic constraints. We find that our methodology enhances accuracy and precision across all control tasks compared to available data-driven linear methods and model-free reinforcement learning. While we focus here on a soft-helicoid trunk robot, our methodology relies solely on hardware data and is therefore applicable to a broader class of soft robotic systems.}




\maketitle
\clearpage
Modeling and control of soft robots has been a grand challenge in the recent decade (see \citet{cescilia_22} and \citet{soft_robot_review}). As the design and fabrication of soft robotic systems have advanced, novel soft robotic architectures with complex geometries have become readily attainable. Once such soft robots are in motion, the dynamics that dictate their evolution are nonlinear. This nonlinear behavior of soft robot dynamics, while challenging to model and control, offers operational advantages in safely navigating sensitive regions routinely encountered during surgical, undersea, and space missions. 

On the hardware side, realizing these objectives requires compliant actuators that grant flexibility and dexterity, and robust controllers that ensure real-time tracking precision. Accurately modeling the nonlinear dynamics of a soft robot is a necessary step to inform both hardware fronts as they converge towards mission-ready soft robots. A classical modeling approach is to derive the governing equations from (typically idealized) physical principles. This leads to partial differential equations, which, for numerical use, are further discretized using finite elements (see \citet{faure2012sofa}), yielding a high-dimensional nonlinear mechanical system. 

Classical control schemes exist for tracking tasks in nonlinear mechanical systems, either by linearizing the system along the target track (see \citet{Sastry1999}) or by deriving analytic closed-loop nonlinear control formulas for the forces (see \citet{Udwadia2003}). While the exactness of such formulas is appealing, the derived forces do not automatically correspond to admissible control inputs for the soft robotic system, which, by design, are underactuated. Also, in this classical approach, one has to make assumptions about the material parameters, the elasticity law, the finite-element geometry, and the geometric nonlinearities arising from boundary conditions. 

In practice, one has to reduce the full dynamics of the soft robot to a low-dimensional model for enabling real-time control. Most such reduction strategies employ POD-Galerkin methods (see \citet{goury18}, \citet{Katzschmann2019} and \citet{pavone211}), which are only fully justified if the system's dynamics are linear. As an alternative, equation-based methods tend to approximate the shape profiles of soft robot geometries using either piecewise constant curvature (see \citet{Katzschmann_cc}) or strain (see \citet{renda18}), or Cosserat-rod models (see \citet{naughton21}). These have low fidelity compared to finite-element methods and may only apply to certain types of soft robotic geometries. As a result, most existing classical approaches are tedious to set up and, in some cases, even impossible to automate for general soft robotic hardware. 

Dynamic data from motion capture systems or lasers is now routinely processed and used in simple feedback control. Hence, a data-driven route to modeling soft robot dynamics has become practically viable. The simplest and most popular data-driven method is to fit a linear, time-invariant system to dynamic data from soft robots. \citet{Proctor2016} first implemented this method, called Dynamic Mode Decomposition with Control (DMDc), and tested it on simple problems. Recent work by \citet{Adibnazari2025} performed a useful case study, comparing various linear data-driven methods, including DMDc, for shape control of a soft robotic fish. 

A linear model yields fast controllers but will perform poorly for inherently nonlinear systems. To address this limitation, \citet{rowley09} sought to fit linear dynamics to nonlinear functions of observables, which has lead to the idea of extended DMD with Controls (or EDMDc; see \citet{bruder21}). Another fundamentally linear idea, Koopman analysis with static pregain, replaces the static input-output relation of a soft robot with a linear function (see \citet{haggerty23}). Comparisons on simulations and hardware consistently show, however, that linear models perform poorly in soft robotics applications involving nontrivial trajectories (see \citet{alora23,alora24} and \citet{Kaundinya2026}, and upcoming experiments in this paper).

A natural next step beyond fitting linear models to data is to fit a neural network that is constrained by the soft robot's physical laws. \citet{Gao2024} learned the forcing due to actuation on the soft robot as a neural network, given the control inputs, assuming that the forcing is differentiable. \citet{bern2020} pursue a similar approach but constrain it to follow soft-robotic kinematics. \citet{Stozle} learn the dynamics as a neural network by shaping potential energy functions they postulate for the soft robotics system, while \citet{valadas_2025} assumes a form for the strain relations. For soft robotics systems with possible memory-dependent behavior, a recent method (see \citet{mori_soft_26}) uses the Koopman-Mori-Zwanzig framework (see \citet{Lin2021} and \citet{Gupta2025}) to account for memory dynamics via memory kernel terms, which are learned by fitting neural networks. Although these methods account for some of the underlying physics, they are black-box approximations and do not directly target the control task of interest. As a consequence, their accuracy is limited, as we will see in examples of the present paper.

Reinforcement learning (RL) directly targets the control task of interest by providing a black-box, model-free methodology that learns a control policy as a neural network optimized for the intended control objectives. RL approaches have found success in rigid robotics for tasks involving repetitive control actions that aid locomotion (see \citet{AnyMal}) and autonomous driving (see \citet{Levine16}). Even in the rigid robotic setting, however, an extremely large dataset spanning the robot's workspace is needed to guarantee a convergent RL control policy. This is a physically demanding challenge for soft robotic systems: it prevents extended on-robot data collection due to excessive actuator wear, and also hysteretic damage to the soft robot geometry. For this reason, most RL methodologies for soft robotics rely on simulators (see \citet{Thuruthel2019ModelBasedRL}, \citet{naughton21} and \citet{pmlr-v229-jitosho23a}) that mimic hardware to generate data for training a reliable RL policy or learn an effective model (see \citet{zheng2025learningsoftroboticdynamics}) that guarantees convergence to a policy by actively exploring the workspace. These model-based RL policies, however, suffer from the same issues as the classical and black-box approaches we have outlined.

A common feature of all soft continuum mechanical systems has emerged in recent years: their dominant dynamics converge to a low-dimensional, attracting invariant manifold in the phase space. Specifically, invariant manifolds known as spectral submanifolds (SSMs) have established themselves as powerful tools for mathematically justified model reduction (see \citet{haller25}). They have been successfully used in structural dynamics and fluid mechanics problems to predict forced response under previously unseen external forcing (see \citet{cenedese22a}, \citet{morsy2024reducingfiniteelementmodels}, and \citet{Bettini2026}).  For SSM construction in soft robotic systems, an additional timescale dichotomy is relevant: the internal vibration rates of the robot decay much faster than its operational speed along tracks in realistic applications. In that case, the invariant structure of interest that contains the global dominant dynamics is an adiabatic spectral submanifold (aSSM) (see \citet{haller24_wa} and \citet{Kaundinya2026}). An aSSM is an attracting invariant manifold composed of trajectories that approach the set of possible operational trajectories of the robot at the slowest rate. 

Reducing the full dynamics onto aSSMs offers a rigorous approach to modeling the originally infinite-dimensional soft robot dynamics via a low-dimensional ODE. As a first step in this direction, \citet{alora24} used a zeroth-order (i.e., near-equilibrium) approximation of aSSMs to model and control a hardware diamond robot on 2D target tasks. \citet{Kaundinya2026} extended this by identifying low-dimensional aSSMs for several soft-trunk simulators and demonstrating accurate control across diverse 3D target tasks. They also performed a systematic comparison of various approximations of aSSMs. \citet{Bettini2026_muscles} offered a further reduction of the aSSM onto a slow manifold to enable modeling and control of one-dimensional artificial muscles. While these have been promising applications of the aSSM-based soft robot control, a hardware demonstration of the aSSM methodology in fully 3D soft robot control has not been available.

We focus here on addressing this hardware gap in the current aSSM control literature. We consider a soft robot made from the sustainable elastomeric resin EPU 46 (see \citet{carbonEPU46}), with a 3D geometry resembling an elephant trunk. Specifically, the soft trunk robot's architecture is composed of trimmed helicoids, inspired by the recent work of \citet{guan2023trimmed}. These architected robots are designed to replicate the compliance of the biological morphologies that inspire them. Controlling soft-helicoid trunk robots has recently attracted significant interest. For instance \citet{guan2023trimmed} use inverse kinematics for closed-loop control under the assumption of piecewise-constant curvature. They propose a strategy to infer the correct tendon lengths that correspond to actuation forces. Related recent work by \citet{zhang2026printedhelicoidsembeddedair} provides design guidelines for the 3D printing of a trimmed helicoid, with results from open-loop control.   Also, \citet{wong2026closedformclfcbfcontrollerwholebody} uses control barrier functions for whole-body avoidance; their strategy relies on a form of the equations of motion for their helicoidal trunk setup. \citet{nan2025efficient} perform model-based RL policy learning for a soft-trunk helicoid using a neural network as a dynamics model. They then perform closed-loop tracking of 2D targets over short tracks compared to the length of the trunk robot. 

Despite these advances, a data-driven closed-loop control method has not been put forward for precision control of soft helicoid trunk robots for dynamic target tasks spanning a global workspace that can also simultaneously perform obstacle avoidance. 

\textbf{Contribution:} Our main contribution here is a fully data-driven, hardware implementation of aSSM-based model-predictive control (MPC). Specifically:
\begin{itemize}
    \item \textbf{[Systematic aSSM model learning from data]} Section~\ref{sec:methods} presents a systematic methodology for learning accurate, low-dimensional aSSM-reduced models from limited observable data. We demonstrate this methodology with a 5D aSSM-reduced model of our soft helicoid trunk robot.
    \item \textbf{[Practical use of aSSM in MPC]} We integrate the aSSM-reduced model into an MPC scheme that exploits the full aSSM geometry. In contrast to prior work by \citet{Kaundinya2026}, the present aSSM-reduced model does not explicitly split the control input into slow and fast components and also does not rely on an inverse-kinematics map to generate guesses for solving the MPC. As a consequence, the model can be readily used in standard control optimization algorithms.
    \item \textbf{[Closed-loop benchmarking]} Sections~\ref{subsec:closed-loop}--\ref{subsec:constraints} report closed-loop trajectory tracking results for aSSM-based MPC, benchmarked against state-of-the-art linear baselines: DMDc, the Koopman-pregain method, a memory-based Koopman--Mori-Zwanzig model, and model-free reinforcement learning via proximal policy optimization (PPO).
    \item \textbf{[Hardware validation]} We provide hardware demonstrations, with video, for all closed-loop control tasks, including obstacle avoidance along 3D tracks.
\end{itemize}
Throughout this work, we assume no information about the equations of motion in any form. Accordingly, we only choose baselines in our comparison that are agnostic to the physics of the robot.

\section{Results}\label{sec:results}
In this section, we describe the closed-loop performance of our data-driven methodology based on the theory of aSSMs on a soft helicoid trunk robot. We first briefly discuss the design, digital fabrication, and morphology of the robot. We then present our data-driven modeling approach to capture the dynamics of the robot and evaluate the model's performance on previously unseen hardware data. We next present real-time, closed-loop results for target tracking across three tracks, two of them 3D, with varying sizes, complexities, and speeds. We close this section by presenting real-time, constraint-aware performance of the soft helicoid robot for 3D static and dynamic obstacles in its workspace.

All of our closed-loop results for 3D target tracking have a video demonstration of the soft helicoid robot and also include comparisons with available linear data-driven modeling baselines: Koopman-pregain (\citet{haggerty23}), and Dynamic Mode Decomposition with Control (DMDc) (\citet{Proctor2016}), a memory-based baseline Mori-Zwanzig-Koopman (MZ-Koopman) (\citet{mori_soft_26}), and a model-free online reinforcement learning (RL) baseline trained using the proximal policy optimization (PPO) approach (\citet{schulman2017proximalpolicyoptimizationalgorithms}). We summarize the model specifications, closed-loop tracking performance, and controller frequency in Table~\ref {tab:baseline_comparison}. We refer to the Supplementary Section \ref{app:baselines} for details on learning the baselines from data. 

\begin{table}[t]
\centering
\caption{Comparison of data-driven aSSM-based modeling and control against linear and learning-based baselines for closed-loop 3D target tracking. The maximum error represents the maximum relative pointwise error obtained across all closed-loop tasks presented in Section~\ref{subsec:closed-loop}.}
\label{tab:baseline_comparison}
\begin{tabular}{lcccccc}
\toprule
Method & Model & Data & Training & Maximum & Constraints & Controller \\
       & dim.\ $d$ & collection time & time & error [cm] & handling & freq.\ [Hz] \\
\midrule
aSSM (ours)     & 5   & 2 hours   & 5 minutes   & 5   & Yes & 50 \\
Koopman-pregain \citep{haggerty23} & 6   & 2 hours  & 30 seconds   & 38   & No   & 100 \\
DMDc \citep{Proctor2016}            & 6   & 2 hours   & 30 seconds   & 18  & Yes   & 100 \\
MZ-Koopman \citep{mori_soft_26}     & 33   & 2 hours   & 15 minutes   & 17   & Yes   & 20 \\
RL (PPO) \citep{schulman2017proximalpolicyoptimizationalgorithms} & NA   & 6 hours   & 6 hours   & 10   & No   & 100 \\
\bottomrule
\end{tabular}
\end{table}



\subsection{Hardware: Design and setup}\label{subsec:design}
We consider a bio-inspired, tendon-driven, soft robotic platform, modeled after an elephant trunk, to validate model-based and model-free control methods. Our design targets two objectives: (i) maximizing the reachable workspace via large bending angles while minimizing parasitic axial compression and viscoelastic hysteresis, and (ii) preserving a wide, unobstructed central lumen to reduce material use and provide protected internal routing for power/signal lines, enabling future end-effector tooling (e.g., a camera or gripper).

\subsubsection{Morphology and fabrication}
Homogeneous hollow-cylinder morphologies are prone to radial expansion and buckling under tendon loads, which severs the monotonic tendon-displacement-to-curvature mapping and precludes reliable control (morphology comparison in Supplementary Section~1). We instead adopt the Trimmed Helicoid (TH) architecture~\citep{guan2023trimmed}, which decouples axial ($K_{axial}$) and bending ($K_{bend}$) stiffness, and can be tuned to obtain a structure that is compliant in bending yet stiffer in the axial direction. This enables a larger workspace over the whole range of tendon displacement.

The platform is fabricated via Digital Light Synthesis (DLS\texttrademark) and Continuous Liquid Interface Production (CLIP\texttrademark) in EPU~46 (Elastomeric Polyurethane, 40\% bio-based, see \citet{carbonEPU46}), yielding monolithic, isotropic parts with injection-molding-like quality and negligible hysteresis ($<1.5\%$). This manufacturing process was chosen over more common ones (e.g. FDM), which introduce defects unsuitable for cyclic tendon loading (details in Supplementary Section \ref{subsec:design_supp}).

\subsubsection{Optimization and final platform}
As no prior characterization exists for this TH--DLS--EPU~46 hardware combination, we used an agile prototyping approach, empirically iterating over sixteen morphological variants to converge on parameters $P1$--$P4$ which we highlight in the Table.~\ref{fig:design_full}a (see Supplementary Section~1 for more detail). The optimized single-module geometry (Fig.~\ref{fig:design_full}b) achieves bending angles $>90^{\circ}$ with axial compression $<15\%$. Our use of microstructure parameters $P1$--$P4$ for optimization is informed by \citet{sabella2023indices}, which reduces cost, mass, and material waste, thereby supporting environmental sustainability.

The final platform comprises three identical modules in series with a total length of $42 \text{ [cm]}$, joined by rigid interface disks that also mount retro-reflective markers (see Fig.~\ref{fig:design_full}c). The actuation system comprises of 12 tendons driven by 6 Dynamixel XM540-W150-R motors (2 per module, each controlling an antagonistic tendon pair for single-plane bending), with 12 pretensioning mechanisms to remove slack and deadzones. Module tips are tracked in real-time using 4 OptiTrack Flex~3 cameras (see Fig.~\ref{fig:design_full}c).

\begin{figure}[t!]
    \centering
    \begin{minipage}[t]{0.58\linewidth}%
        \vspace{0pt}
        \centering
        \begin{subfigure}[t]{\linewidth}
            \centering
            \small
            \begin{tabular}{llc}
                \toprule
                \textbf{ID} & \textbf{Parameter} & \textbf{Value} \\
                \midrule
                -- & Outer Radius, $R_{ext}$ & 20.0 mm \\
                -- & Height, $H$ & 135.0 mm \\
                -- & Number of helices, $N_{hel}$ & 6 \\
                \textbf{P1} & Helix thickness, $T$ & 2.2 mm \\
                \textbf{P2} & Width over radius, $W/R_{ext}$ & 0.35 \\
                \textbf{P3} & Number of revolutions, $N_{rev}$ & 1.5 \\
                \textbf{P4} & Microstructures, $V$ & Absent \\
                \bottomrule
            \end{tabular}
            \caption{}
            \label{tab:optimized_parameters}
        \end{subfigure}

        \vspace{1.0em}

        \setcounter{subfigure}{2}
        \begin{subfigure}[t]{\linewidth}
            \centering
            \includegraphics[width=\linewidth]{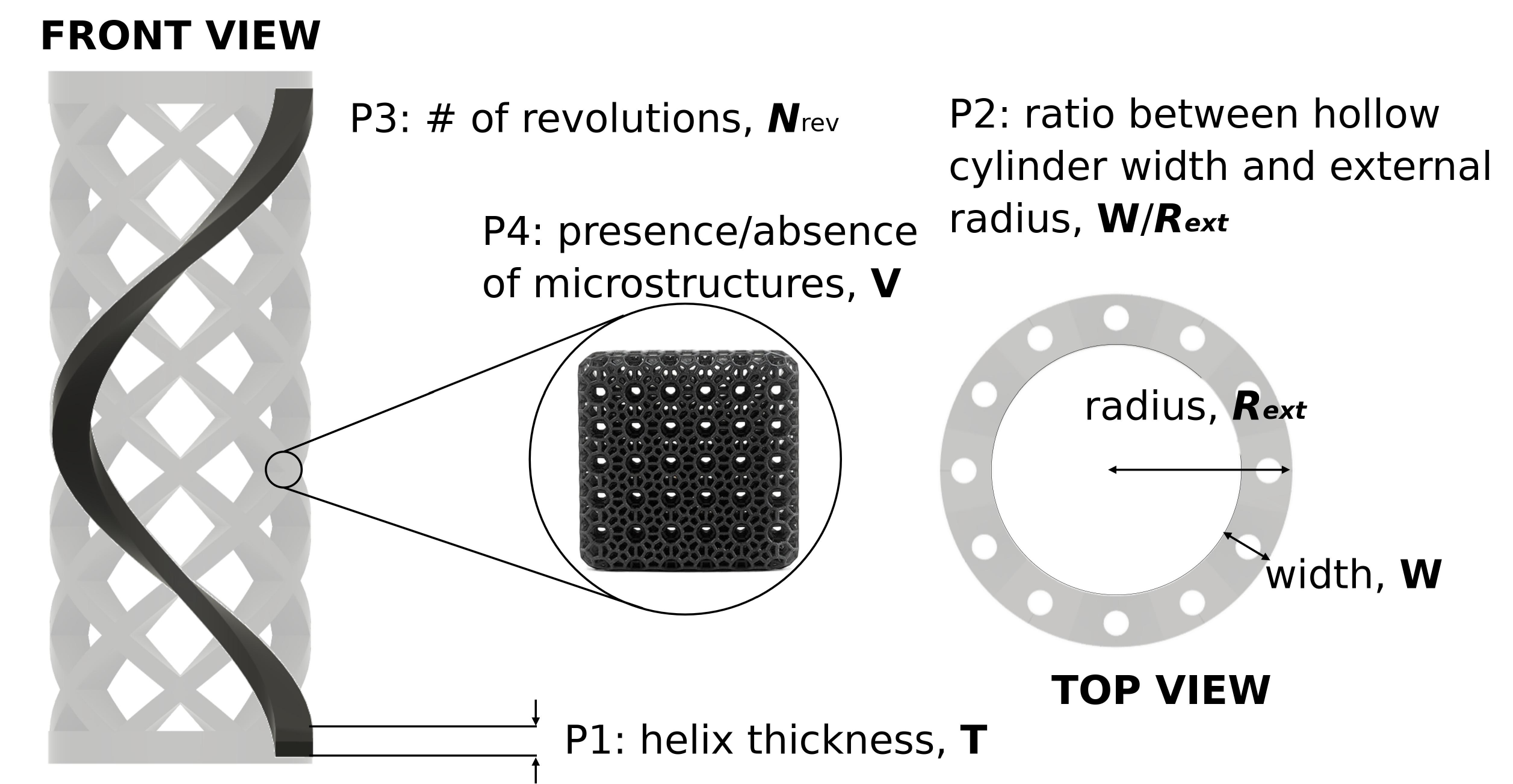}
            \caption{}
            \label{fig:design_space}
        \end{subfigure}
    \end{minipage}%
    \hfill%
    \begin{minipage}[t]{0.40\linewidth}%
        \vspace{0.8pt}%
        \centering
        \setcounter{subfigure}{1}
        \begin{subfigure}[t]{\linewidth}
            \centering
            \includegraphics[height=0.30\textheight]{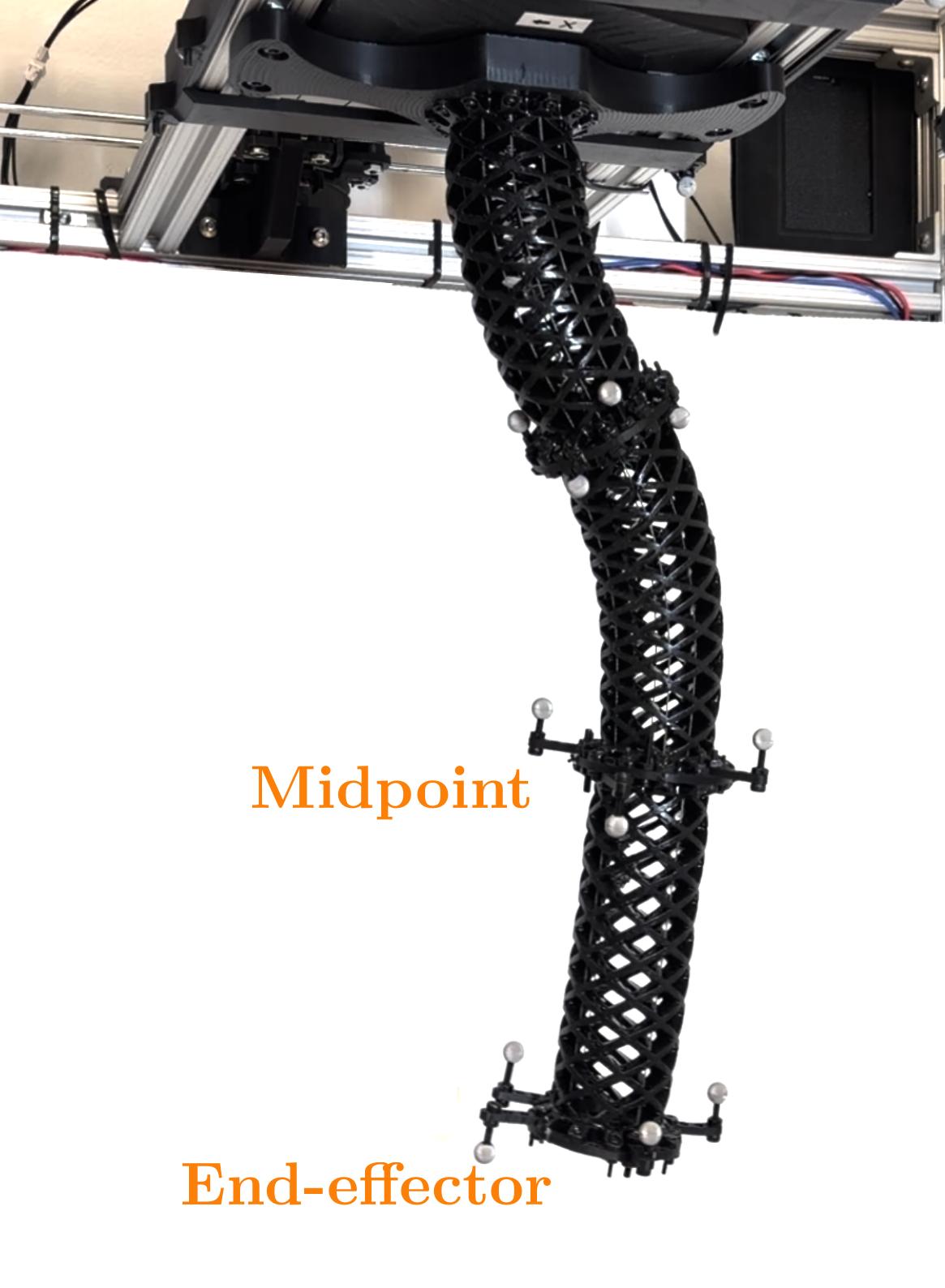}
            \caption{}
            \label{fig:experimental_setup}
        \end{subfigure}
    \end{minipage}
    \caption{(a) Final optimized geometric parameters of the Trimmed Helicoid module. (b) The three-module soft helicoid trunk robot is vertically mounted. The end-effector and midpoint positions are motion-tracked using OptiTrack markers (gray). (c) Design space: Helix thickness ($P1$), Trimming ratio ($P2$), Revolutions ($P3$), Microstructures ($P4$).}
    \label{fig:design_full}
\end{figure}

\subsection{Adiabatic spectral submanifolds for precision control}\label{subsec:aSSM_overview}
Motivated by general hardware setups for robotics, we formulate our control problem in the robot's workspace, defined by $\mathbf{z} \in \mathbb{R}^{n_w}$, under the admissible control inputs $\mathbf{u} \in \mathbf{R}^{n_u}$. The goal is to find optimal control inputs $\mathbf{u}^*(t)$ that force the robot to track a dynamic target track $\boldsymbol \Gamma (t) $ defined in its workspace. This objective can be formulated as the following finite-horizon optimal control problem 

\begin{align}
    \label{eq:mpc_general}
        \mathbf{u}^*(t) =\text{minimize}_{\mathbf{u}(\cdot)}~ \quad
        &\int_{0}^{t} \Big(||\mathbf{z}(s)-\mathbf{\Gamma}(s)||^2_\mathbf{Q_z} + ||\mathbf{u}(s)||^2_\mathbf{R_u} + ||\mathbf{\dot{u}}(s)||^2_\mathbf{R_{\dot{u}}}\Big) ds, 
\end{align}
subject to the dynamic evolution of the robot, state, and actuation constraints. Here, $\mathbf{Q_z}$, $\mathbf{R_u}$, and $\mathbf{R_{\dot{u}}}$ are tunable positive semi-definite control cost matrices that penalize the target tracking, the controls, and control rates, ensuring that the optimal control problem is well-posed.  

Our objective is to achieve precision control, which means solving eq.~(\ref{eq:mpc_general}) accurately with a reliable dynamic model of the robot. For soft robotic hardware, the exact governing equations are generally unknown, and only limited observational data are available. We denote this observable vector as $\mathbf{y} \in \mathbb{R}^{n_y}$, which in our soft helicoid system corresponds to its module ends tracked by the OptiTrack system (see Section~\ref{subsec:design}). 

To infer a reliable dynamic model, we observe that most soft robotic systems are heavily damped and hence exhibit an inherent dichotomy in time scales between the fast internal vibrations of the soft robot and its substantially slower trajectory speed during typical control tasks. This practical observation turns out to provide a representation of the soft robot's dynamics on slow invariant manifolds in its phase space. The underlying dynamics, when reduced onto these low-dimensional slow invariant manifolds, offer a rigorously justified modeling method to capture the core dynamics of the soft robot. 

More generally, spectral submanifolds (SSMs) and adiabatic spectral submanifolds (aSSMs) are low-dimensional attracting invariant surfaces in the phase space of the robot. Trajectories synchronize exponentially fast with their internal dynamics which therefore act as reduced-order models for the full system. SSMs are constructed locally near fixed points of the uncontrolled robot, while aSSMs can be constructed globally in the presence of the time-scale separation we assume. Recent work by \citet{Kaundinya2026} demonstrates clear performance gains when an aSSM-reduced model used in MPC instead of an SSM-reduced model.

The theory of aSSM reduction applies to general control systems of the form 
\begin{equation}
\label{eq:eom0}
\mathbf{\dot{x}} = \mathbf{F}\left(\mathbf{x}, \mathbf{u}(t) \right),
\end{equation}
where $\mathbf{x} \in \mathbb{R}^{n_x}$ denotes points in the phase space of the robot, $\mathbf u \in \mathbb{R}^{n_u}$ the control inputs and $\mathbf F$ is a smooth nonlinear function in its arguments. We assume that system eq.~(\ref{eq:eom0}) satisfies the following:
\begin{itemize}
    \item \textit{(A1)} The control input $\mathbf{u}(t)$ admits a splitting of the form
    \begin{equation}
    \label{eq:control_split}
    \mathbf{u}(t) \equiv \mathbf{u}^s(\epsilon t) + \mathbf{u}^d(t), \quad 0<\epsilon \ll 1, \quad \|\mathbf{u}^d(t)\| < \Delta \ll \|\mathbf{u}^s(\epsilon t)\|,
    \end{equation}
    for all times and for some constant $\Delta>0$.
    \item \textit{(A2)} System (\ref{eq:eom0}) admits a normally attracting critical manifold (i.e., manifold of fixed points), $\mathcal{L}_0$, in the limit $\mathbf{\dot{u}}(t)\equiv 0$ of static control inputs: 
    \begin{align}
    \label{eq:cm_def}
    \mathcal{L}_0 = \Big\{ \mathbf{x} \in \mathbb{R}^{n_x}:  \mathbf{x} = \mathbf{x}_0(\mathbf{u});\quad &\mathbf{F}(\mathbf{x}_0(\mathbf{u}),\mathbf{u}) \equiv 0, \quad  \mathbf{u} \in \mathcal{U} \Big\},
    \end{align}
    for some open set $\mathcal{U}\subset \mathbb{R}^{n_u}$.
    \item \textit{(A3)} The $d$ slowest eigenvalues of the spectrum of the linear operator family $\mathbf{D}_{\mathbf{x}}\mathbf{F}(\mathbf{x}_0(\mathbf{u}), \mathbf{u})$, evaluated along the critical manifold $\mathcal{L}_0$, satisfy non-resonance conditions detailed in \citet{haller24_wa}. 
\end{itemize}

Then eq.~(\ref{eq:eom0}) can rigorously be reduced to an $(n_u+d)$-dimensional aSSM-reduced model for $\epsilon$ and $\Delta$ small enough (see \citet{haller24_wa} and \citet{Kaundinya2026}). The geometry of this $(n_u+d)$-dimensional aSSM is a fiber bundle whose base, in the frozen limit ($\dot{\mathbf{u}}(t)\equiv \mathbf{0}$), is the $n_u$-dimensional critical manifold, and whose fibers are the $d$-dimensional slow SSMs anchored at each point of this critical manifold. In practice, these results extend to larger values of $\epsilon$ and $\Delta$ as well; the smallness of these parameters is required only for the rigorous existence proofs of the aSSM.

The above reduction assumes access to the full state $\mathbf{x}$. For soft robotic hardware, access to the full state space is absent ($n_y \ll n_x$). In fact, most motion capture systems record only positions and orientations from rigid-body markers attached to the soft robot. In such situations, we seek embeddings of the aSSM and its dynamics in the robot's observable space, $\mathbf{y} \in \mathbb{R}^{n_y}$, from data. Since direct velocity observations are unavailable, we need to use delay embeddings of the position observables, and seek the aSSM in this delay-embedded observable space. Takens' embedding theorem (see \citet{takens81}) guarantees that a $d$-dimensional fiber in the aSSM needs at least $p=\Big\lfloor \frac{2d+1}{n_y} \Big\rfloor$ delays of a nondegenerate observable $\mathbf {y} $ to be generically embeddable in a $(p+1)n_y$ delay-embedded observable space. We define this $p$ delay embedding function with delay parameter $\tau$ as
\begin{equation}
\label{eq:delay_embedding}
\boldsymbol \mu(\mathbf y(t)) = \begin{pmatrix}
\mathbf{y}(t) \\
\mathbf{y}(t-\tau) \\
\vdots \\
\mathbf{y}(t-p\tau)
\end{pmatrix}.
\end{equation}
As we will show in Section \ref{subsec:setup_theory}, the $(n_u+d)$-dimensional aSSM-reduced model in this delay embedded observable space is of the form 
\begin{align}
\label{eq:data-driven_case0}
\mathbf{r} &= \mathbf{V}^\mathrm{T}(\mathbf u)\boldsymbol \mu (\mathbf{y}-\mathbf{y}_0(\mathbf{u})), \qquad \mathbf{u} \in \mathcal{U}, \nonumber \\
\mathbf{\dot{r}} &= \mathbf{R}(\mathbf{r}, \mathbf{y}_0(\mathbf{u}(t))) + \mathbf{B}\left(\mathbf{u}(t)\right)   \mathbf{D}_{\mathbf{u}}\mathbf{y}_0(\mathbf u(t)) \dot{\mathbf{u}}, \\
\mathbf{y}(t) &=   \mathbf{W}(\mathbf{r}(t), \mathbf{y}_0(\mathbf{u}(t)))+ \mathbf y_0(\mathbf{u}(t)). \nonumber 
\end{align}
Here $\mathbf{r} \in \mathbb{R}^d$ are the reduced coordinates of the aSSM transverse to the critical manifold; $\mathbf{V}(\cdot) \in \mathbb{R}^{(p+1)n_y \times d}$ is the chart map whose column vectors span the tangent space of the aSSM; $\mathbf{y}_0(\cdot)$ is the parametrization of the critical manifold in the observable space; $\mathbf{R}(\mathbf{r},\mathbf{y}_0 (\cdot))$ is the leading-order reduced dynamics on the aSSM; $\mathbf{B}(\cdot)$ is a calibration matrix for the higher-order terms of the aSSM-reduced dynamics; and $\mathbf{W}(\mathbf{r},\mathbf{y}_0 (\cdot))$ is the leading-order aSSM parametrization. In Section \ref{subsec:setup_theory}, we provide a self-contained derivation of the aSSM-reduced model for this setting, which differs in the choice of parametrization of the aSSM from \citet{haller24_wa} and \citet{Kaundinya2026}.

\begin{figure}[t!]
    \centering
    \begin{subfigure}[b]{0.48\linewidth}
        \centering
        \includegraphics[width=\linewidth]{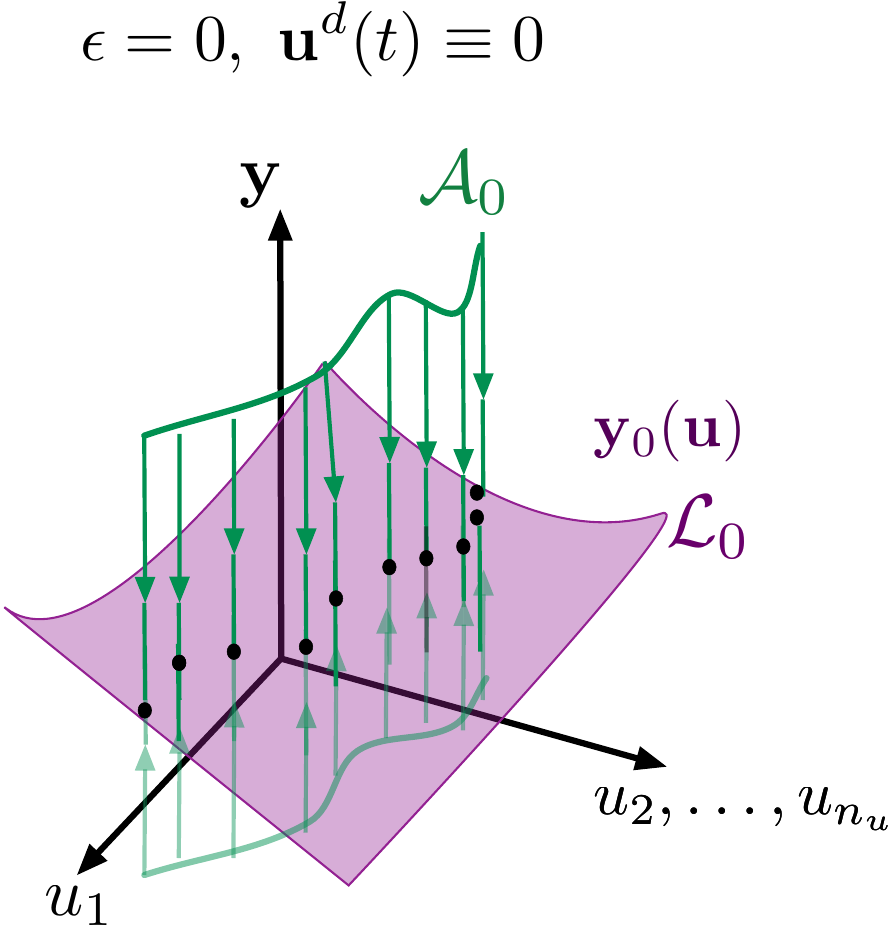}
        \caption{}
        \label{fig:aSSM_frozen}
    \end{subfigure}
    \hfill
     \begin{subfigure}[b]{0.48\linewidth}
        \centering
        \includegraphics[width=\linewidth]{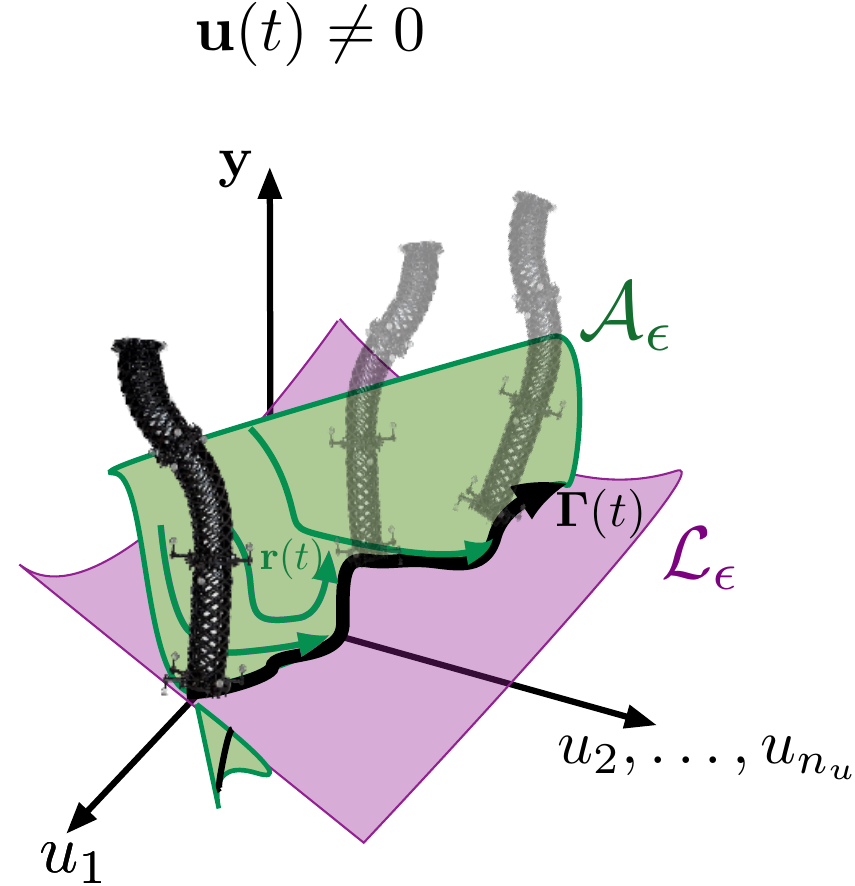}
        \caption{}
        \label{fig:aSSM_u}
    \end{subfigure}

    \caption{Schematic for (a) the frozen limit of the aSSM geometry and (b) the dynamic aSSM geometry for time-varying control inputs.}
    \label{fig:aSSM_geometry}
\end{figure}

The functions $\mathbf{V}$, $\mathbf{y}_0$ $\mathbf{R}$, $\mathbf{B}$, and $\mathbf{W}$ are all learnable from data. To this end, we need to pass control inputs to the robot that probe aSSM geometry. The frozen aSSM geometry ($\epsilon =0$ and $\mathbf{u}^d(t) \equiv 0$ in eq.~(\ref{eq:control_split})) comprises the critical manifold $\mathcal{L}_0$ and the static slow SSM bundle, $\mathcal{A}_0$, attached to it (see Fig.~\ref{fig:aSSM_frozen}). This frozen aSSM geometry is directly learnable from data. Specifically, passing random step control inputs to the robot allows one to probe $\mathcal{L}_0$ and $\mathcal{A}_0$. The leading-order aSSM-reduced model is obtained by passing dynamic control inputs $\mathbf{u}(t)$ to the learned frozen aSSM model (see Fig.~\ref{fig:aSSM_u}).

\FloatBarrier

In Section \ref{subsec:data-driven}, we provide details on learning the leading-order aSSM-related functions from random step control input data, $\{\mathbf{Y}_{\mathrm{step}}, \mathbf{U}_{\mathrm{step}}\}$. We also provide details to learn the higher-order calibration function  $\mathbf{B}$ from random control input data, $\{\mathbf{Y}_{\mathrm{calib}}, \mathbf{U}_{\mathrm{calib}}\}$, and the learned leading-order aSSM-related functions.

We use the aSSM-reduced model (\ref{eq:control_split}) learned from data to make predictions on previously unseen general control inputs $\mathbf{u}(t)$ given initial conditions in the robot's observable space obtained from motion capture. The aSSM-reduced model obtained in this way accurately captures the robot's dynamics and hence can be used as a constraint in a model predictive control (MPC) scheme with an objective function defined in eq.~(\ref{eq:mpc_general}). For MPC, we also define the state ($\mathcal{Y}$) and control ($\mathcal{U}$) constraint sets, and a mapping from the observable space to the workspace,   
\begin{equation}
\label{eq:target_map}
\mathbf{z} = \mathbf{C}(\mathbf y), \quad  \mathbf y \in \mathcal{Y}, \quad  \mathbf{u} \in \mathcal{U}. 
\end{equation}
Solving eq.~(\ref{eq:mpc_general}) subject to constraints eq.~(\ref{eq:data-driven_case0})-(\ref{eq:target_map}) provides us with a closed-loop controller for target tracking tasks. In Section~\ref{subsec:mpc}, we provide the techniques to solve this optimization problem in real-time.

\subsection{Modeling the soft trunk dynamics via aSSM}\label{subsec:modeling}

The camera setup for our soft trunk platform can record the end effector and the mid-section of the trunk. We henceforth denote our observable space of the soft trunk as $\mathbf{y} = \{x_{ee},y_{ee},z_{ee},x_{mid},y_{mid},z_{mid}\}$. Our workspace comprises the end-effector positions $\mathbf{z} = \{x_{ee},y_{ee},z_{ee}\}$, for which we will plan trajectory tracking tasks. These tracks are executed via a total of six motors that tension the tendons along the trunk's arm. This amounts to $n_u = 6$ control inputs. 

The soft trunk decay profiles indicate a strongly damped system. The symmetry of the trunk with respect to the $x_{ee}$ and $y_{ee}$ axes yields a near $1-1$ resonance between the slowest modes of vibration near the origin $\mathbf {y} = \mathbf{0}$. We note the resonance is allowed by SSM theory, as we recall in \citet{haller16}. The soft trunk's geometry also allows for at least one decaying mode that is present in the $z_{ee}$ direction. Therefore, depending on the nature of the decays (overdamped or underdamped), we conclude that at least a $d=5$ or $ d=6$\textemdash dimensional fiber in the aSSM can capture the dynamics of the soft trunk robot. We note that these observations also applied to the aSSM-reduced model derived for trunk simulators with similar geometries by \citet{Kaundinya2026}. From our analysis in Section \ref{subsec:aSSM_overview}, we expect these $d=5$ or $d=6$ dimensional fibers of the aSSM to be embedded in a $6(p+1)$ dimensional delay-embedded observed space with $p\geq 2$. 

As a first step in building the aSSM from data, we pass random step control inputs to the soft trunk robot for 30 minutes. These step inputs are sampled to ensure that the observable space of the trunk is uniformly probed (see Section \ref{subsec:data-driven}). Next, we collect 1 hour 30 minutes of slow, smooth, random control inputs, with the sampling done via a Perlin noise scheme (see \citet{perlin85}). These experimental runs provide us with the datasets $\{\mathbf{Y}_{\mathrm{step}},\mathbf{U}_{\mathrm{step}}\}$ and $\{\mathbf{Y}_{\mathrm{calib}},\mathbf{U}_{\mathrm{calib}}\}$.

\begin{figure}[t!]
    \centering
    \begin{subfigure}[b]{0.49\linewidth}
        \centering
        \includegraphics[width=\linewidth]{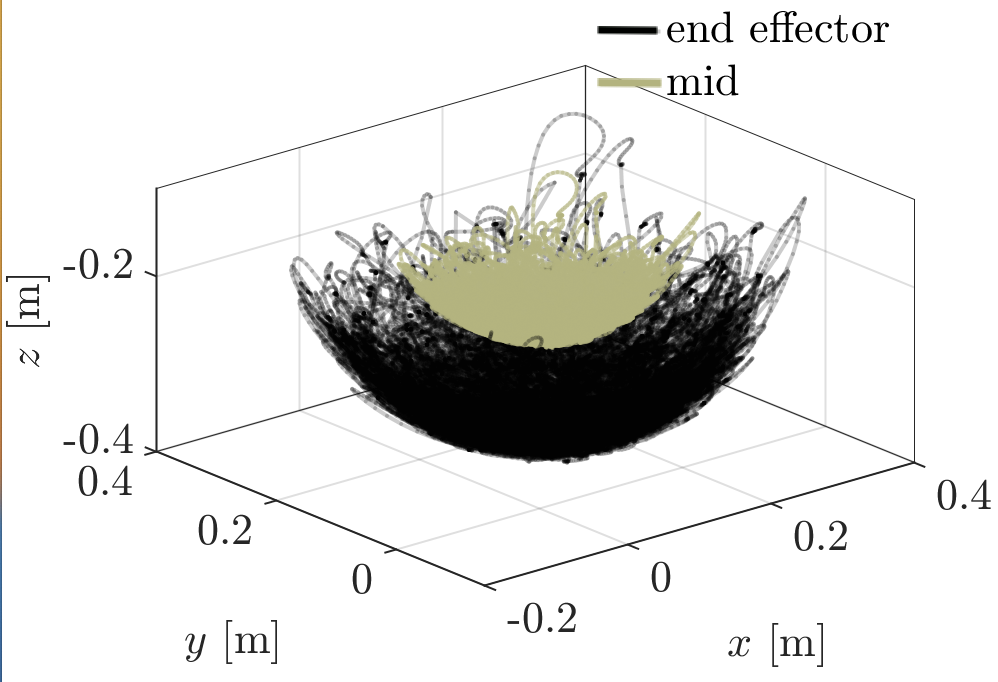}
        \caption{}
        \label{fig:data}
    \end{subfigure}
    \begin{subfigure}[b]{0.49\linewidth}
        \centering
        \includegraphics[width=\linewidth]{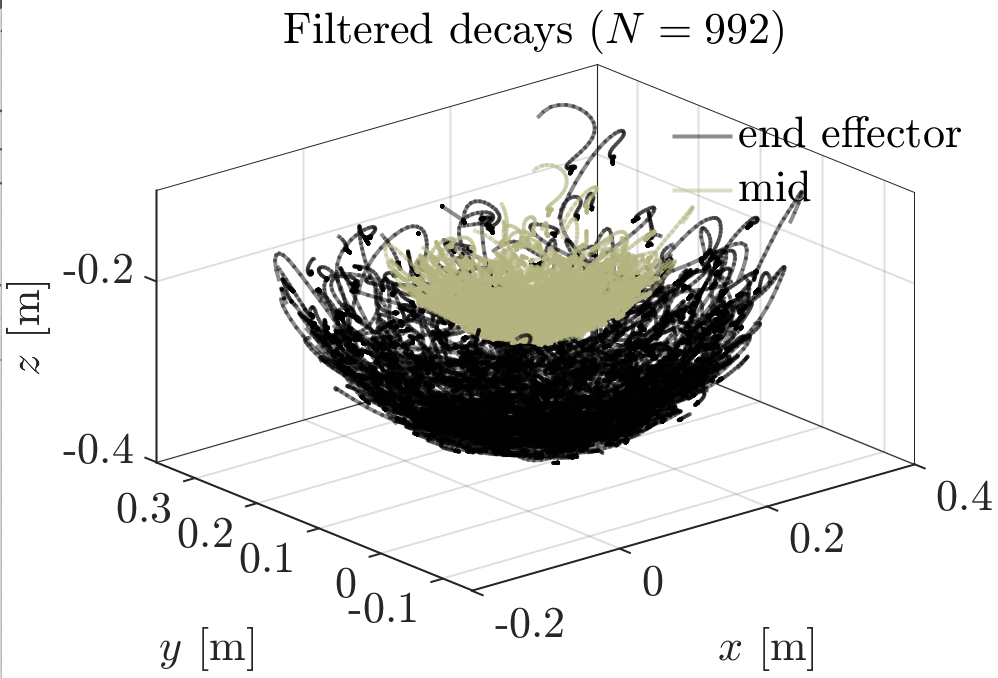}
        \caption{}
        \label{fig:f_data}
    \end{subfigure}

    \begin{subfigure}[b]{0.6\linewidth}
        \centering
        \includegraphics[width=\linewidth]{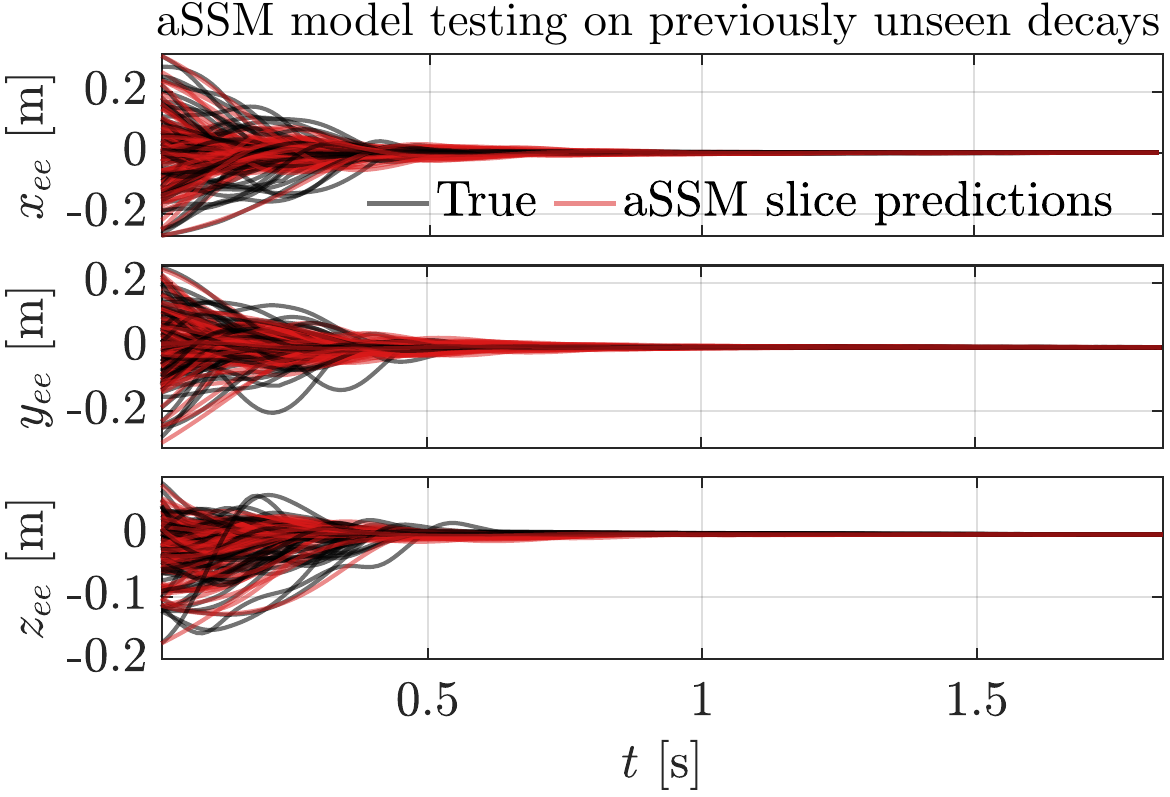}
        \caption{}
        \label{fig:pred0}
    \end{subfigure}

    \begin{subfigure}[b]{0.7\linewidth}
        \centering
        \includegraphics[width=\linewidth]{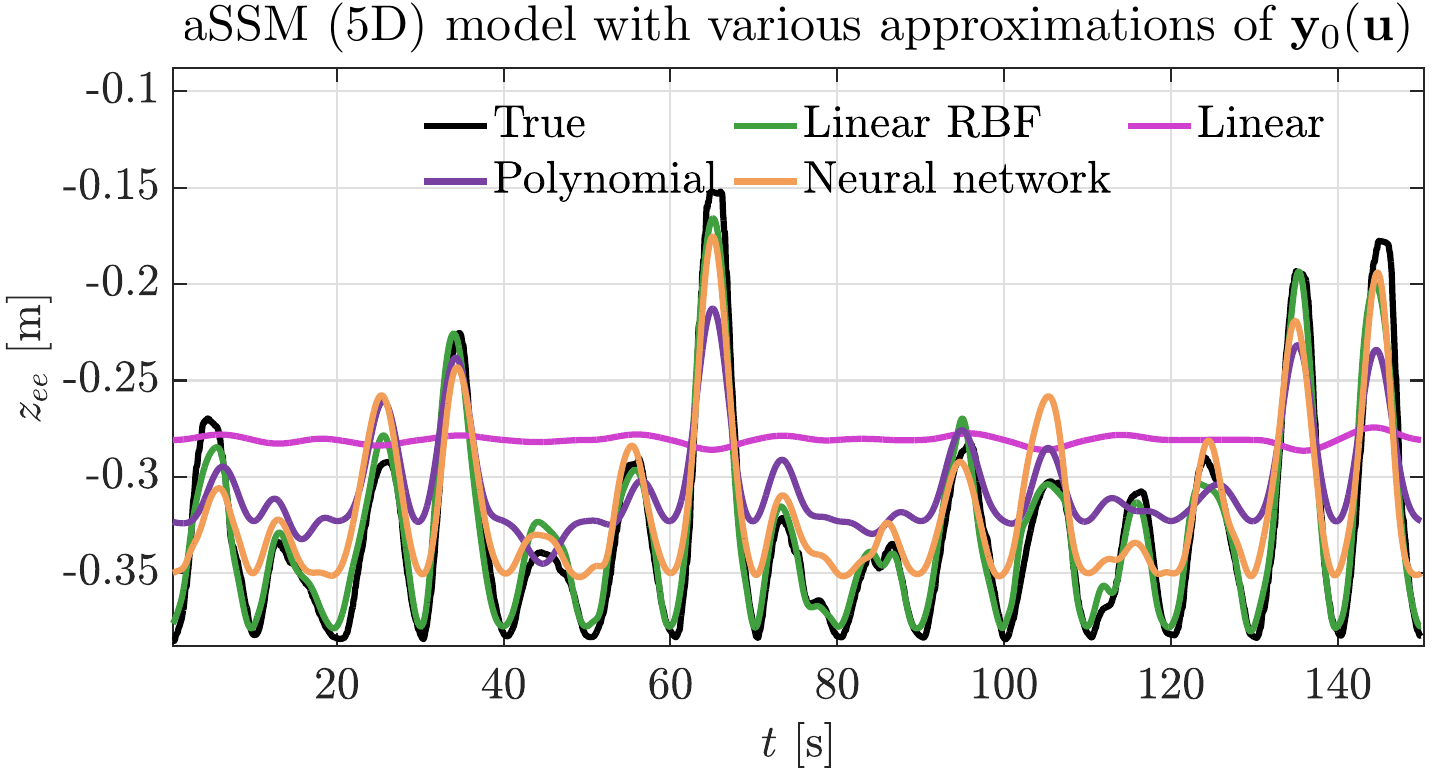}
        \caption{}
        \label{fig:pred}
    \end{subfigure}

    \caption{(a) Step and random input sequence responses in the soft trunk's observable space. (b) Filtered decay plots after pre-processing the step input dataset $\mathbf{Y}_{\mathrm{step}}$. (c) Predictions on previously unseen decay data of the leading-order aSSM-reduced model for static inputs. (d) Prediction of the aSSM-reduced model for unseen random control inputs using various approximations of the critical manifold $\mathcal{L}_0$.}
    \label{fig:modeling}
\end{figure}

In Fig.~\ref{fig:data}, we plot the collected trajectories in the observable space. For learning aSSM-reduced models, we also need to post-process the decaying dataset $\{\mathbf{Y}_{\mathrm{step}},\mathbf{U}_{\mathrm{step}}\}$ dataset as outlined in Section \ref{subsec:data-driven}. The post-processed dataset isolates decays and static states. For this dataset, we isolated $N=992$ static states, each with $10$ clean decays (see Fig.~\ref{fig:f_data}). We further split these datasets randomly into $90 \%$  for training and $10 \%$ for testing. We use the training dataset and follow the steps outlined in Section \ref{subsec:learn_cm} for learning the critical manifold $\mathbf{y}_0(\mathbf{u})$ and in Section \ref{subsec:learn_aSSM} for learning the aSSM-reduced model. 

We learn a $(n_u+d)=(6+5)$\textemdash dimensional aSSM-reduced model (see eq.~(\ref{eq:aSSMLearn_1})-(\ref{eq:aSSMLearn_fin})) with parameters $k^{W}_{r} = 2$ and $k^{W}_{y} = 2$  for the aSSM parametrization,  $k^{R}_{r} = 2$ and $k^{R}_{y} = 2$  for the reduced dynamics, and $k^{B}_{u} = 2$ for the calibration in a $6(p+1)$- dimensional delay-embedded observer space with $p=6$. The normalized mean trajectory error on unseen test decaying data for the leading-order aSSM-reduced model is $9.6 \%$ (see Fig. \ref{fig:pred0} for centered predictions). We also test the aSSM-reduced model with various approximations of the critical manifold $\mathcal{L}_0$. We find that a linear radial basis function with $n_c=6,000$ centers yields the lowest error compared to linear, polynomial, or neural network models (see Fig. \ref{fig:pred}) on unseen control input data. Hence, we identify a $(n_u+d)=(6+5)$\textemdash dimensional aSSM-reduced model with a critical manifold approximated by linear radial basis functions to model the soft trunk's dynamics accurately. Since $n_u$ is the dimension of the control inputs that parametrize the model, and the reduced ODE is $d$-dimensional, we hereafter denote the aSSM-reduced model as aSSM ($d$D).

\FloatBarrier

\subsection{Closed-loop results for target tracking}\label{subsec:closed-loop}

We use our data-driven aSSM-reduced model for the closed-loop control of target tracking in the workspace of the robot. The model is justified with full mathematical rigor when the target speed is slow compared to the robot's internal dynamics. To quantify such regimes of applicability, we recall from \citet{Kaundinya2026} the non-dimensional relative slowness measure $r_s$ defined for a target track $\boldsymbol{\Gamma}(t)$ over a finite time interval $[0, T]$ as 
\begin{equation}
    r_{s} = \frac{\frac{1}{T}\int_{0}^{T} \|\dot{ \boldsymbol{\Gamma}}(s)\| ds}{\overline{\frac{1}{t_d}\int_{0}^{t_d} \|\dot{\mathbf{z}}_{\text{free decays}}(s)\| ds}},
\end{equation}
Here ${\mathbf{z}}_{\text{free decays}}(t)$ are the intrinsic vibrations of the soft robot near its static states; the overbar denotes averaging over various initial conditions in the workspace of the robot; and $t_d$ is the total duration of the non-zero part of the decaying signals. We expect accurate target tracking when $r_s < 1$, as this implies strict time-scale separation, a key assumption for the theory of aSSMs to hold. Note that low values of $r_s$ correspond to slowly varying tracks relative to the robot's intrinsic dynamics, such tracks do not necessarily indicate targets that are perceived as slow by a human observer.

In Section \ref{subsec:mpc}, we provide details for the closed-loop controller for target tracking with an aSSM-reduced model. We found that choices $\mathbf{R_u} = 10^{-4}\cdot  \mathbb{I}_{6}$ and $\mathbf{R_{\dot{u}}} = 10^{-3}\cdot \mathbb{I}_{6}$ for control and control rate costs in eq.~(\ref{eq:aSSM_mpc0}) led to physically relevant control input magnitudes for the aSSM-reduced model. For 2D target tracking tasks, we set $\mathbf{Q_z} = \text{diag}\left(1000,1000,0\right)$ and for 3D target tracking tasks, we set $\mathbf{Q_z} = \text{diag}\left(100,100,100\right)$. For MPC horizon lengths up to $M=20$, we can operate the aSSM-based controller at 50 Hz. We used $M=10$ for the target tracking task and $M=20$ for the obstacle avoidance tasks. 

We also present here results for four baselines: a 6D DMDc model (see \citet{Proctor2016}), a 6D Koopman-pregain model (see \citet{haggerty23}), a 33D MZ-Koopman model (see \citet{mori_soft_26}), and an RL (PPO) control policy (see \citet{schulman2017proximalpolicyoptimizationalgorithms}). We report the optimal controller settings for these models in the Supplementary Section \ref{app:baselines}.

Our interest is to evaluate the dynamic precision of the various methods for several closed-loop tasks. During closed-loop hardware operations, various delays can occur for all models at the software and hardware levels. These delays result in errors between the target and the closed-loop prediction $\mathbf{z}_{\text{pred}}$ at the time instant $t$. To this end, to measure dynamic accuracy as well as spatial accuracy, we adopt the sliding-window-type relative pointwise error metric 
\begin{equation}
\text{Pointwise relative error}(t) = \min_{\tau \,\in\, \left[t-\delta t,\; t+\delta t\right]} \left\| \mathbf{z}_{\text{pred}}(t) - \boldsymbol{\Gamma}(\tau) \right\|,
\end{equation}
with the half-window spacing set to $\delta t =1$ [s]. 

We also note that our current motor actuation mechanisms allow for reproducible dynamic responses with a mean maximum error of $3$ [cm] observed for the end-effector. Hence, we expect our model predictions to perform well if they are within an uncertain tube of this radius around the target track. We will plot such tubes around our target tracks in the workspace and also show these as lines in the pointwise relative error plots.   

\subsubsection{Figure 8 track}\label{subsec:8}

We consider here a large Figure-8 track defined in the $x_{ee}-y_{ee}$ plane. The maximum width and length span of the Figure-8 is $0.2$ [m]. This is six times larger than the Figure-8 track considered by \citet{alora24} for the hardware diamond robot, and it is nearly half the length of the soft trunk robot ($0.42$ [m]). We run closed-loop experiments for a total duration of $30$ [s] with two different target speeds: (i) a Figure-8 track with the mean instantaneous speed $v_{\boldsymbol{\Gamma}} = 8 \text{ [cm/s]}$ and slowness measure $r_s = 0.17$ (ii) a 1.5$\times$ faster varying track with the mean instantaneous speed $v_{\boldsymbol{\Gamma}} = 13 \text{ [cm/s]}$ and slowness measure $r_s = 0.25$.

The closed-loop results for the target speed case (i) for all models are presented in the left panel in Fig.~\ref{fig:8} and for the faster speed case (ii) in the right panel of the same figure. Qualitatively, we observe across both target speeds, only the aSSM-reduced MPC scheme tracks the Figure-8 within the gray uncertainty region. The Koopman-pregain comes second, with a visible Figure-8, but the plots are misaligned in orientation and hence exit the uncertainty region. The RL (PPO) policy's performance resembles a Figure-8, but it jitters quite a bit around the traced track. Compared with the Koopman-pregain method, it shows a larger misalignment from the original track. The MZ-Koopman controller also predicts a Figure-8 track, but it is off-center, exhibits jitter, and is significantly smaller than the original target track. The DMDc model-based controller performs poorly, and the resulting track hardly resembles the Figure-8. 

We also provide quantitative results with pointwise error plots for the speed case (i) in Fig.~\ref{fig:8_slow_error} and the speed case (ii) in Fig.~\ref{fig:8_fast_error}. We find that the aSSM-reduced MPC scheme offers consistent performance across the target track and track speeds, with pointwise relative error remaining mostly below the tolerance line at $3$ [cm]. Unlike the gray tube in Fig.~\ref{fig:8}, which only checks whether the trajectory stays close to the target track, the pointwise error is evaluated at each time instance and therefore also penalizes delays and deviations from the reference. This explains the three instances where the pointwise error of the aSSM exceeds the tolerance, even though the trajectory remains within the tube. All other models consistently cross the tolerance line. We also see that overall errors increase as the target speed is increased. We also note that on the faster track, there are regions where the aSSM-reduced model has consistently low errors that are comparable to its performance on the regular speed track.

\newcommand{\panelw}{0.26\linewidth}              
\newcommand{\labfont}{\footnotesize}              
\newcommand{\xlab}{$x_{ee}\ [\mathrm{m}]$}
\newcommand{\ylab}{$y_{ee}\ [\mathrm{m}]$}
\newcommand{\ylabsep}{2pt}                        
\newcommand{\xlabsep}{1pt}                        
\newcommand{\panelgap}{6pt}                       
\newcommand{\rowgap}{\noalign{\vskip\panelgap}}   
\newcommand{\trimmedimg}[1]{%
  \includegraphics[width=\panelw,trim=34 36 16 7,clip]{#1}}

\newcommand{\vc}[1]{$\vcenter{\hbox{#1}}$}
\newcommand{\ylabbox}{\vc{\rotatebox{90}{\labfont\ylab}}}
\newcommand{\panel}[1]{%
  \begin{tabular}{@{}c@{\hspace{\ylabsep}}c@{}}
    \ylabbox & \vc{\trimmedimg{#1}} \\[\xlabsep]
             & {\labfont\xlab}
  \end{tabular}}
\newcommand{\overimg}[1]{%
  \begin{tabular}{@{}c@{\hspace{\ylabsep}}c@{}}
    \hphantom{\ylabbox} & \makebox[\panelw]{#1}
  \end{tabular}}
\newcommand{\mtitle}[2]{\overimg{\textcolor[RGB]{#1}{\bfseries\small #2}}}

\begin{figure}[H]
    \centering
    \setlength{\tabcolsep}{0pt}
    \renewcommand{\arraystretch}{0}
    \begin{tabular}{@{}c@{\hspace{0.06\linewidth}}c@{}}
        \overimg{\small$v_\Gamma = 9\ [\mathrm{cm/s}]$} &
        \overimg{\small$v_\Gamma = 13\ [\mathrm{cm/s}]$} \\[5pt]
        \mtitle{18,126,4}{aSSM (5D)} & \mtitle{18,126,4}{aSSM (5D)} \\
        \panel{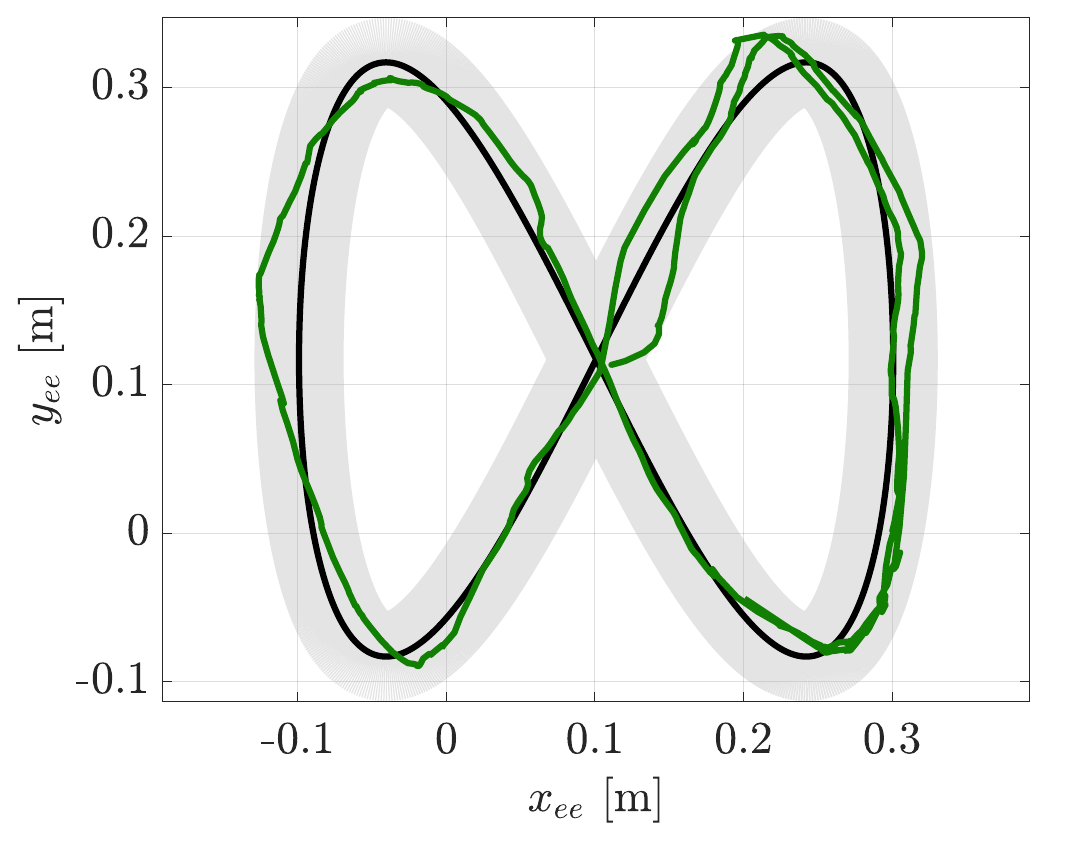} &
        \panel{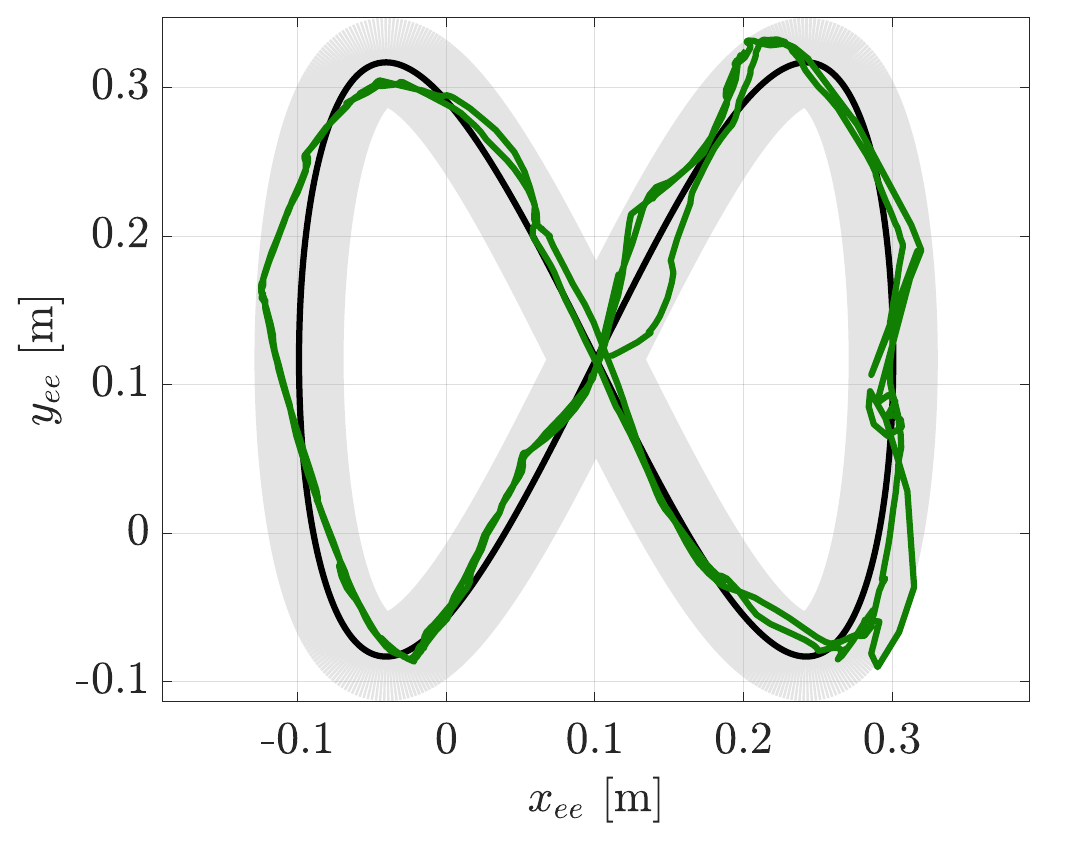} \\ \rowgap
        \mtitle{128,0,128}{DMDc (6D)} & \mtitle{128,0,128}{DMDc (6D)} \\
        \panel{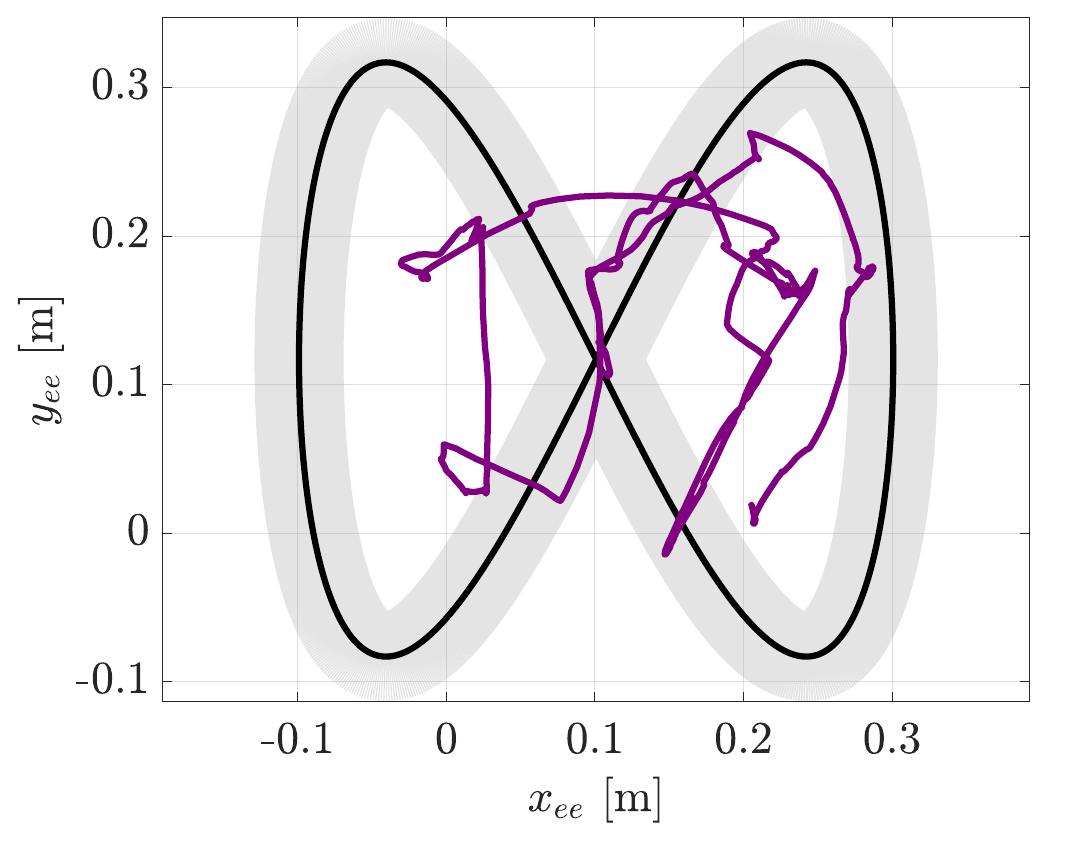} &
        \panel{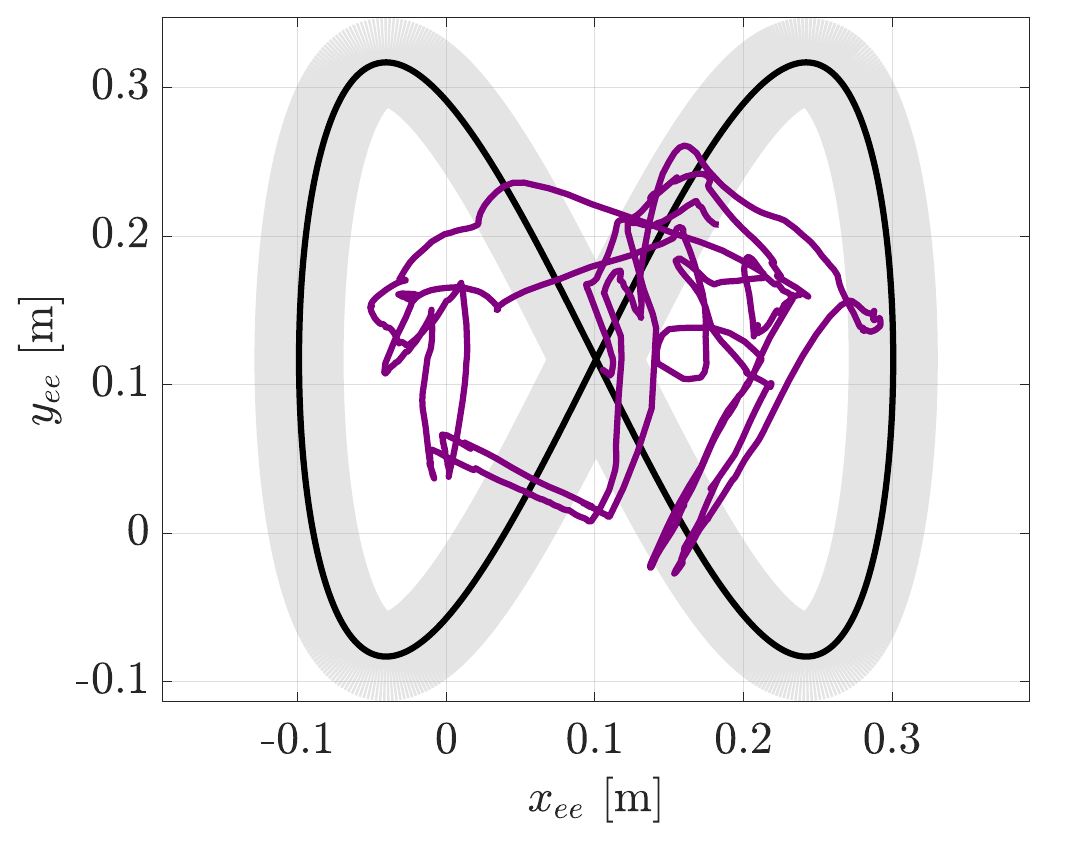} \\ \rowgap
        \mtitle{237,28,36}{Koopman-pregain (6D)} & \mtitle{237,28,36}{Koopman-pregain (6D)} \\
        \panel{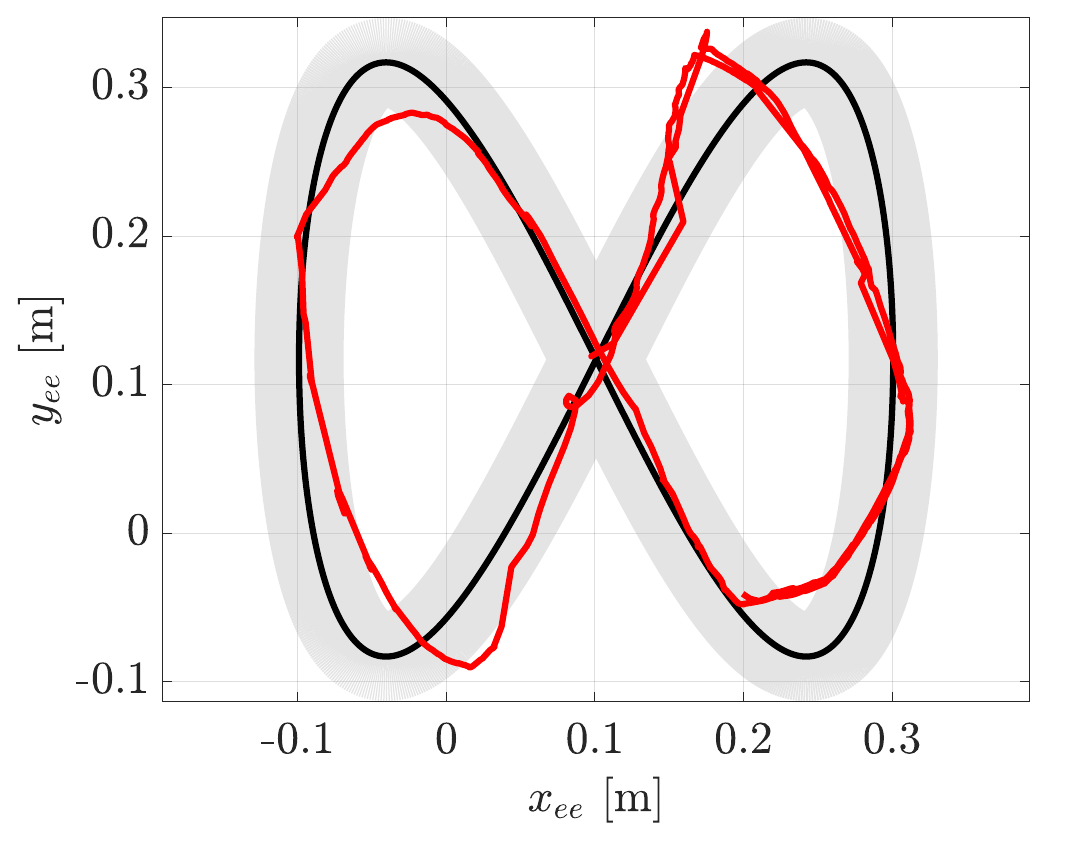} &
        \panel{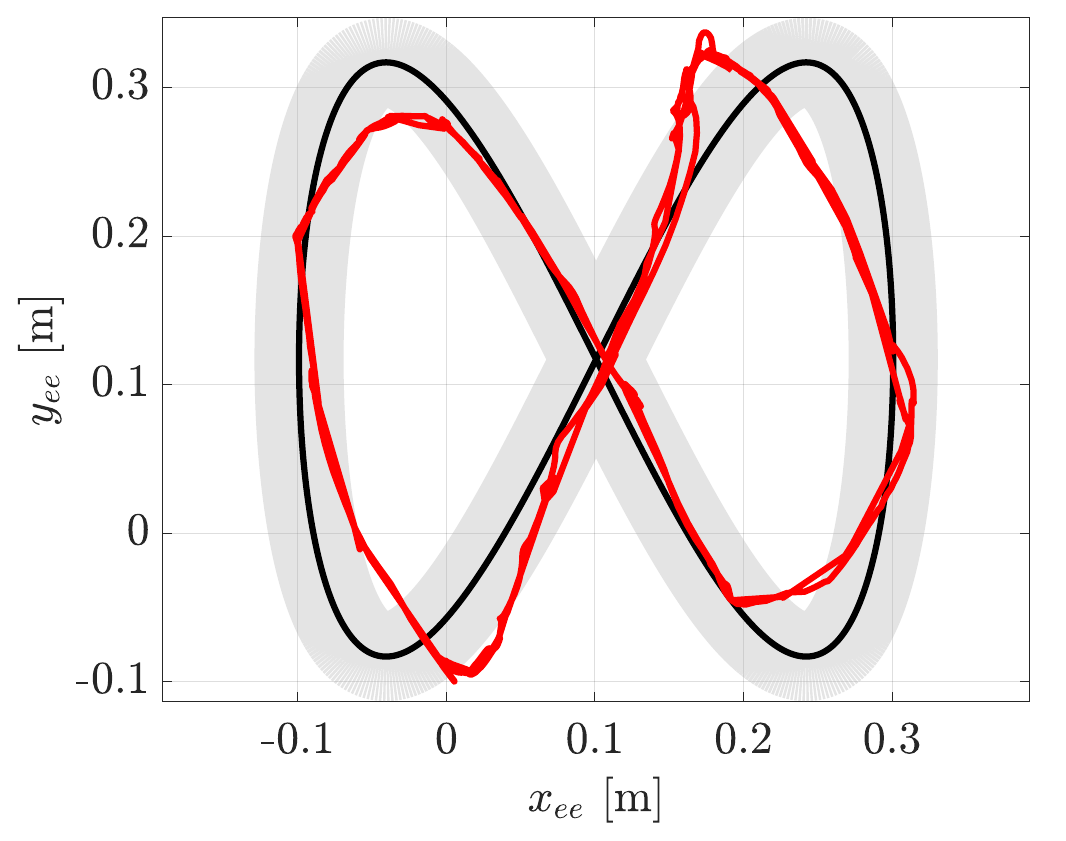} \\ \rowgap
        \mtitle{237,125,49}{MZ-Koopman (33D)} & \mtitle{237,125,49}{MZ-Koopman (33D)} \\
        \panel{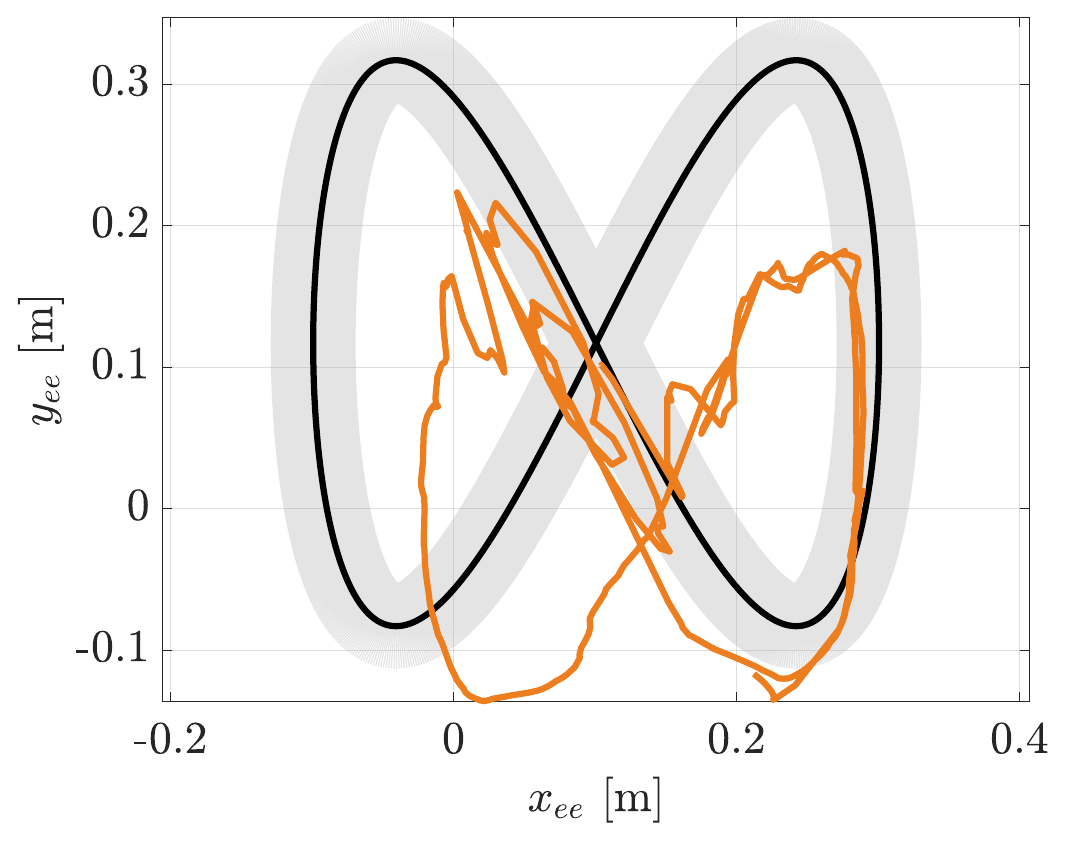} &
        \panel{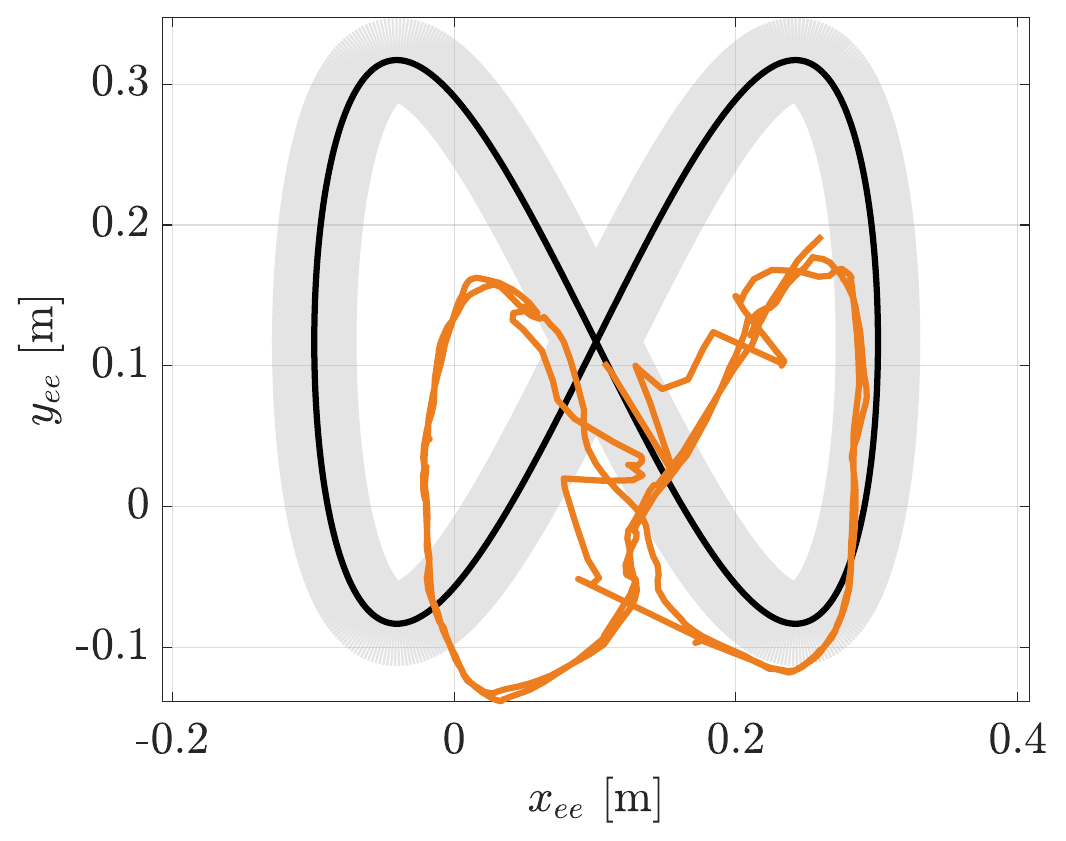} \\ \rowgap
        \mtitle{0,114,189}{RL (PPO)} & \mtitle{0,114,189}{RL (PPO)} \\
        \panel{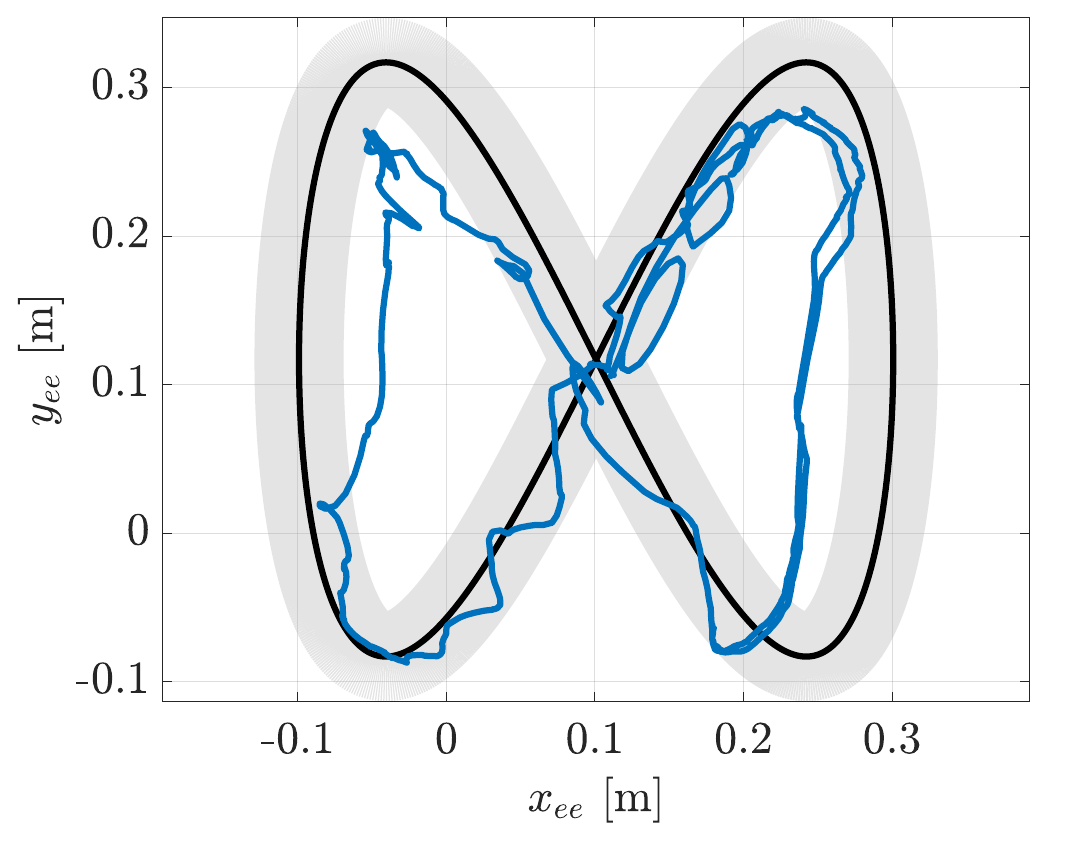} &
        \panel{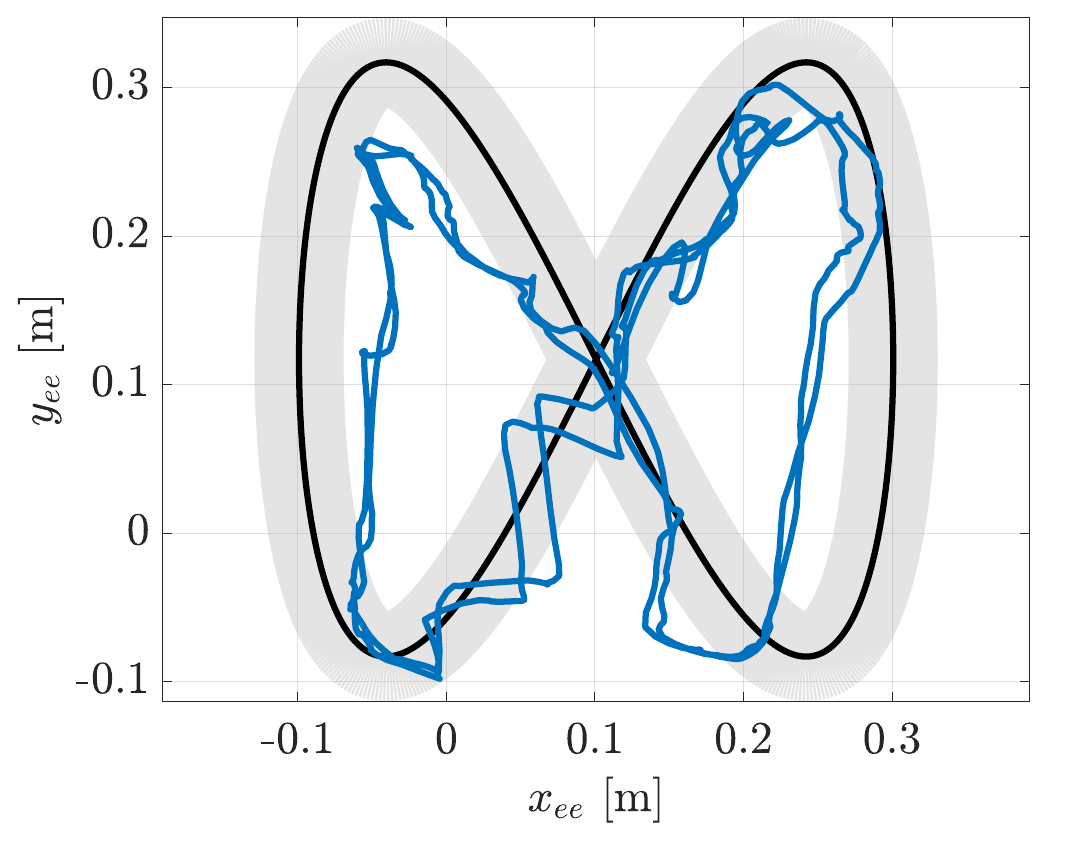} \\
    \end{tabular}
    \caption{Closed-loop tracking results for a Figure-8 target (black) with the uncertainty band of width $3$ [cm] plotted in gray.}
    \label{fig:8}
\end{figure}

\begin{figure}[H]
    \centering
    \includegraphics[width=0.66\linewidth]{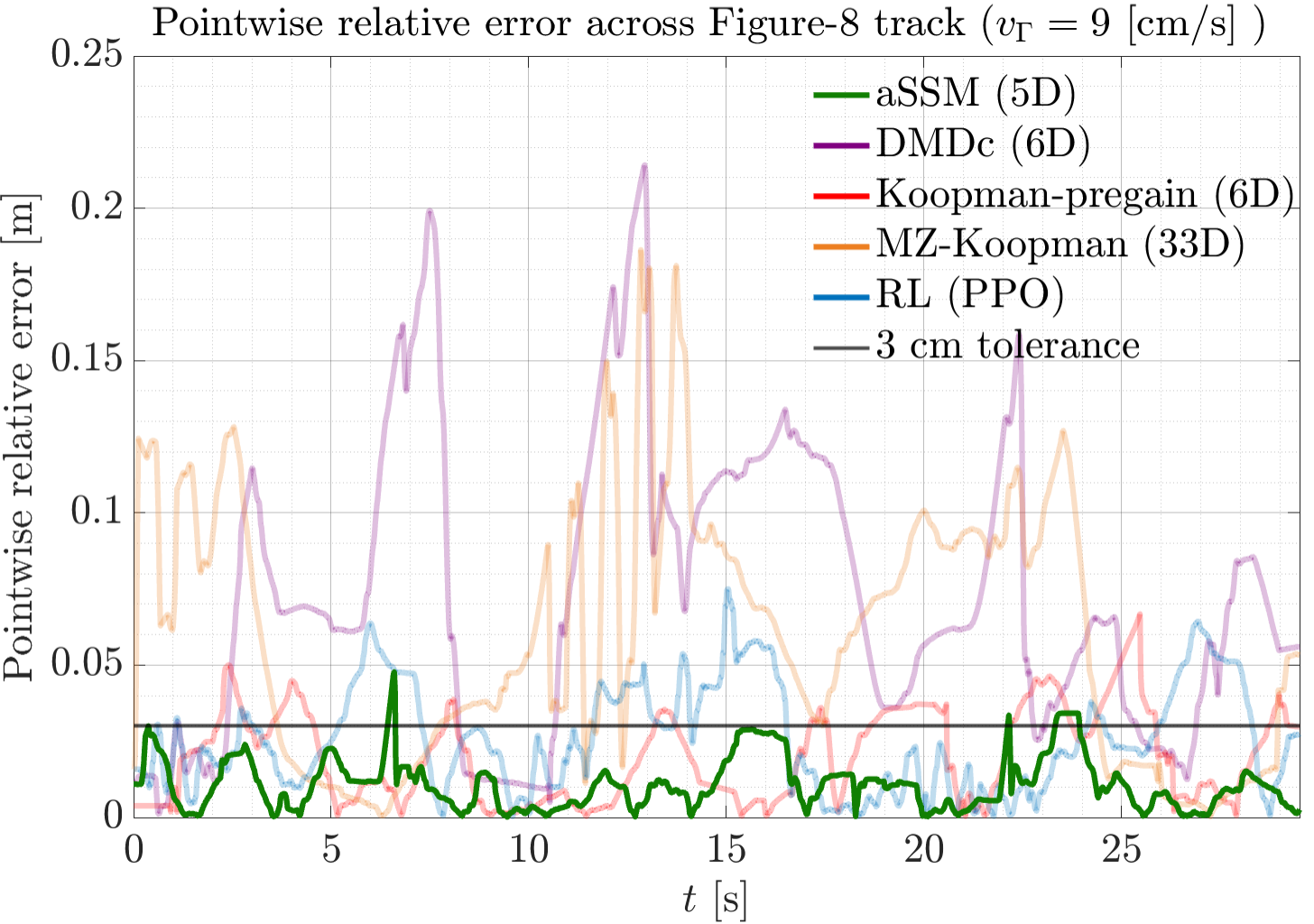}
    \caption{Pointwise relative error for closed-loop tracking of a Figure-8 track with speed $9$ [cm/s].}
    \label{fig:8_slow_error}
\end{figure}

\begin{figure}[H]
    \centering
    \includegraphics[width=0.66\linewidth]{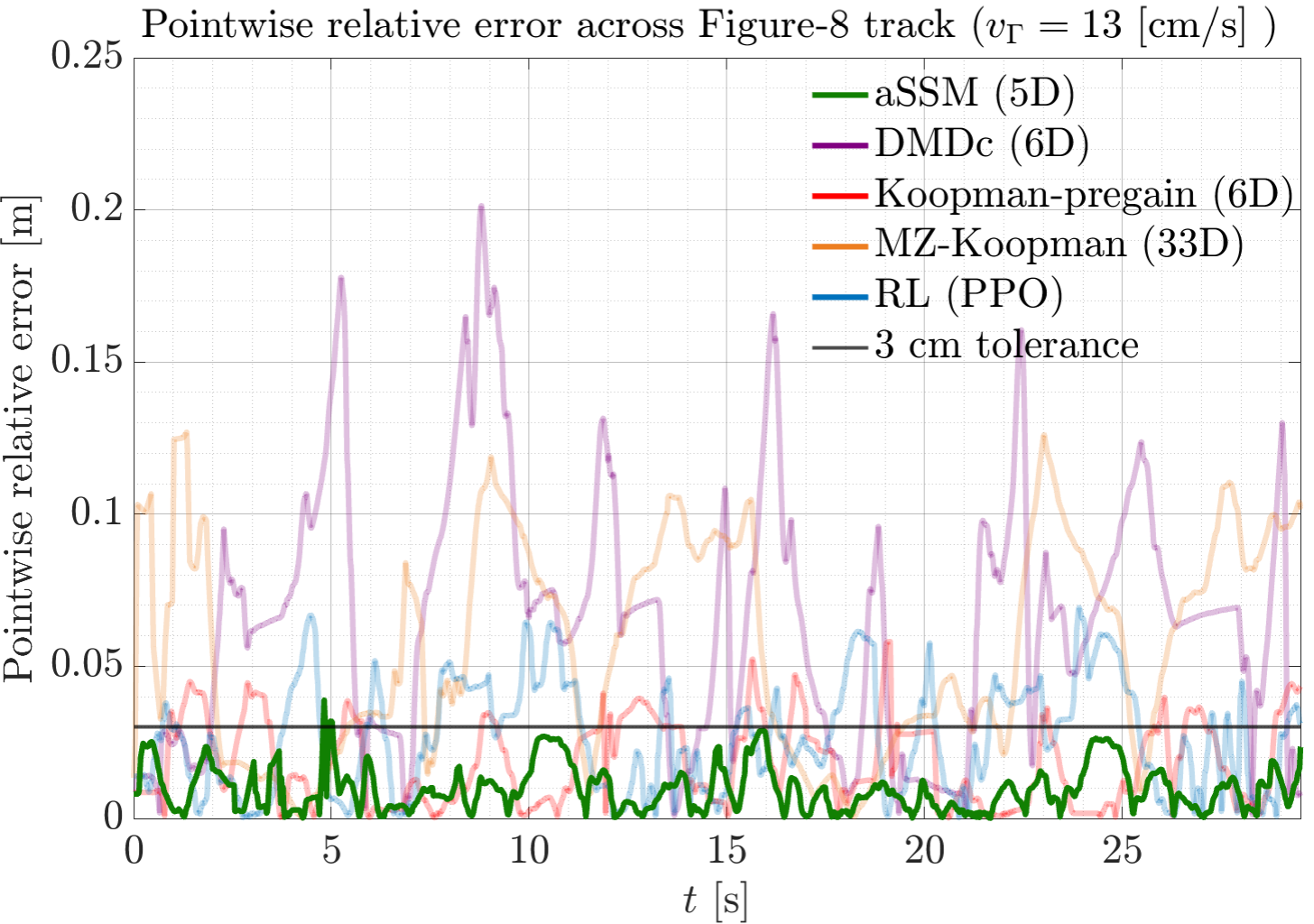}
    \caption{Pointwise relative error for closed-loop tracking of a faster Figure-8 track.}
    \label{fig:8_fast_error}
\end{figure}

\FloatBarrier

\subsubsection{3D Random track}\label{subsec:random}

Our second target track is a challenging 30-second-long 3D Random target track in the trunk's workspace, generated by passing a smooth random control input sequence. This is not seen in the training control data used to learn the various models and the RL policy. The mean instantaneous speed of the target is $v_{\boldsymbol{\Gamma}} = 5$ [cm/s] with a slowness measure $r_s = 0.1$.

\begin{figure}[H]
    \centering
    \setlength{\tabcolsep}{0pt}
    \begin{tabular}{@{}p{0.49\linewidth}@{\hspace{0.02\linewidth}}p{0.49\linewidth}@{}}
        \centering\textcolor[RGB]{18,126,4}{\bfseries\normalsize aSSM (5D)} &
        \centering\textcolor[RGB]{128,0,128}{\bfseries\normalsize DMDc (6D)} \tabularnewline[-2pt]
        \includegraphics[pagebox=mediabox, trim=0pt 4pt 51pt 30pt, clip, width=\linewidth]{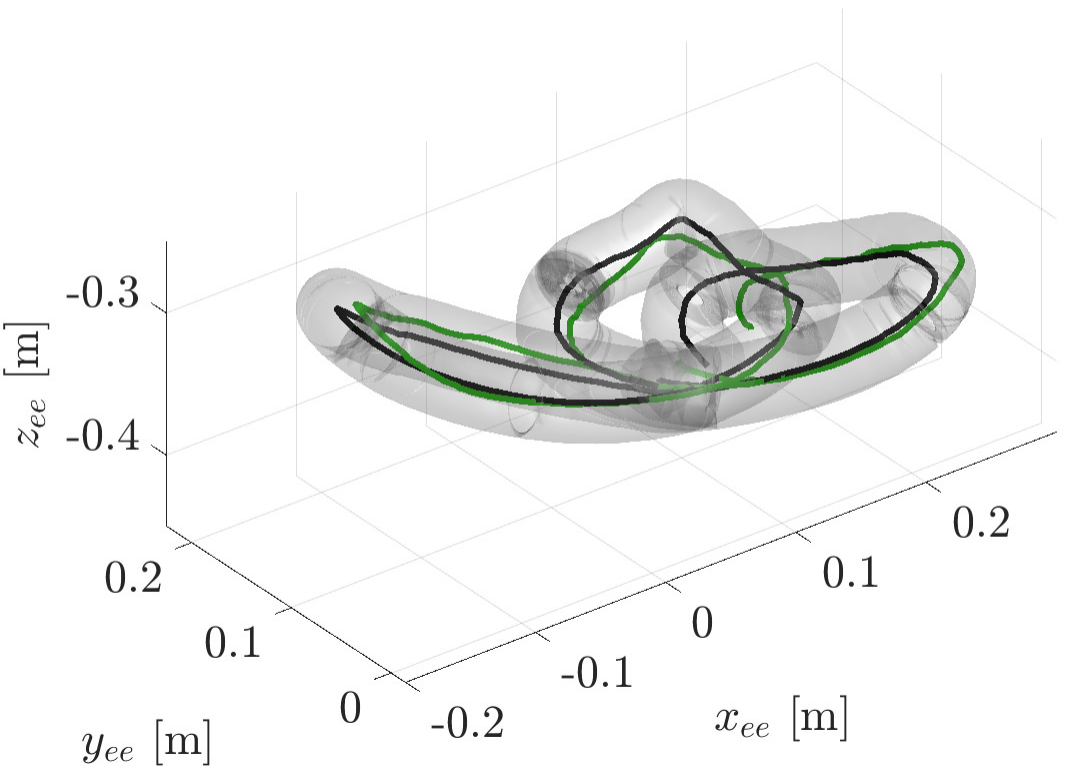} &
        \includegraphics[pagebox=mediabox, trim=0pt 4pt 51pt 30pt, clip, width=\linewidth]{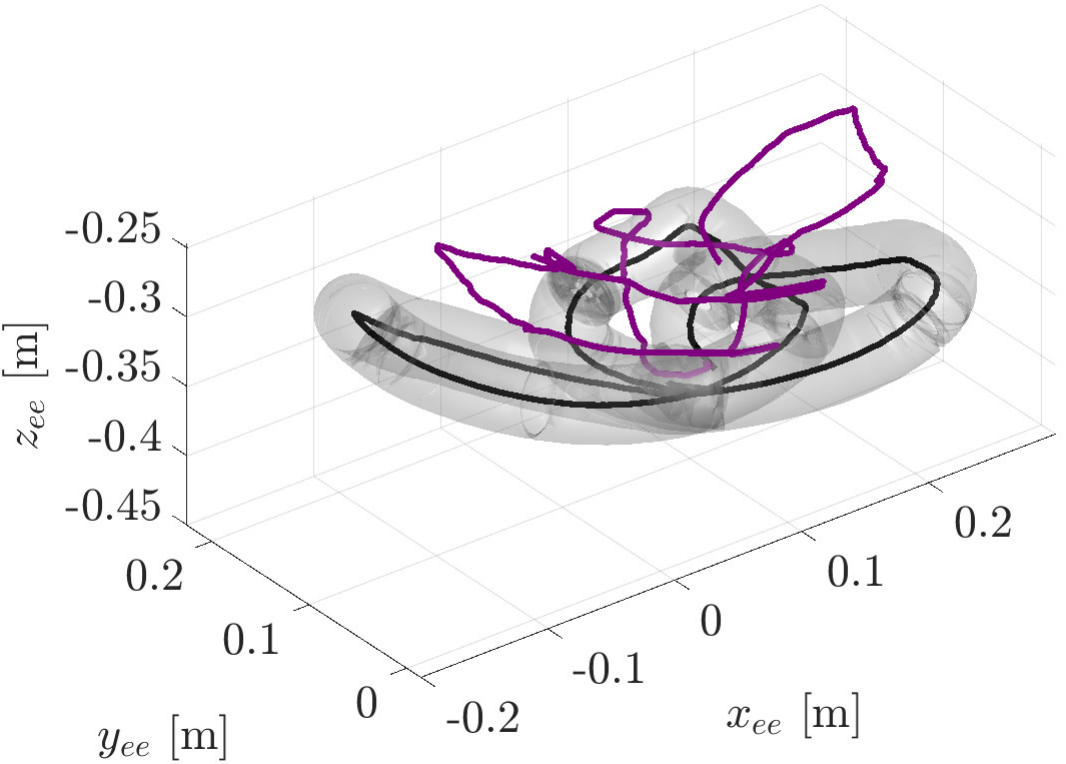} \tabularnewline[4pt]
        \centering\textcolor[RGB]{237,28,36}{\bfseries\normalsize Koopman-pregain (6D)} &
        \centering\textcolor[RGB]{237,125,49}{\bfseries\normalsize MZ-Koopman (33D)} \tabularnewline[-2pt]
        \includegraphics[pagebox=mediabox, trim=0pt 4pt 51pt 30pt, clip, width=\linewidth]{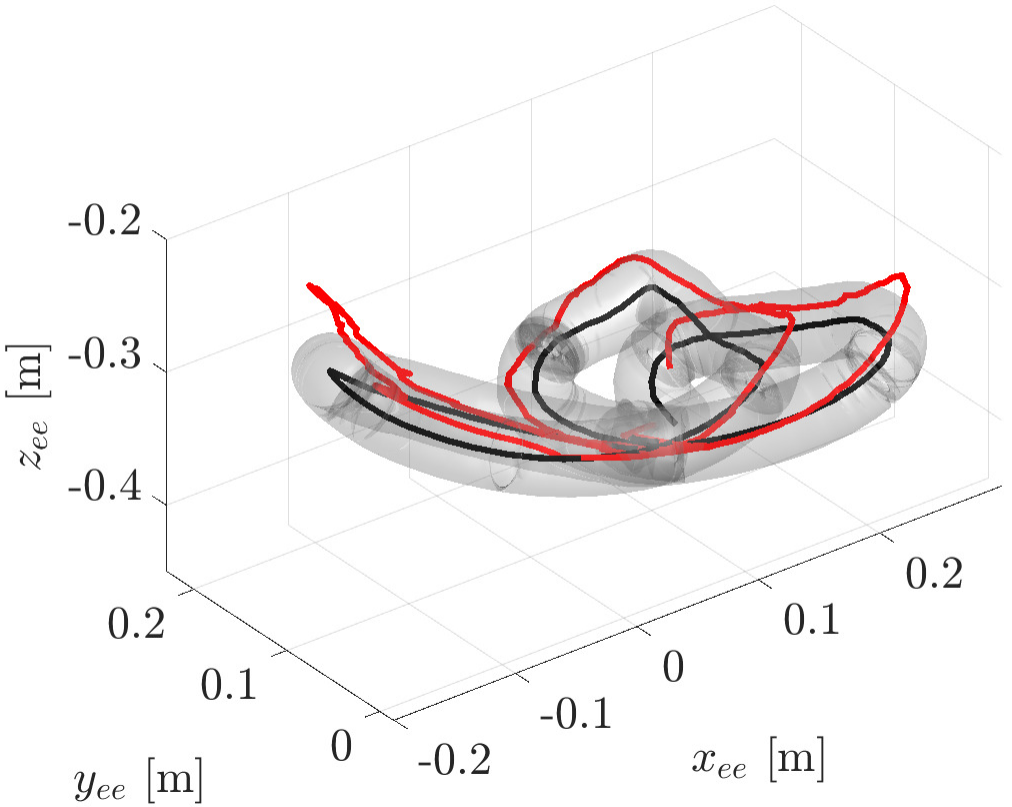} &
        \includegraphics[pagebox=mediabox, trim=0pt 4pt 51pt 30pt, clip, width=\linewidth]{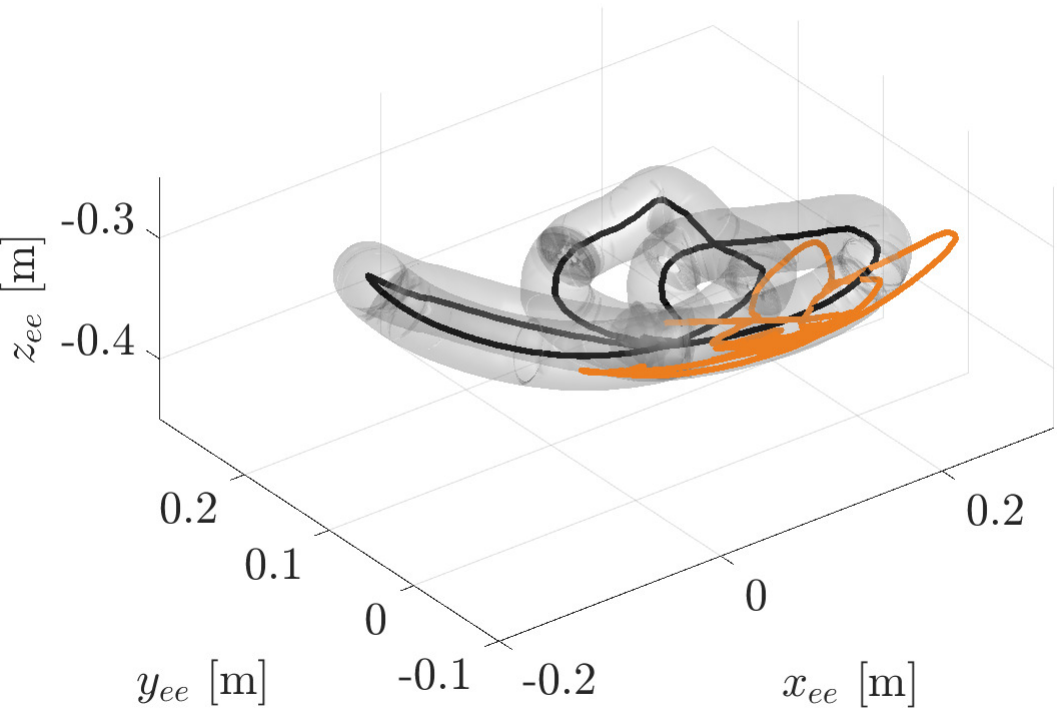} \tabularnewline[4pt]
        \multicolumn{2}{c}{\textcolor[RGB]{0,114,189}{\bfseries\normalsize RL (PPO)}} \tabularnewline[-2pt]
        \multicolumn{2}{c}{\parbox{0.49\linewidth}{%
            \includegraphics[pagebox=mediabox, trim=0pt 4pt 51pt 30pt, clip, width=\linewidth]{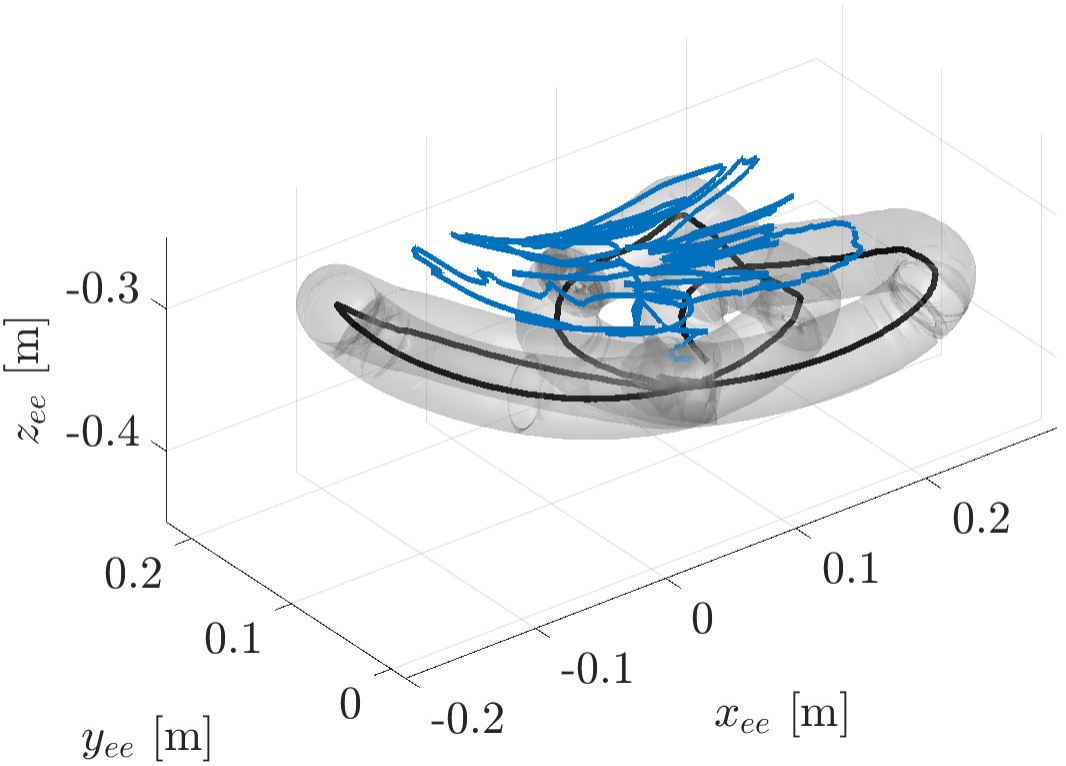}}}
        \tabularnewline
    \end{tabular}
    \caption{Closed-loop tracking results for a 3D Random track (black) with the
    uncertainty tube of radius $3$ [cm] plotted in gray.}
    \label{fig:random_cl}
\end{figure}

Along the 3D Random track, we reach a maximum height of $0.1$ [m], and along the $x_{ee}-y_{ee}$ plane, we have a maximum width of $0.4$ [m]. This track explores the extremal boundaries of the trunk robot's workspace.

\newcommand{\modeltitle}[2]{%
  \textcolor{#1}{\sffamily\bfseries\large #2}\par\vspace{2pt}}
\begin{figure}[H]
    \centering
    \begin{subfigure}[b]{0.35\linewidth}
        \centering
        \modeltitle{cOurs}{aSSM (5D)}
        \includegraphics[width=\linewidth]{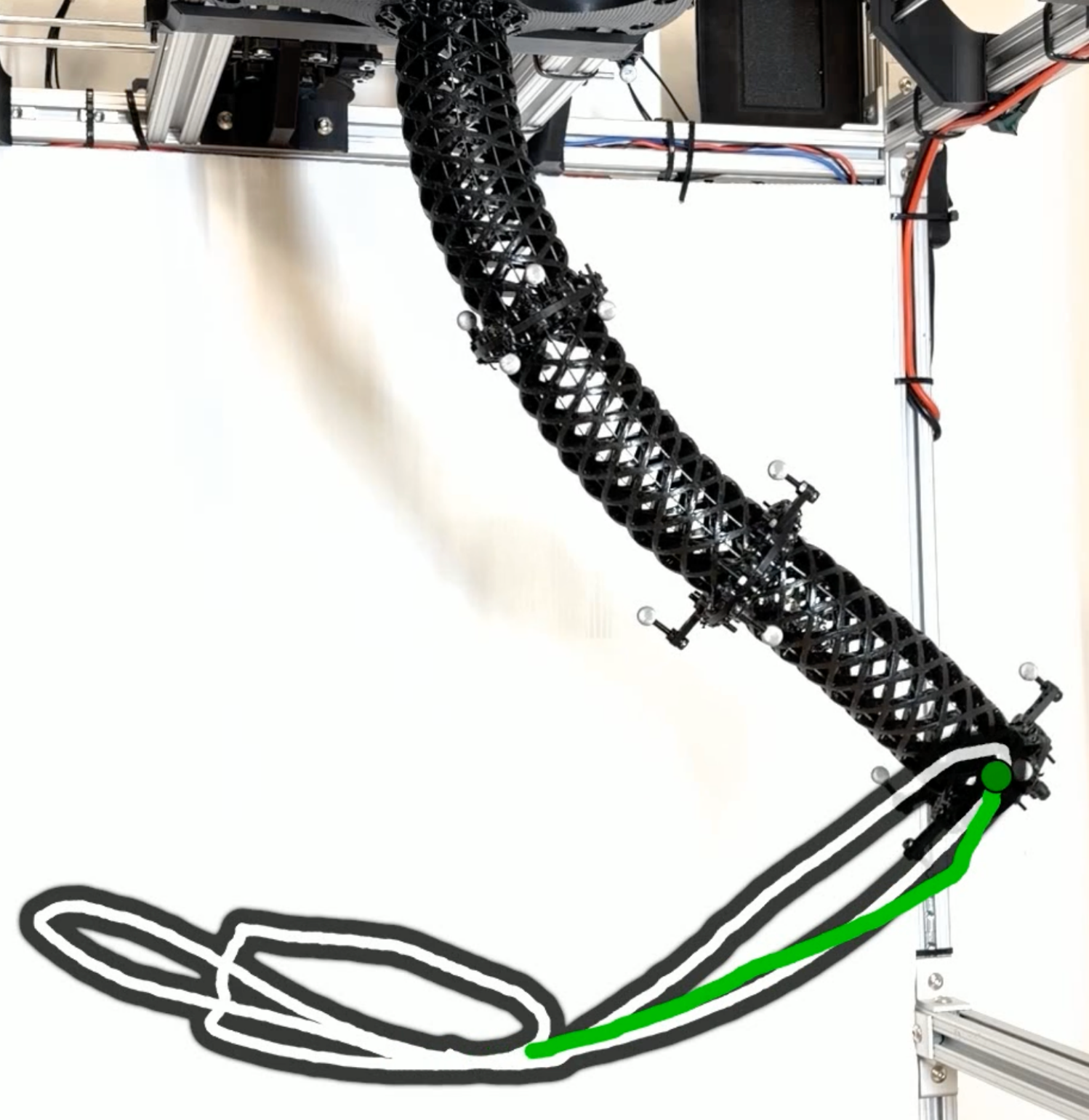}
        \label{fig:hw_aSSM}
    \end{subfigure}
    \hspace{0.02\linewidth}
    \begin{subfigure}[b]{0.35\linewidth}
        \centering
        \modeltitle{cDMDc}{DMDc (6D)}
        \includegraphics[width=\linewidth]{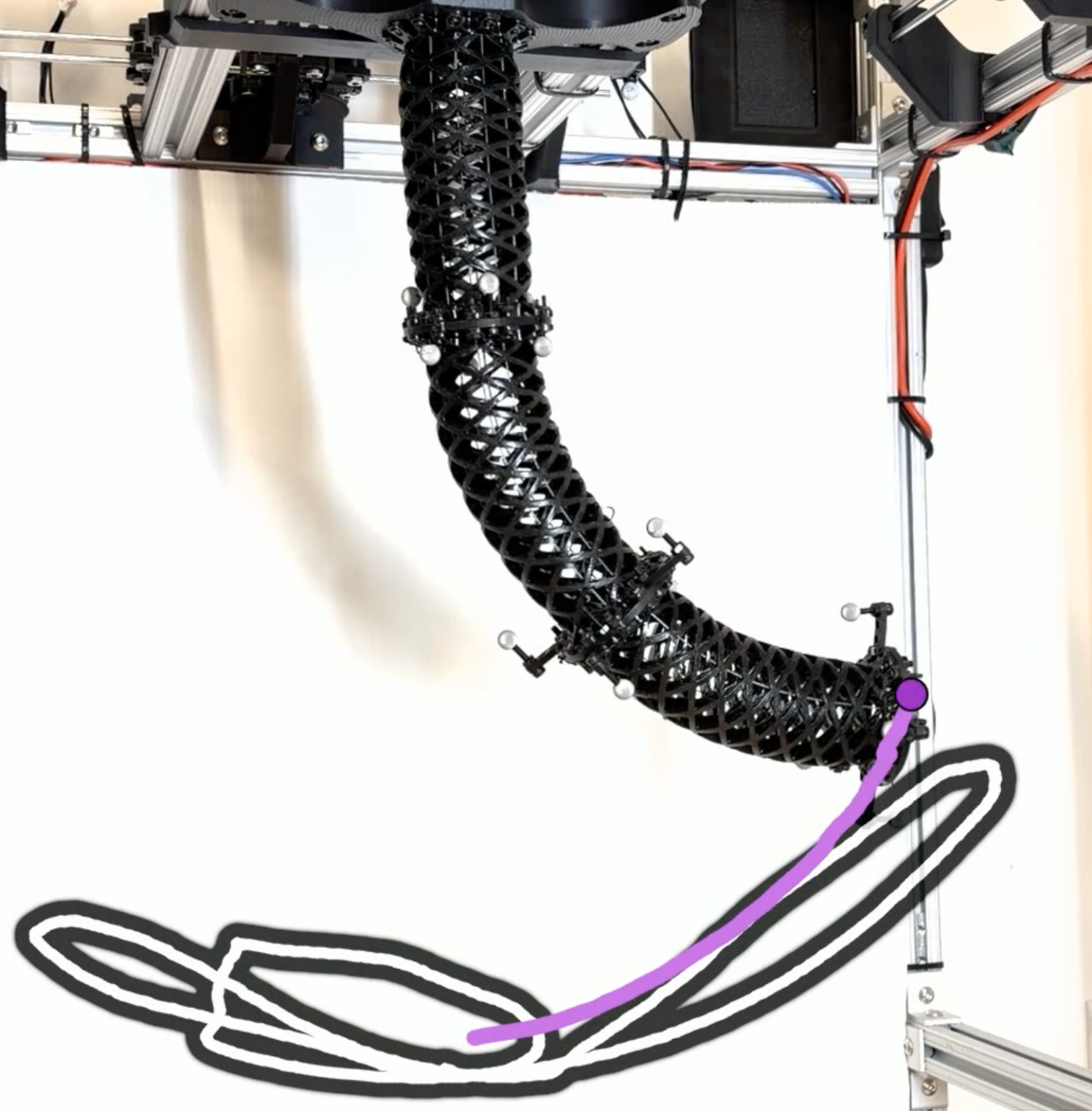}
        \label{fig:hw_dmdc}
    \end{subfigure}

    \begin{subfigure}[b]{0.35\linewidth}
        \centering
        \modeltitle{cKoop}{Koopman-pregain (6D)}
        \includegraphics[width=\linewidth]{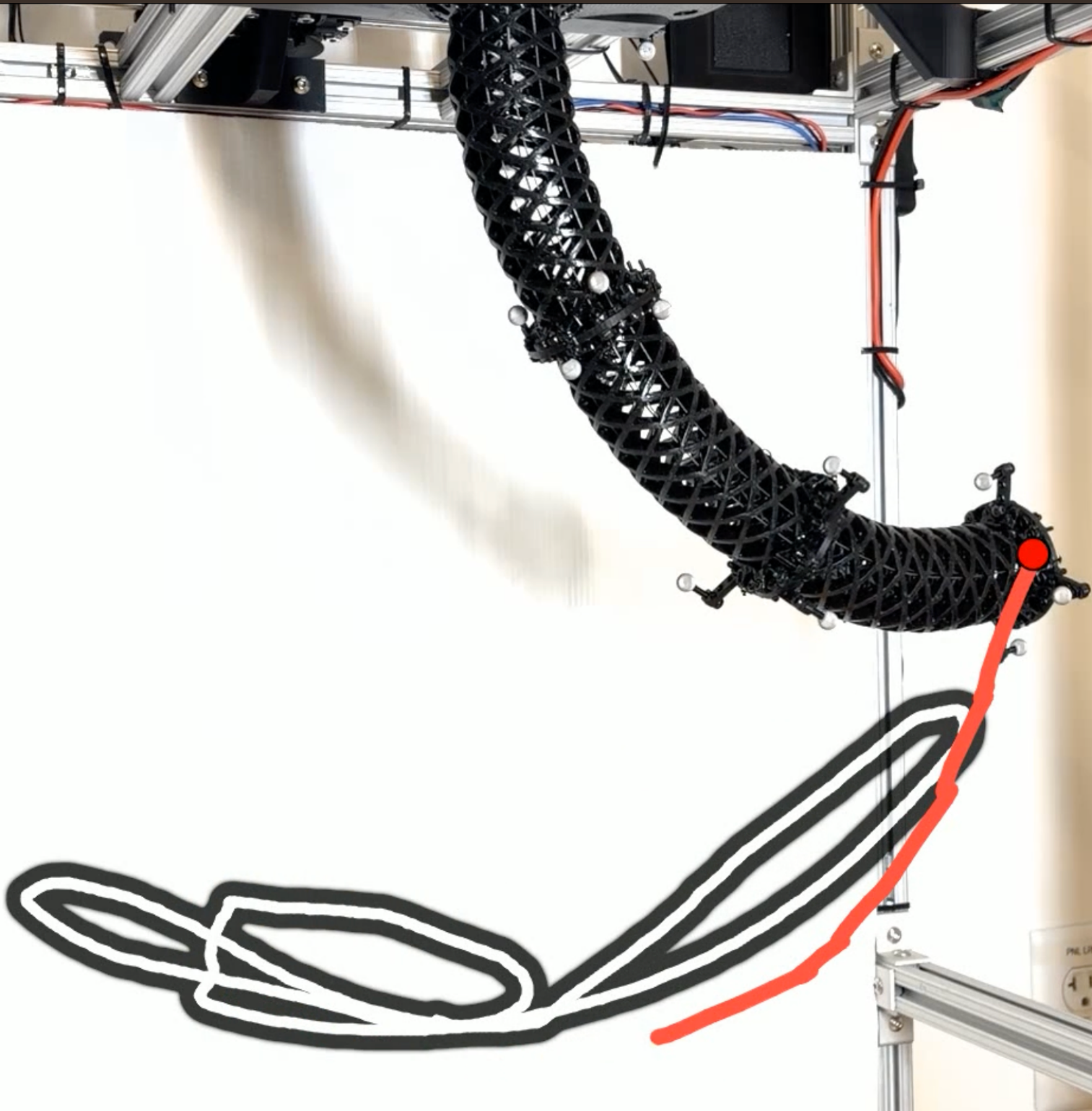}
        \label{fig:hw_koop}
    \end{subfigure}
    \hspace{0.02\linewidth}
    \begin{subfigure}[b]{0.36\linewidth}
        \centering
        \modeltitle{cMZ}{MZ-Koopman (33D)}
        \includegraphics[width=\linewidth]{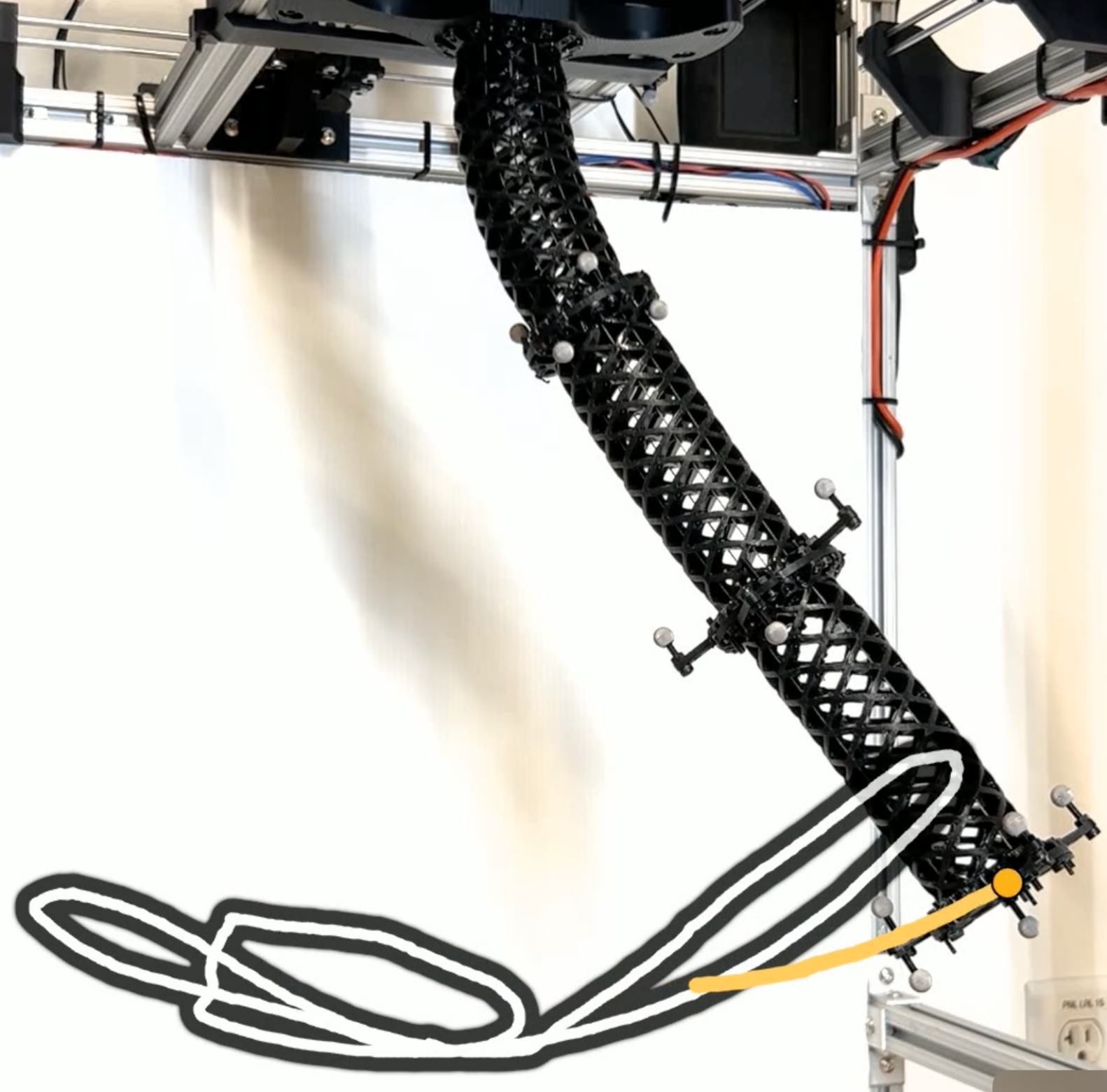}
        \label{fig:hw_mz}
    \end{subfigure}

    \begin{subfigure}[b]{0.35\linewidth}
        \centering
        \modeltitle{cRL}{RL (PPO)}
        \includegraphics[width=\linewidth]{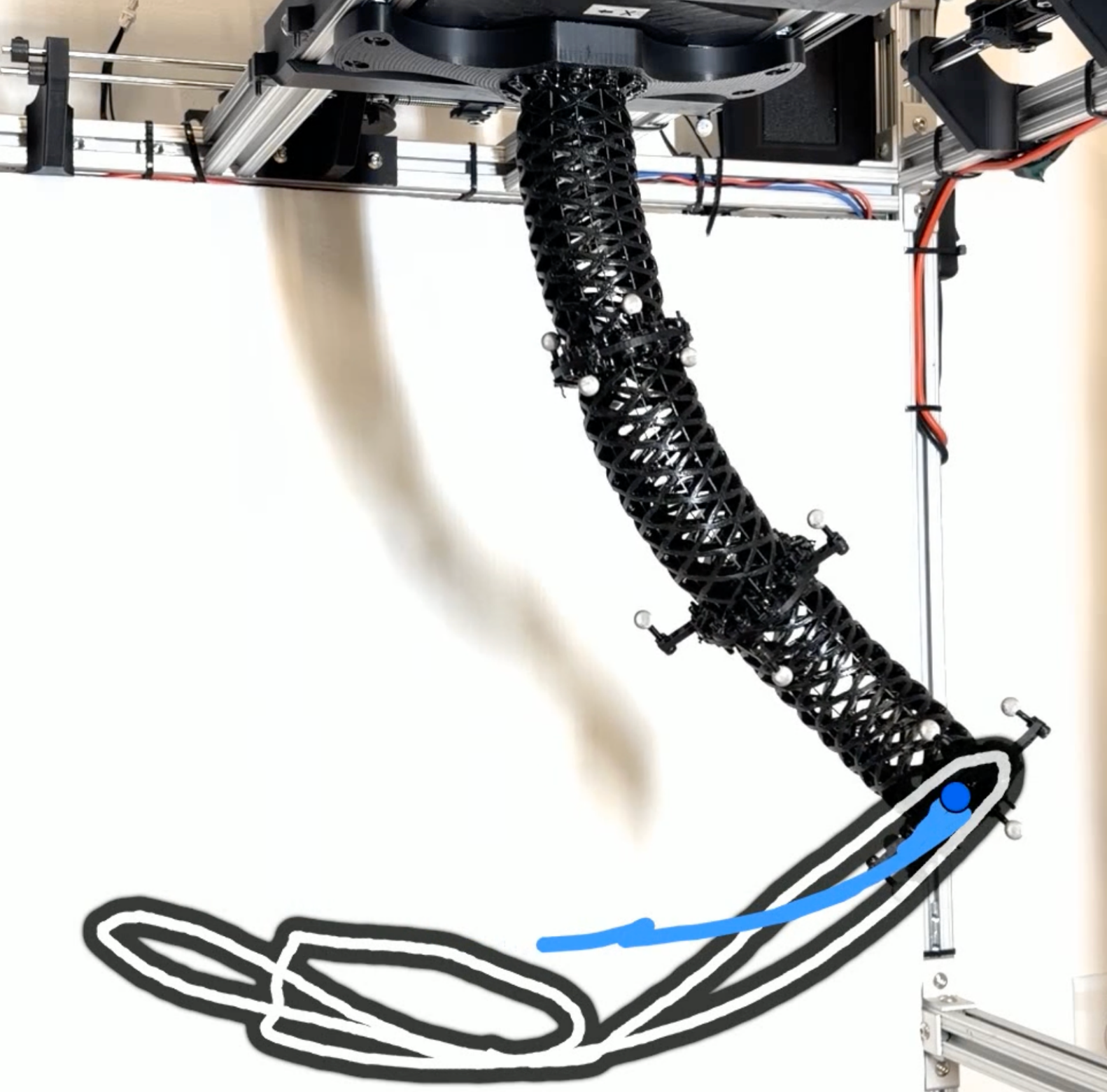}
        \label{fig:hw_ppo}
    \end{subfigure}

    \caption{Hardware snapshots during closed-loop tracking of the $3$D Random track. See Supplementary Movie for all closed-loop hardware demonstrations.}
    \label{fig:hw_cl}
\end{figure}

The closed-loop experimental results are plotted in Fig.~\ref{fig:random_cl} for all the model-based controllers and the RL policy. The 3D tracks require the use of complex nonlinear bending shapes for the trunk robot to achieve low tracking errors. Our aSSM-reduced controller achieves precisely this objective, as shown in the closed-loop prediction plots, and the snapshot of the trunk in Fig.~\ref{fig:hw_cl} validates our expectation. 

The linear methods (Koopman-pregain and DMDc) fail to achieve this objective. The workspace plots indicate sharp deviations from the target track, and the hardware snapshot plots (see Fig.~\ref{fig:hw_cl}) show that both models cause the trunk to bend linearly, sending the trunk's end effector off the intended track and outside the tube of uncertainty. 

The memory-based MZ-Koopman controller suffers from an off-centering issue, as observed in the Figure-8 track. It is a sizeable model, and hence real-time MPC struggles to correct for the errors as we traverse the target track. We attribute the model's off-centered performance to the ad hoc manner in which the observables are chosen. The burden of overfitting of the model's many parameters leads to an incorrect prediction. 

\begin{figure}[t]
    \centering
    \includegraphics[width=0.75\linewidth]{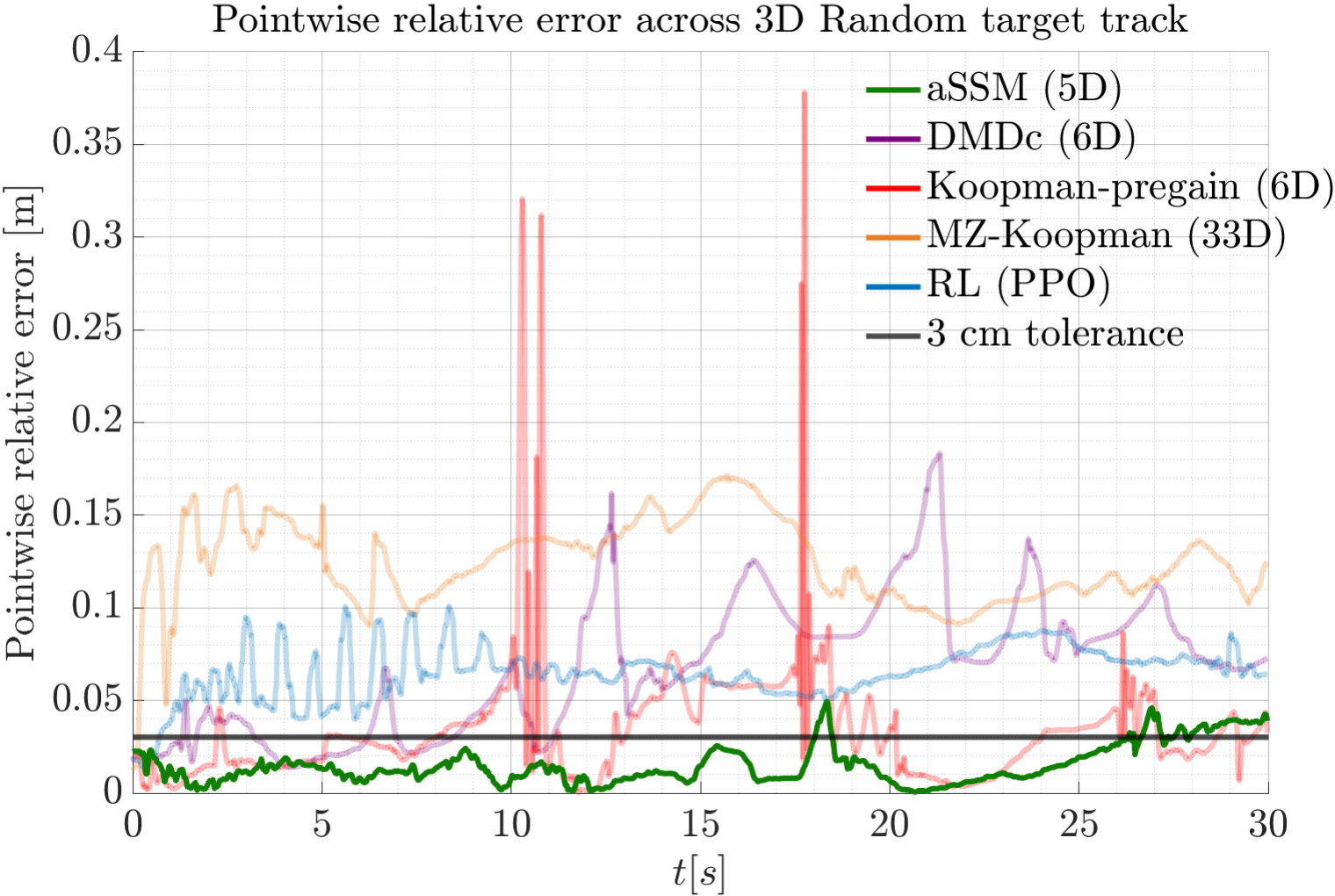}
    \caption{Pointwise relative error for closed-loop tracking of a 3D Random track.}
    \label{fig:rand_error}
\end{figure}

The RL policy has even more parameters than the MZ-Koopman model and hence is susceptible to the same problems. Indeed, the RL policy performs very poorly. Despite longer training on similar random target tracks, it fails to generalize to an unseen random target track. We support all these observations quantitatively in Fig.~\ref{fig:rand_error} with pointwise relative error plots.

\subsubsection{3D Pacman track}\label{subsec:pacman}

We conclude this section with closed-loop tracking for a 3D Pacman track. This is a significantly harder task than the 3D Random track, as it is a apriori unclear if the track is physically feasible for the soft trunk to traverse, and it also contains sharp, rapid jumps. The 3D Pacman track rises to a height of $0.05$ [m] and has a radius of $0.1$ [m]. For a total duration of 30 [s], we perform closed-loop experiments for three different target speeds: (i) mean instantaneous speed $v_{\boldsymbol{\Gamma}} = 3$ [cm/s] with slowness measure $r_s = 0.05$ (ii) twice as fast with slowness measure $r_s = 0.12$  (i) (iii) four times faster than (i) with slowness measure $r_s = 0.2$.

\begin{figure}[H]
    \centering
    \newcommand{\pacpanel}[1]{%
        \includegraphics[trim=0pt 54pt 20pt 90pt, clip, width=\linewidth]{#1}}
    \newcommand{\pacmodel}[2]{\textcolor[RGB]{#1}{\bfseries\small #2}}
    \setlength{\tabcolsep}{0pt}
    \begin{tabular}{@{}p{0.49\linewidth}@{\hspace{0.02\linewidth}}p{0.49\linewidth}@{}}
        \centering$v_\Gamma = 3\ [\mathrm{cm/s}]$ &
        \centering$v_\Gamma = 6\ [\mathrm{cm/s}]$ \tabularnewline[3pt]
        \centering\pacmodel{18,126,4}{aSSM (5D)} &
        \centering\pacmodel{18,126,4}{aSSM (5D)} \tabularnewline[-2pt]
        \pacpanel{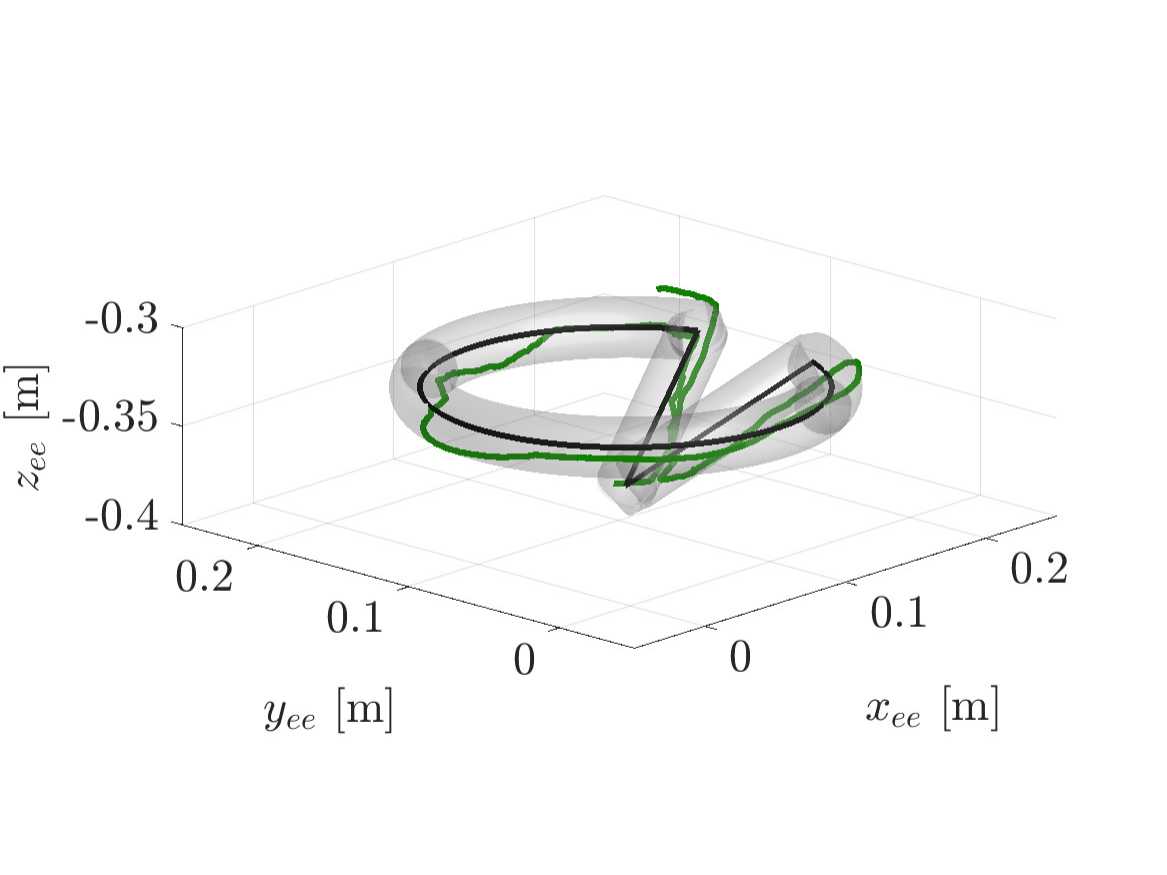} & \pacpanel{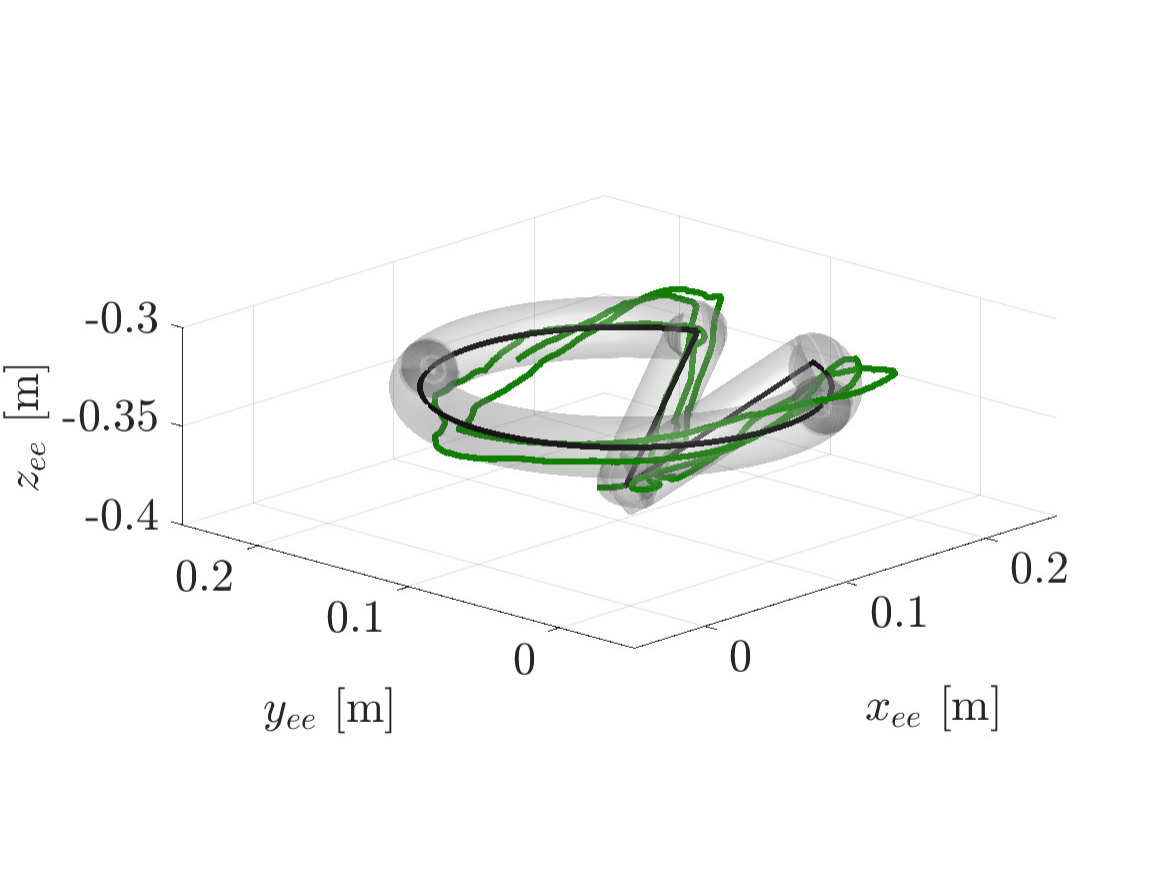} \tabularnewline[2pt]
        \centering\pacmodel{128,0,128}{DMDc (6D)} &
        \centering\pacmodel{128,0,128}{DMDc (6D)} \tabularnewline[-2pt]
        \pacpanel{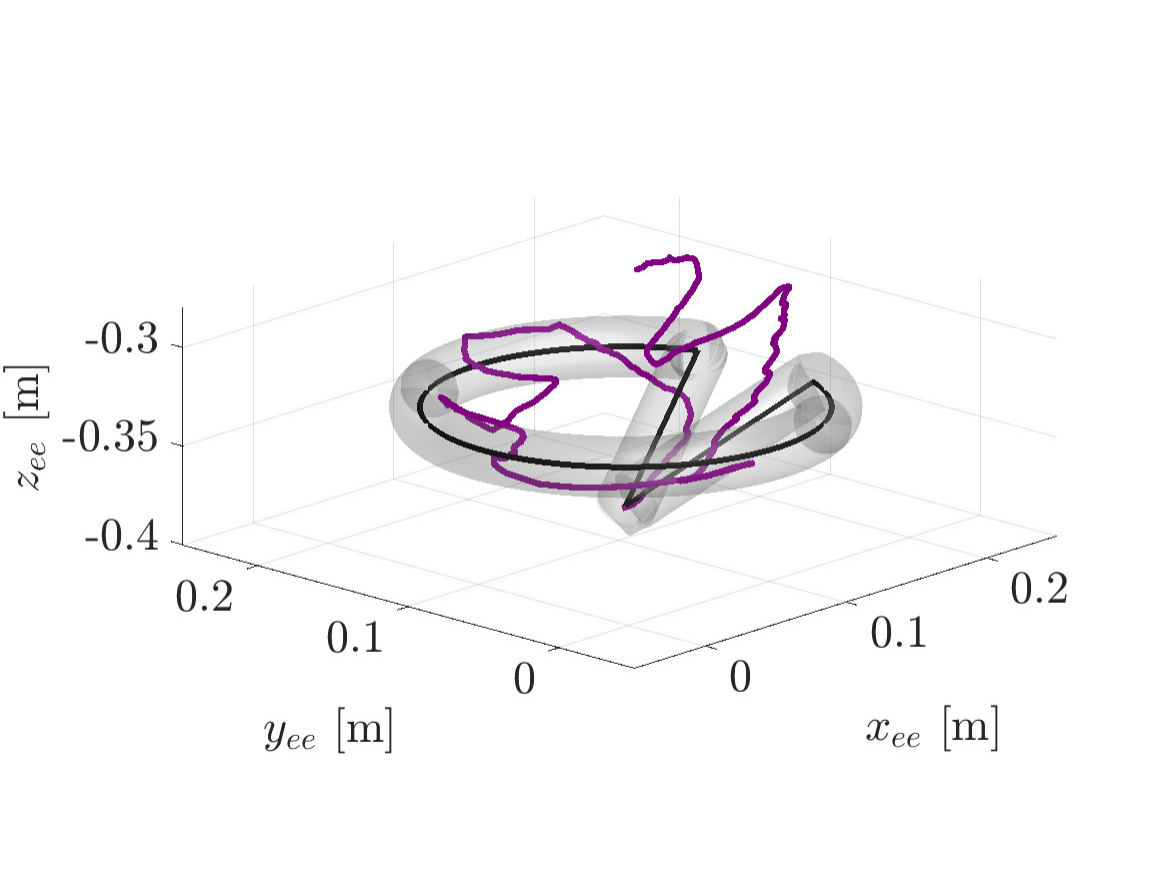} & \pacpanel{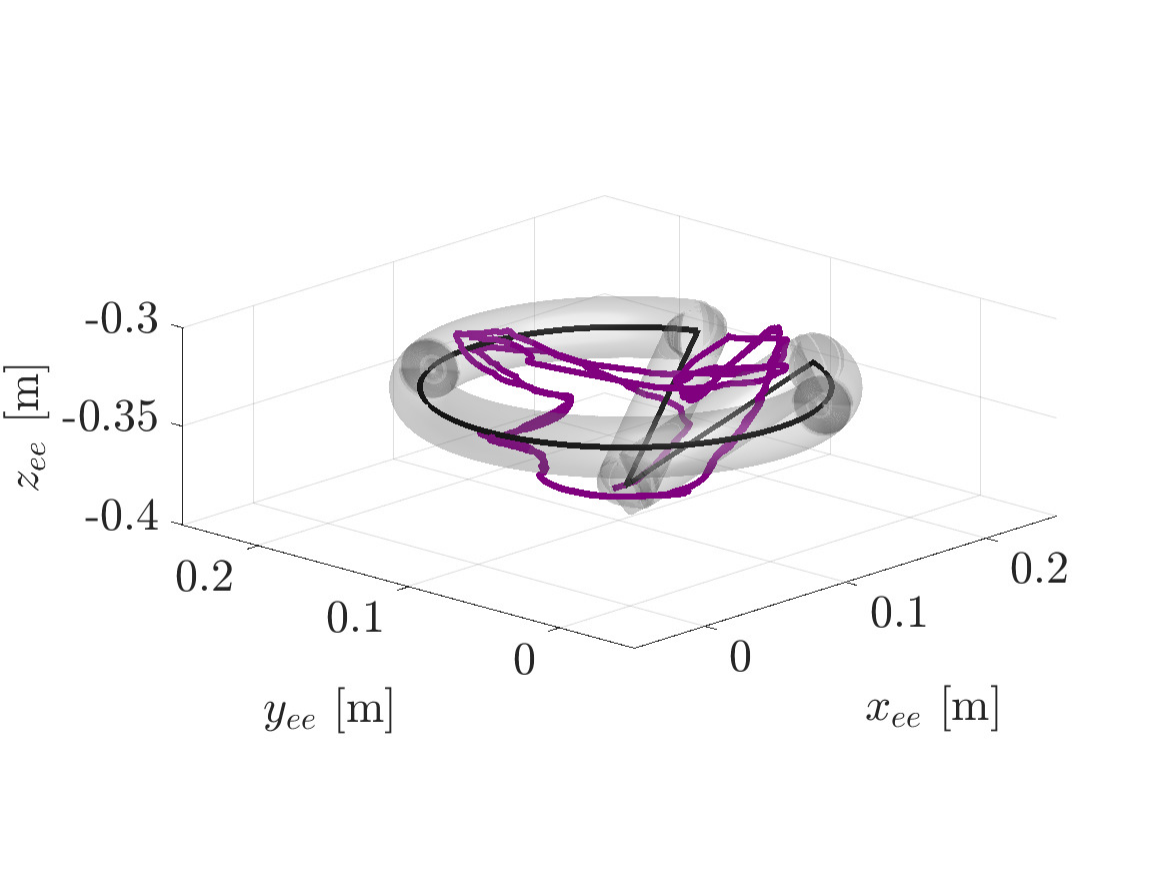} \tabularnewline[2pt]
        \centering\pacmodel{237,28,36}{Koopman-pregain (6D)} &
        \centering\pacmodel{237,28,36}{Koopman-pregain (6D)} \tabularnewline[-2pt]
        \pacpanel{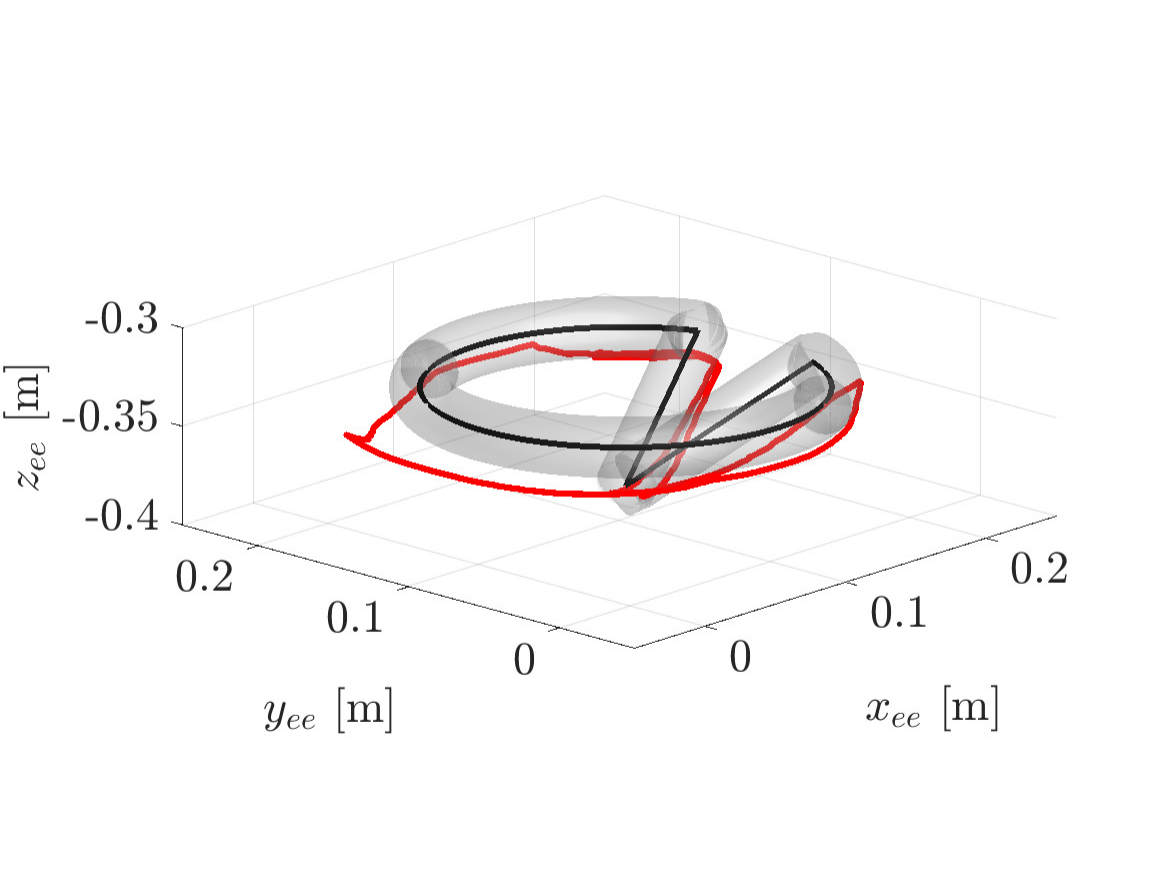} & \pacpanel{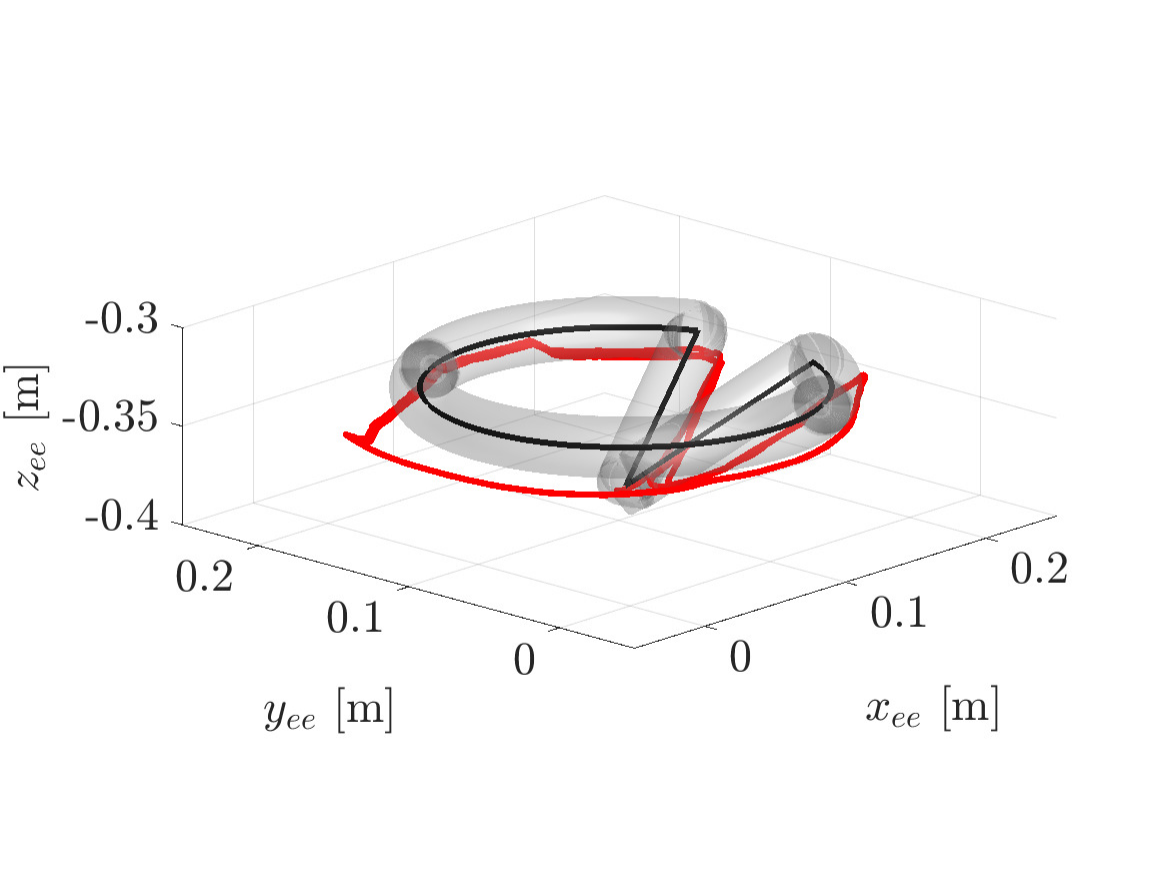} \tabularnewline[2pt]
        \centering\pacmodel{237,125,49}{MZ-Koopman (33D)} &
        \centering\pacmodel{237,125,49}{MZ-Koopman (33D)} \tabularnewline[-2pt]
        \pacpanel{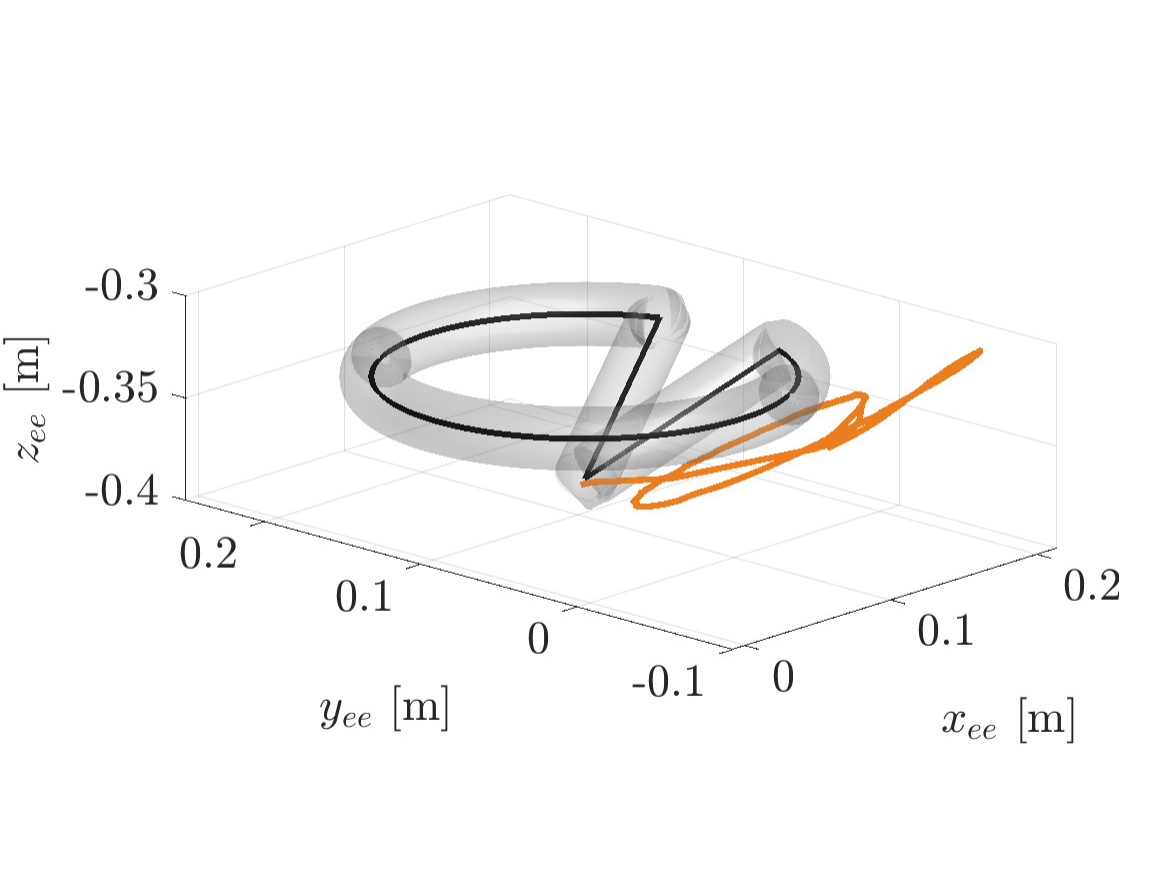} & \pacpanel{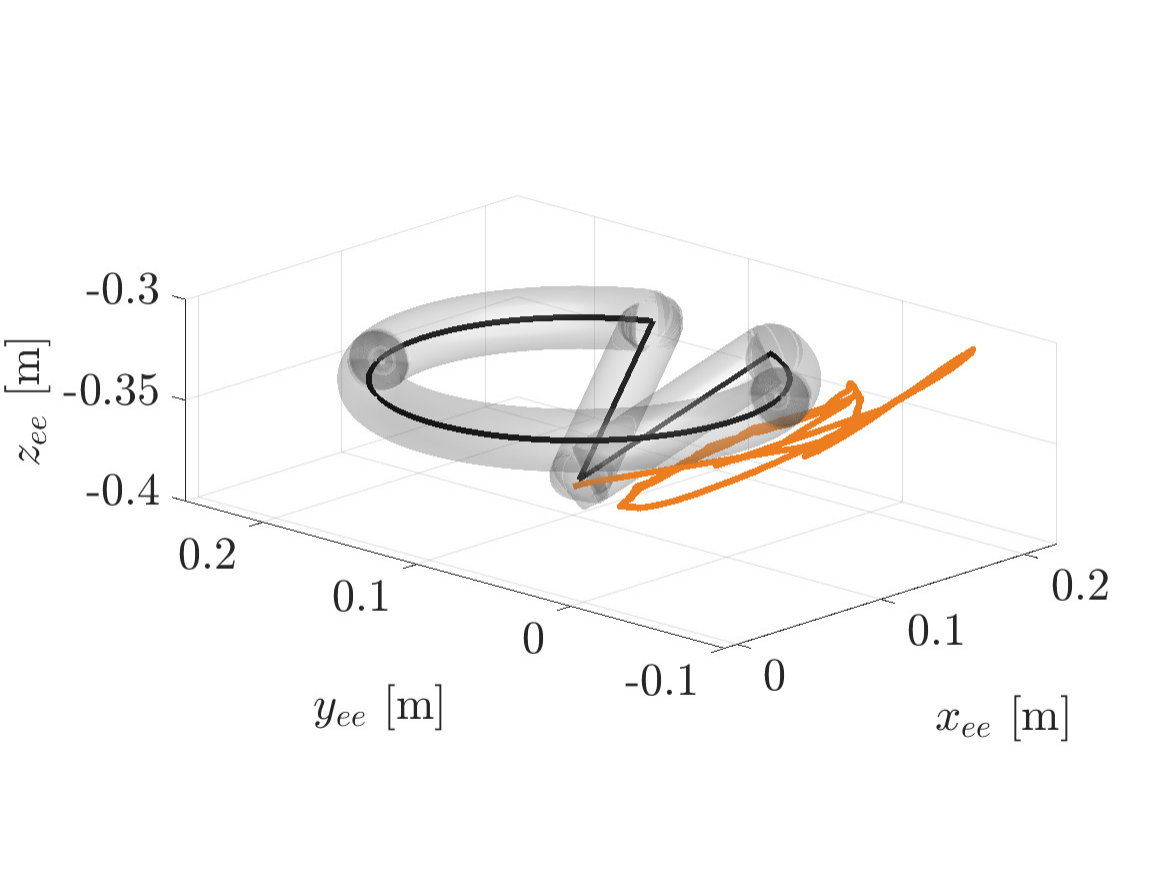} \tabularnewline[2pt]
        \centering\pacmodel{0,114,189}{RL (PPO)} &
        \centering\pacmodel{0,114,189}{RL (PPO)} \tabularnewline[-2pt]
        \pacpanel{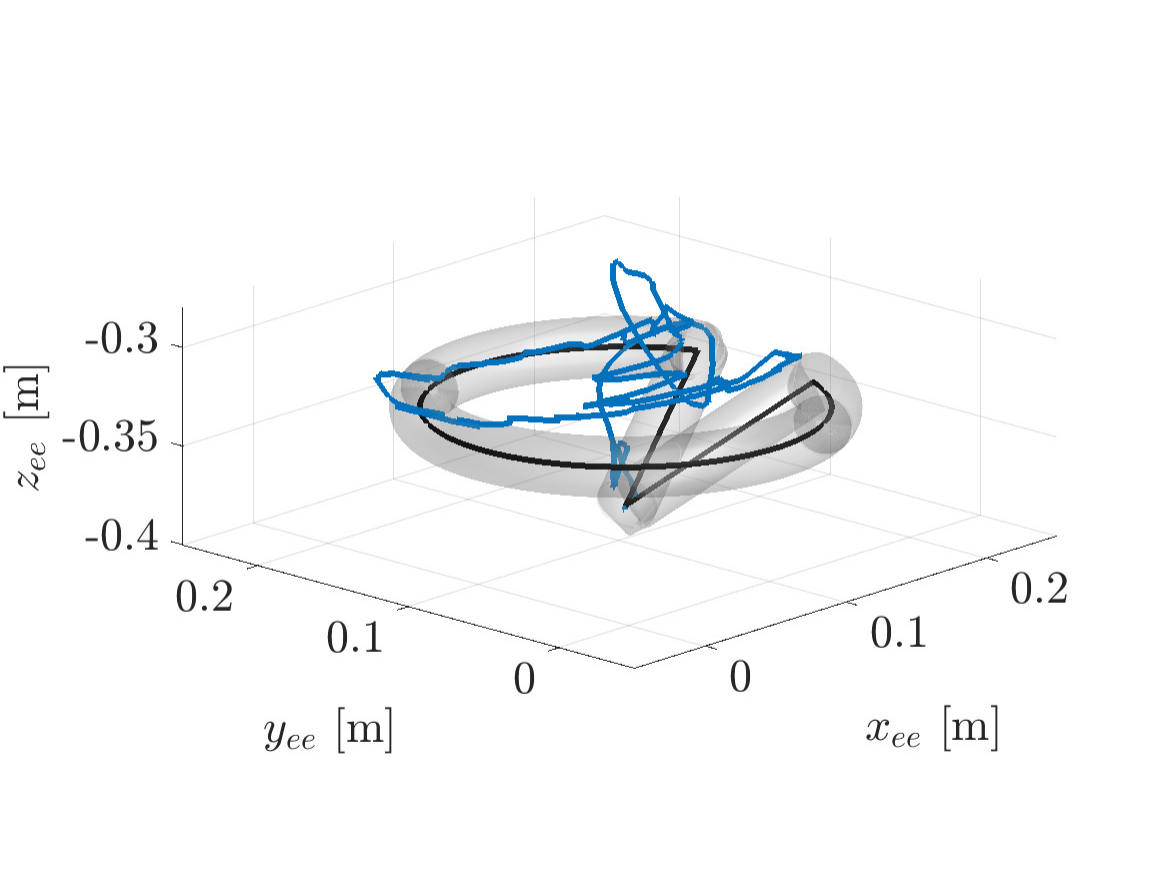} & \pacpanel{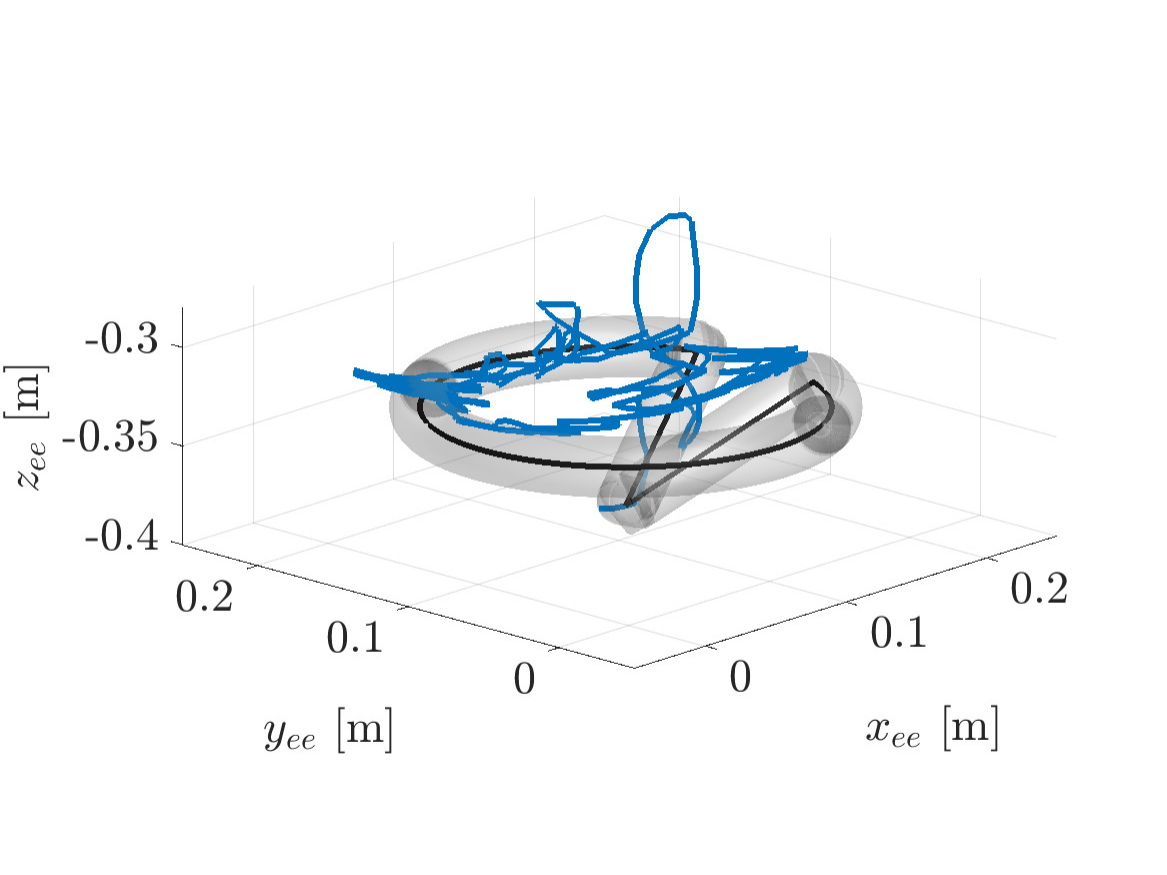} \tabularnewline
    \end{tabular}
    \caption{Closed-loop tracking results for a 3D Pacman track (black) with the uncertainty tube of radius $3$ [cm] plotted in gray. The left panel shows results for the average speed along the target track of $3$ [cm/s], and the right panel shows results for a speed that is 2 times faster.}
    \label{fig:pac3d}
\end{figure}

We plot the closed-loop predictions for all the models in the workspace in Fig.~\ref{fig:pac3d}, with the left panel for target speed case (i) and the right panel for the speed case (ii). Only the aSSM-reduced model achieves the maximal target height and tracks the 3D Pacman across both speeds. The linear methods fail to lift off from the origin state. Either they are completely off from the target or they underestimate the height as the controllers based on them have no nonlinear information to access the large heights in the trunk's workspace. They also linearly extrapolate, which causes them to stray outside the tolerance tube. 

\begin{figure}[H]
    \centering
    \includegraphics[width=0.7\linewidth]{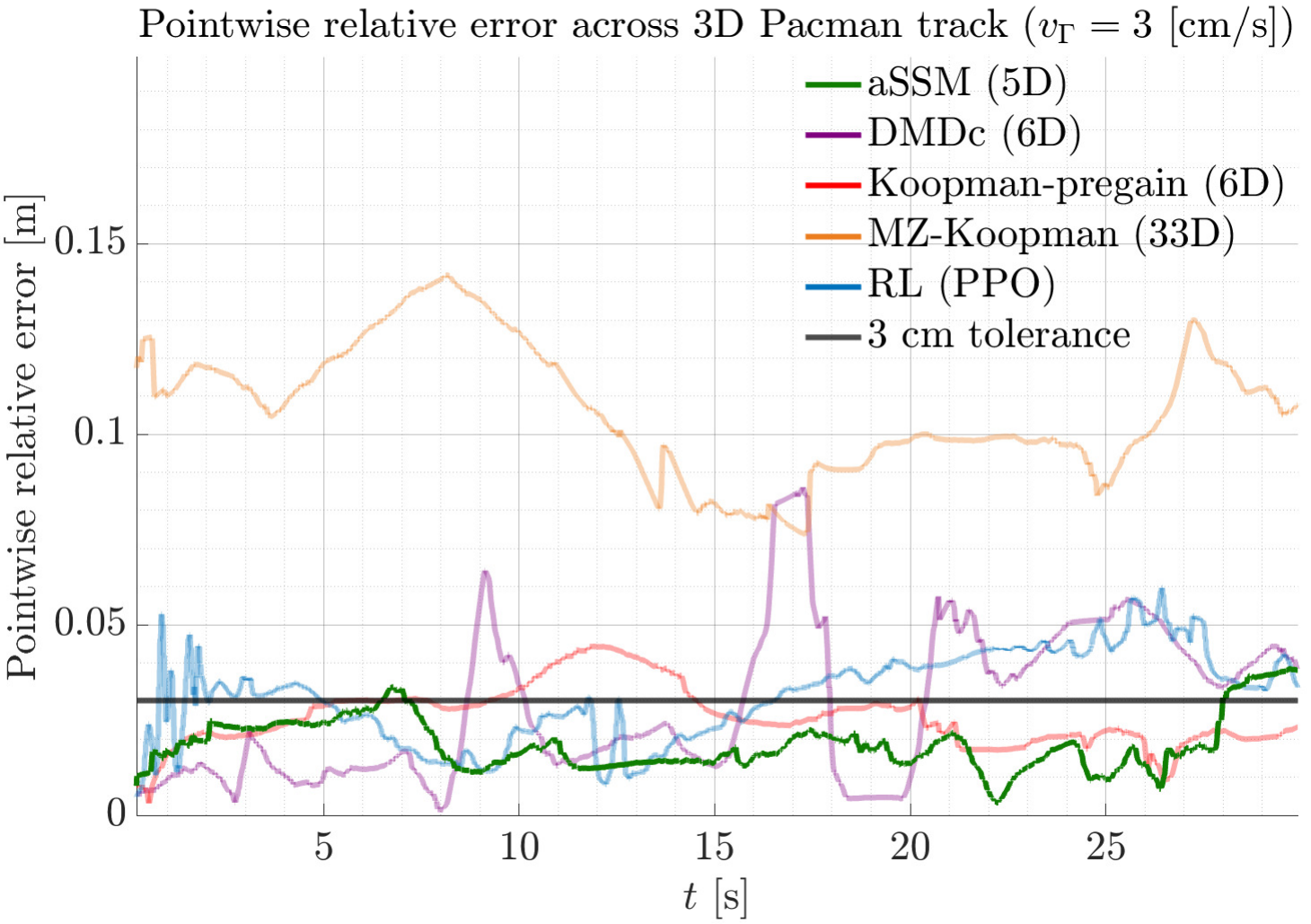}
    \caption{Pointwise relative error for closed-loop tracking of a 3D Pacman track with speed $3$ [cm/s].}
    \label{fig:p3D_slow}
\end{figure}

\begin{figure}[H]
    \centering
    \includegraphics[width=0.7\linewidth]{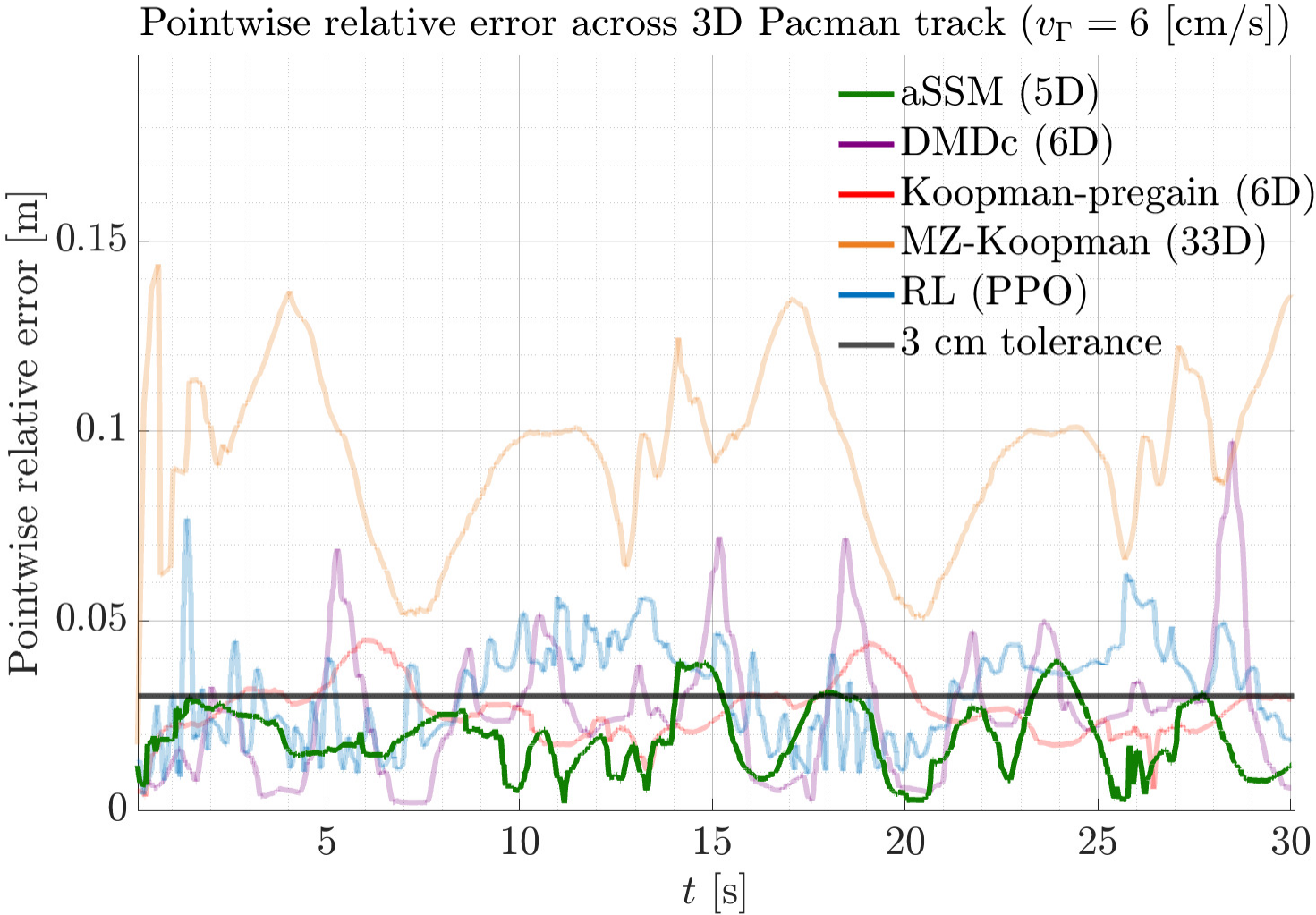}
    \caption{Pointwise relative error for closed-loop tracking of a faster 3D Pacman track.}
    \label{fig:p3D_fast}
\end{figure}

The MZ-Koopman model provides similar closed-loop predictions as before, but in this problem, it is unable to reproduce even the qualitative shape of the 3D Pacman trajectory. The RL policy at best reaches the necessary height, but it is unable to trace out the track and starts oscillating rapidly as it tries to align with the track. These qualitative visual assessments of the closed-loop result are corroborated with pointwise relative error plots in Fig.~\ref{fig:p3D_slow} for case (i) and in Fig.~\ref{fig:p3D_fast} for case (ii).

\FloatBarrier

In the Supplementary Section \ref{app:pacman_fast}, we present the results for case (iii) with error plots. Despite the track being extremely fast, the aSSM-reduced model still provides consistent performance with lower errors across the target track than the other baselines.

\subsection{Closed-loop results for obstacle avoidance}\label{subsec:constraints}
    
We conclude the results section 1 with closed-loop experiments conducted in the presence of static and dynamic obstacles within the trunk's observable space. Here, we will set additional constraints for the MPC problem defined in eq.~(\ref{eq:aSSM_mpc0}). For the static obstacle task, we fix the target $\boldsymbol{\Gamma}(t)$ to be the 3D Random target track used in Section \ref{subsec:random}. For the dynamic obstacle tasks, we focus only on avoiding a moving obstacle; hence, we set the target-tracking costs to zero, $\mathbf{Q_z} = \text{diag}(0,0,0)$. In all obstacle-avoidance tracking problems, we only record the center points of the trunk’s end-effector and mid-section positions. To safely account for the trunk’s radial width when avoiding three-dimensional objects, we therefore add an effective buffer length of 
$r_{\text{eff}} = 0.25$ cm.

We also note that none of our obstacles are real physical objects. They are virtual spatial constraints designed by the user. However, we still expect our system to work seamlessly when the constraints are physical, provided we have enough motion capture cameras to accurately observe the trunk's end effector and the obstacle's position. At present, we do not have the camera system to carry out this task. Hence, we resort to using virtual obstacles for a proof of concept.

 \subsubsection{Avoiding static spherical obstacles}\label{subsec:sphere}

We add static spherical constraints to the MPC by enforcing the condition, 
\begin{equation}
\|\mathbf{z}-\mathbf{c}\|\geq r+r_{\text{eff}} 
\end{equation}
for the trunk's end effector under closed-loop tracking. We pick centers $c$ for the sphere locations as random nearby points along the target track. This will ensure that the track is non-trivially obstructed our goal is to track the target as precisely as possible while strictly avoiding the spherical obstacles. We denote the radius of the spherical obstacle as $r$ and always plot it in dark red. We also plot a red translucent spherical shell of width $r_{\text{eff}}$ centered at $\mathbf{c}$.

\begin{figure}[H]
    \centering
    \begin{subfigure}[b]{0.8\linewidth}
        \centering
        \includegraphics[width=\linewidth]{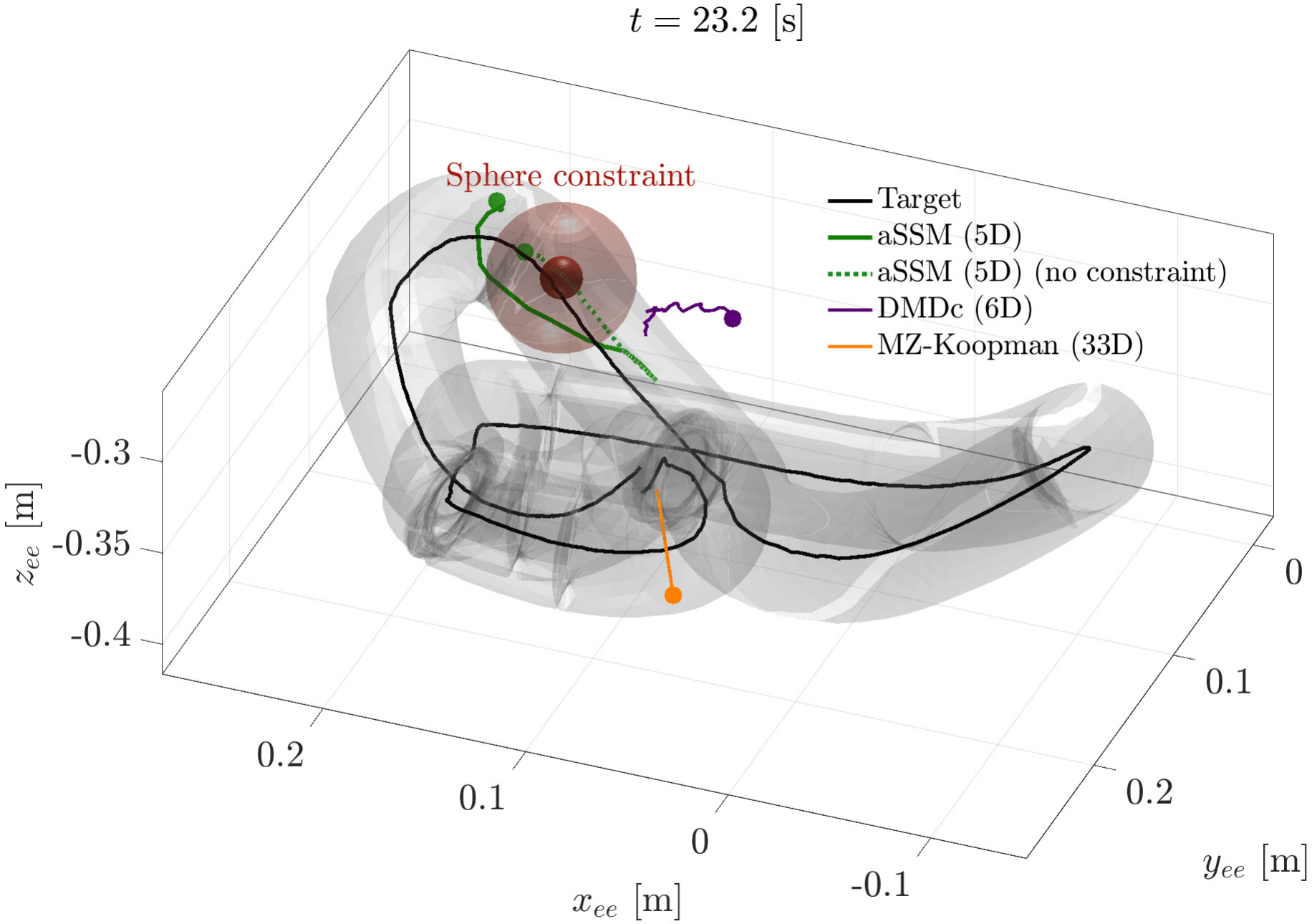}
        \caption{}
        \label{fig:small_obstacle}
    \end{subfigure}
    \hfill
    \begin{subfigure}[b]{0.8\linewidth}
        \centering
        \includegraphics[width=\linewidth]{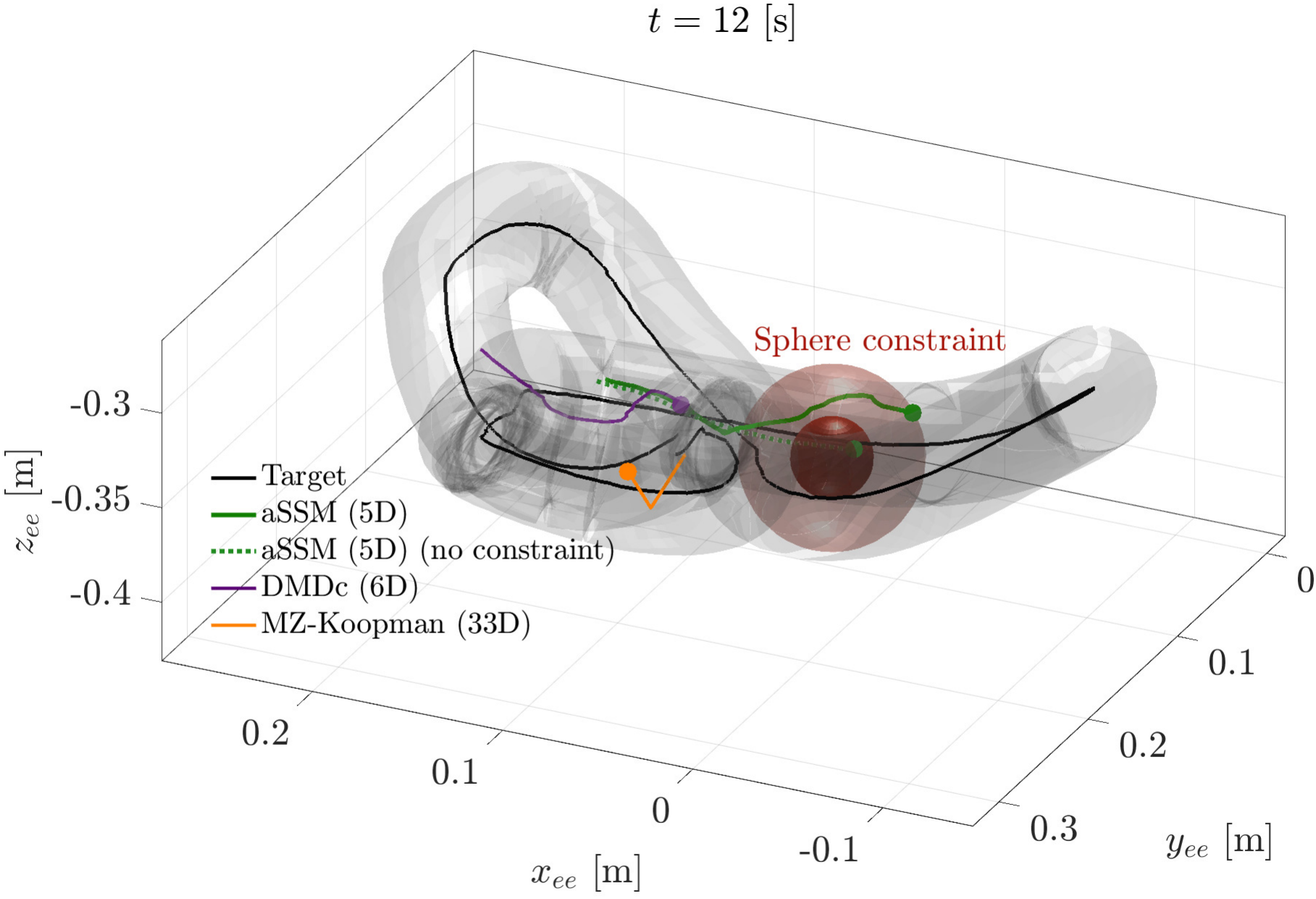}
        \caption{}
        \label{fig:medium_obstacle}
    \end{subfigure}
    \caption{Snapshots for closed-loop avoidance of static spherical obstacles while simultaneously tracking a 3D Random track for (a) a small-sized spherical obstacle, and (b) a medium-sized spherical obstacle. Each plot contains closed-loop results for the 5D aSSM-reduced (green solid line), the 6D DMDc model (purple), the 33D MZ-Koopman model (yellow), and the 5D aSSM-reduced with deactivated constraints (green dotted line). The uncertainty tube of radius $3$ [cm] is plotted in gray around the target track. The dark red sphere is the obstacle, and the translucent sphere shell surrounding it is a safety buffer zone with a width of $2.5$ [cm].}
    \label{fig:constrain_1}
\end{figure}

\begin{figure}[t!]
    \centering
    \begin{subfigure}[b]{0.8\linewidth}
        \centering
        \includegraphics[width=\linewidth]{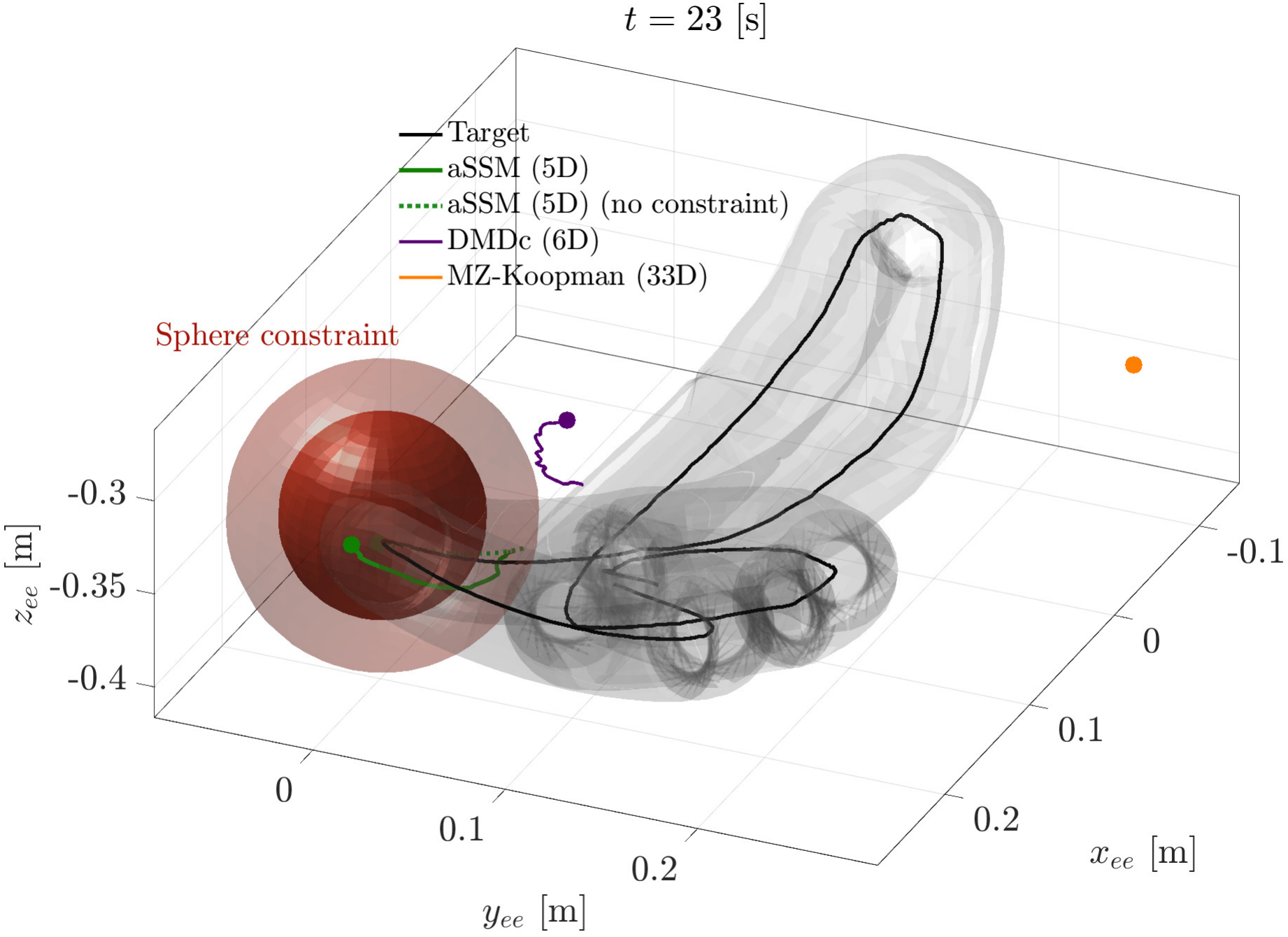}
        \caption{}
        \label{fig:large_obstacle}
    \end{subfigure}
    \hfill
    \begin{subfigure}[b]{\linewidth}
        \centering
        \includegraphics[width=\linewidth]{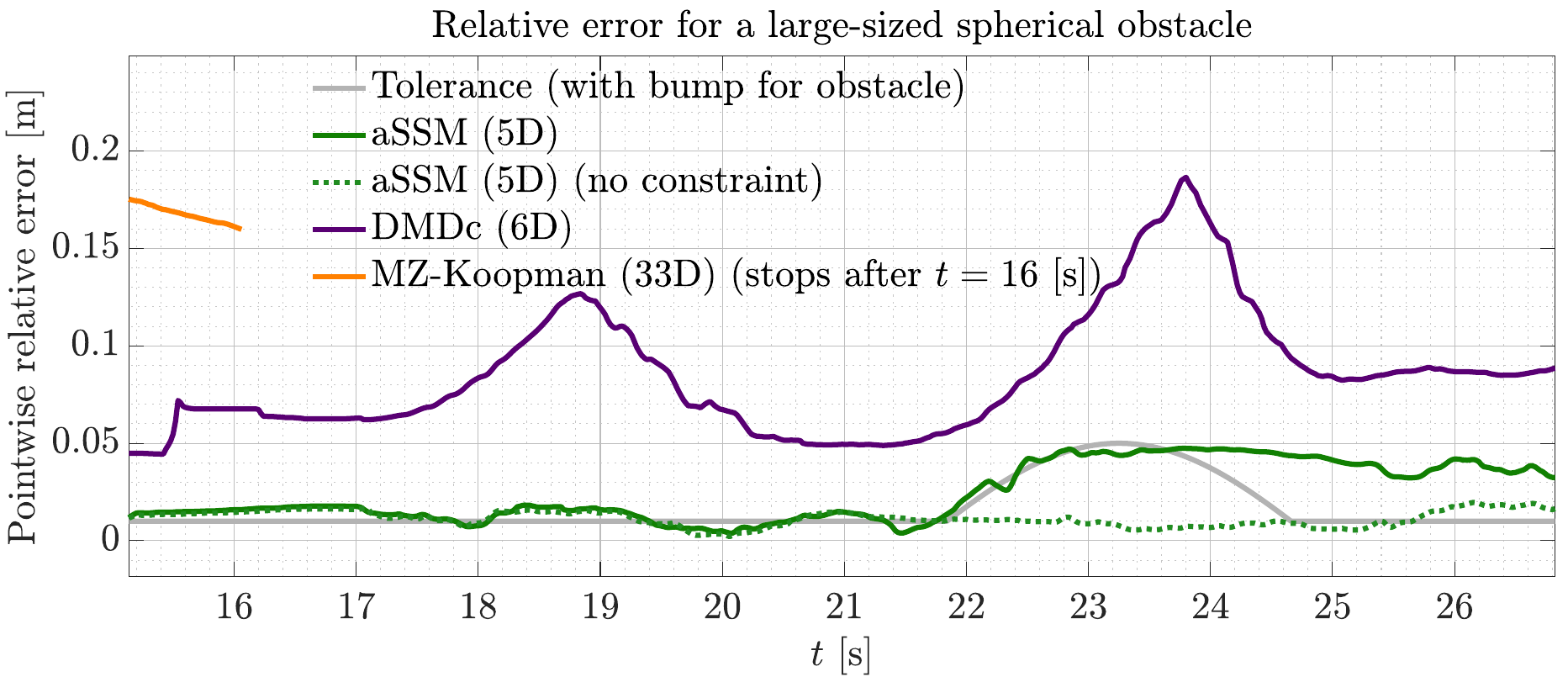}
        \caption{}
        \label{fig:rpe_obstacle_error}
    \end{subfigure}
    \caption{(a) Snapshot for avoiding a large spherical obstacle while tracking the 3D Random target. (b) Pointwise relative error plots for the large-sized spherical obstacle avoidance task. The tolerance line accounts for the sphere's location in real time as the target track is traversed.}
    \label{fig:constraints_2}
\end{figure}

We demonstrated closed-loop experiments for three spherical obstacles of different sizes: (i) a small obstacle of radius $r=0.05$ [cm], (ii) a medium obstacle of radius $r=0.1$ [cm], and (iii) a large obstacle of radius $r=0.2$. All these spherical obstacles are placed at random locations close to the 3D Random track. We plot temporal snapshots in which the trunk must avoid the spherical obstacle. See Fig.~\ref{fig:small_obstacle} for case (i), see Fig.~\ref{fig:medium_obstacle} for case (ii), and Fig.~\ref{fig:large_obstacle} for case (iii). 

We compare our aSSM-reduced model with the constraint condition switched off, to ensure that the avoidance is a non-trivial occurrence. Across all figures, the aSSM-reduced controller without constraints passes through the obstacle, whereas with constraints active, the model's closed-loop predictions steer the trunk's end effector away from it. For the small- and medium-sized spherical obstacles, the aSSM-reduced controller avoids the entire spherical constraint, including the buffer zone. The large spherical constraint is more challenging. In that task, the aSSM-reduced controller steers the end-effector away from the dark red spherical obstacle but is still trapped in the buffer region.

We also compare these results with those from the DMDc and MZ-Koopman models, the two benchmarks that theoretically allow for constraint handling. The DMDc model operates at a different height, avoiding the spherical obstacles, but ultimately misses the equally important task of tracking the target. The MZ-Koopman model's predictions across all tasks consistently blow up, leading to the deployment of the controller's safety codes to preserve the trunk hardware. The reason for this poor performance is that a 33D model with constraints is not fast enough to run in real time and is also not accurate, leading to poor guesses by the optimization solver and, consequently, longer solve times.

We collected episodes to train RL policies for almost 8 hours, but the method did not converge and caused the trunk's tendons to snap, resulting in extensive downtime for maintenance. For these reasons, we have not been able to construct an effective RL control policy to handle constraints. In Fig.~\ref{fig:rpe_obstacle_error}, we provide error plots that quantitatively assess whether we avoid the large spherical constraints for all models. The error plots for the small- and large- sized obstacles can be found in Supplementary Section \ref{app:sphere_appendix}.

\FloatBarrier

\subsubsection{Avoiding dynamic cylindrical obstacles}\label{subsec:cylinder}

Our final closed-loop experiment involves the trunk avoiding a moving cylindrical obstacle. We present results only for our aSSM-reduced model. The MZ-Koopman model has longer solve times and cannot handle a complex moving obstacle, causing the controller to stop. In contrast, the linear DMDc closed-loop controller produces erratic initial predictions that violate the moving constraint, leading the controller to stop abruptly, as shown in the previous section.

\begin{figure}[H]
    \centering
    \begin{subfigure}[b]{\linewidth}
        \centering
        \includegraphics[width=0.8\linewidth]{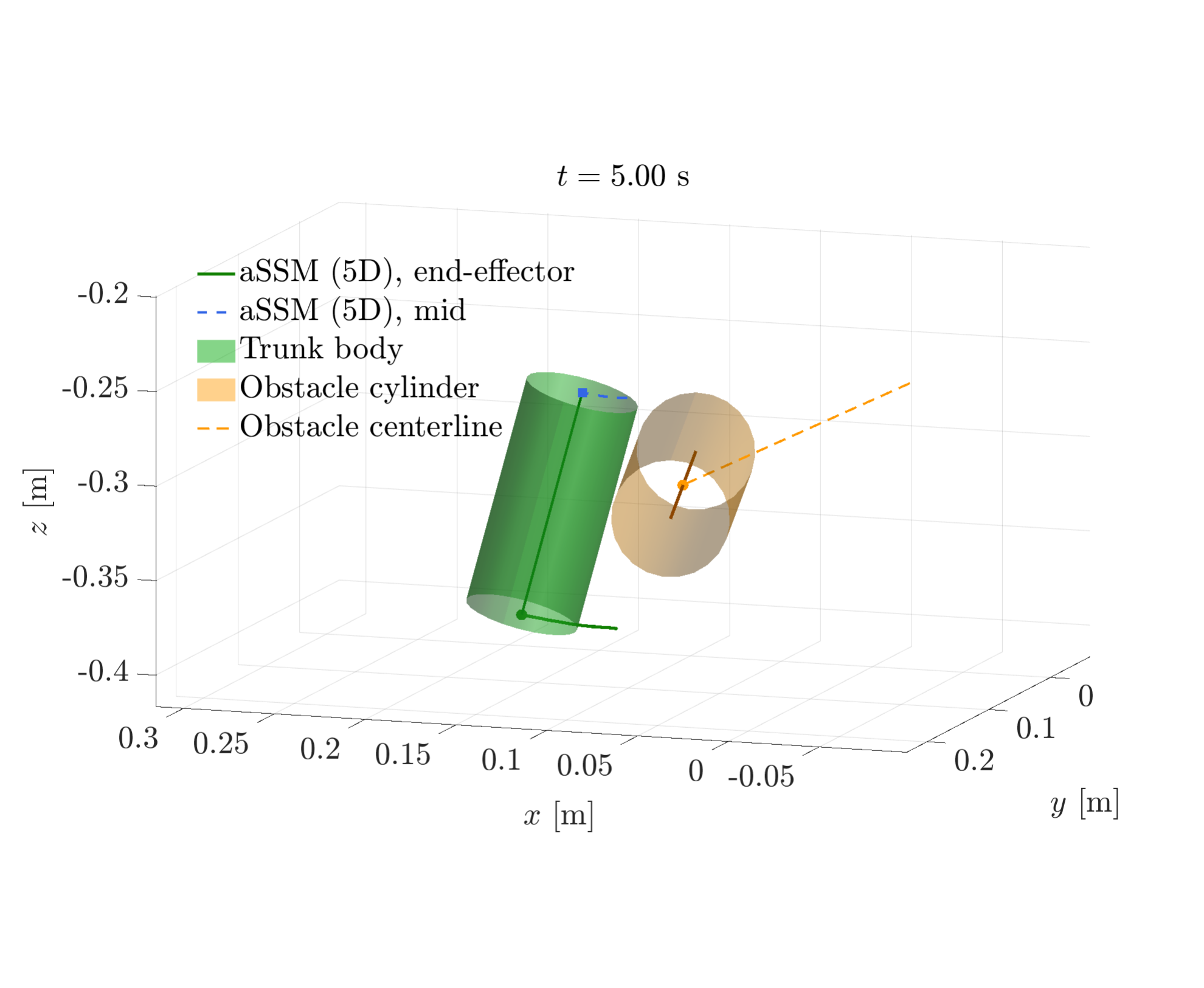}
        \caption{}
        \label{fig:cyl_snap1}
    \end{subfigure}
    \hfill
    \begin{subfigure}[b]{\linewidth}
        \centering
        \includegraphics[width=0.8\linewidth]{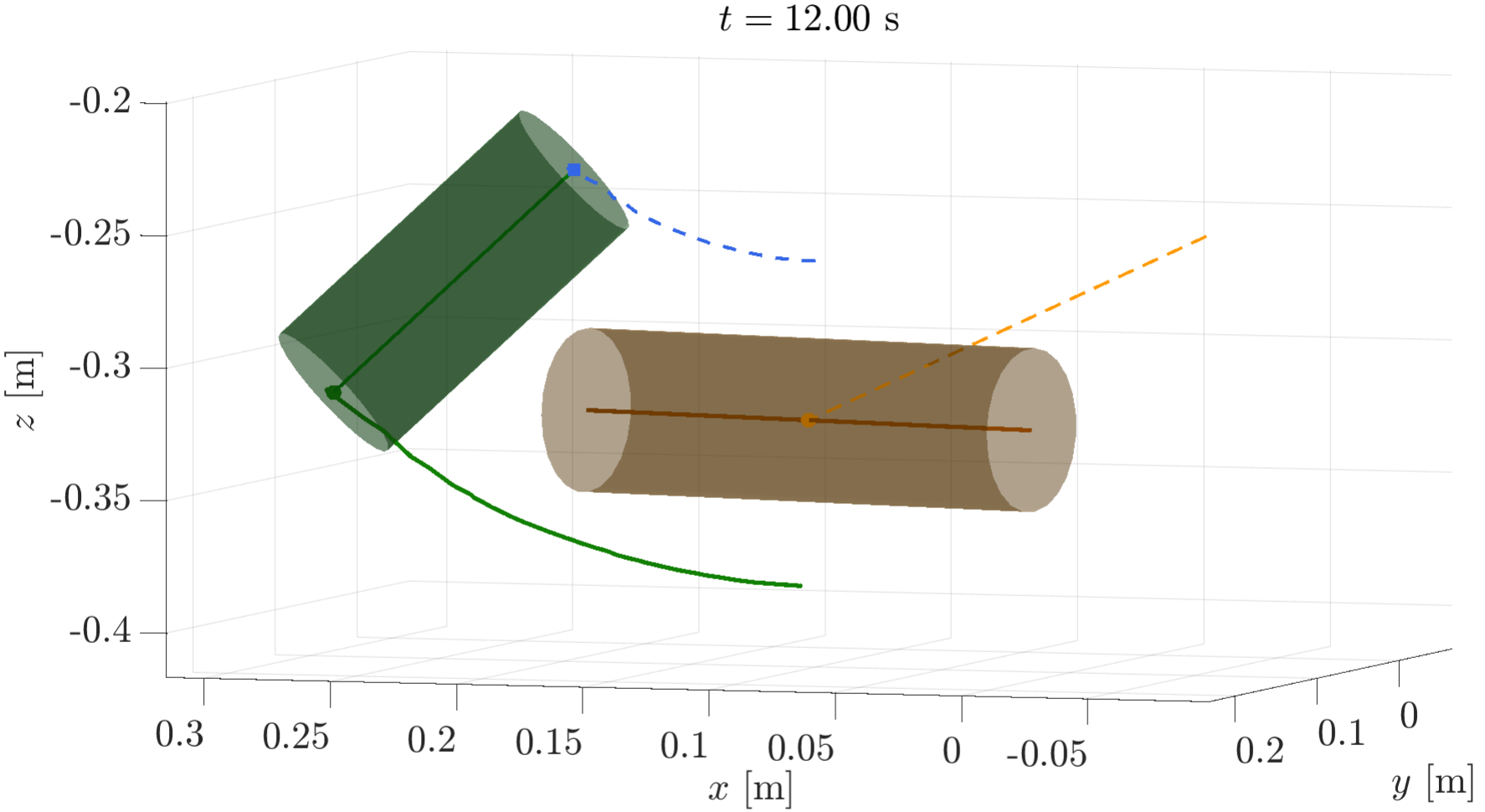}
        \caption{}
        \label{fig:cyl_snap2}
    \end{subfigure}
    \caption{Snapshots of the trunk main body avoiding a moving cylindrical obstacle at (a) $t=5$ [s] as it approaches the rest state and (b) $t=15$ [s] as the obstacle rotates near the rest state via aSSM-based closed-loop control. The trunk body is green, and the cylindrical obstacle is yellow.}
    \label{fig:constrain_3}
\end{figure}

Here, we will need to use the full trunk geometry to avoid the obstacle. The trunk's observable space allows us to approximate its geometry between the mid-position and the end-effector as a cylinder with radius $r_{\text{eff}} = 0.25$ [cm]. We use this cylinder approximation of the trunk in the observable space to set up a constraint that avoids a moving cylindrical obstacle, given by, 
\begin{equation}
\min_{s,\,u \,\in\, [0,1]} \left\| ((1-s)\,\mathbf{z}_{\text{mid}}(t) + s\,\mathbf{z}(t)) - \Big(\mathbf{c}(t) + \left(u-\tfrac{1}{2}\right) l\,\mathbf{d}(t)\Big) \right\| \;\geq\; r_{\text{eff}} + r.
\end{equation}
Here the observable space is split into mid and end-effector positions $\mathbf{y} =\{\mathbf{z},\mathbf{z}_{\text{mid}}\}$; $l$ and $r$, denote the length and the radius of the cylindrical obstacle, $\mathbf{c}(t)$ is the moving centerline of the obstacle, and $\mathbf{d}(t)$ is the orientation of the cylinder axis.

\FloatBarrier
\begin{figure}[H]
    \centering
    \begin{subfigure}[b]{0.8\linewidth}
        \centering
        \includegraphics[width=\linewidth]{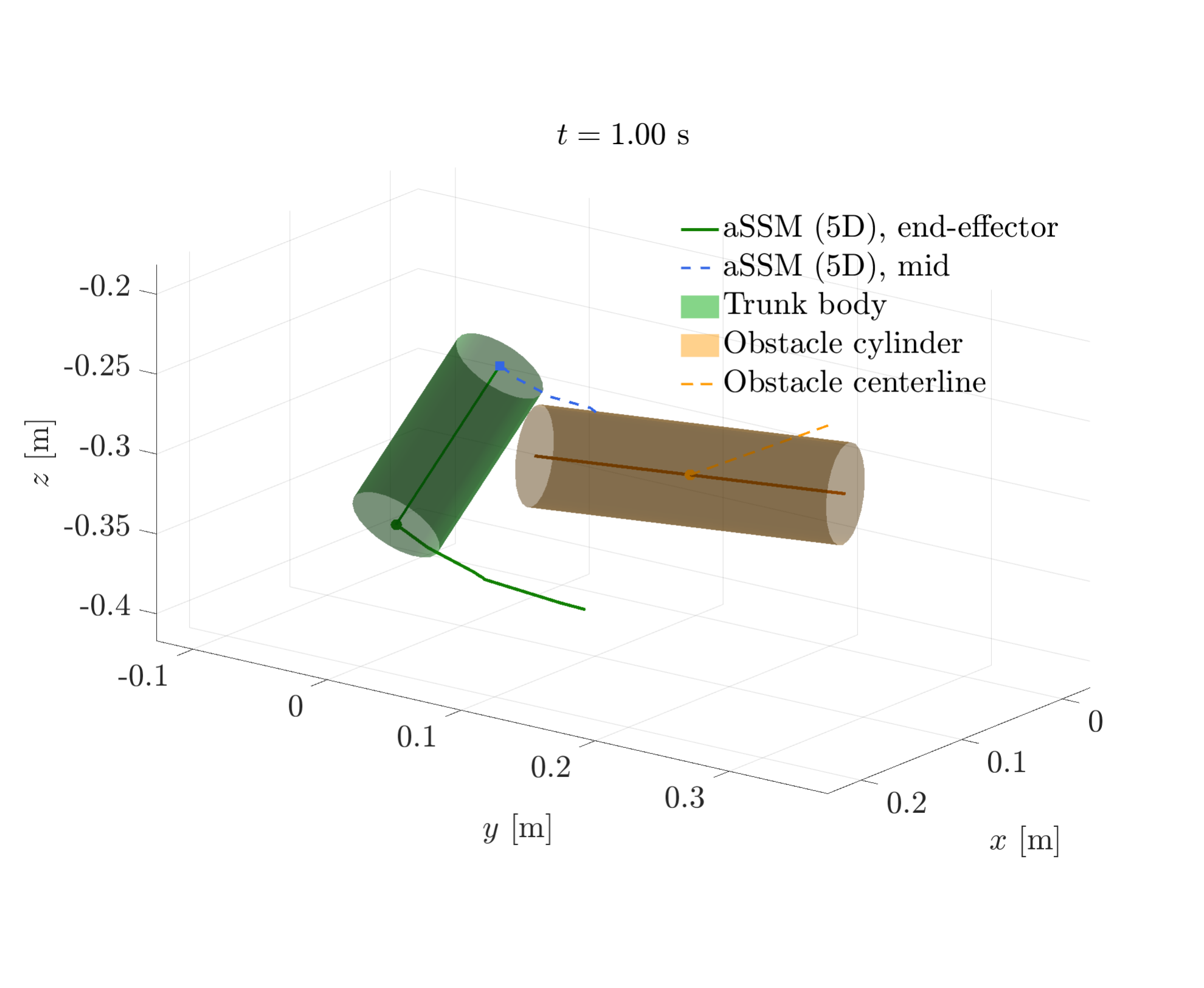}
        \caption{}
        \label{fig:fast_cyl_1}
    \end{subfigure}
    \begin{subfigure}[b]{0.8\linewidth}
        \centering
        \includegraphics[width=\linewidth]{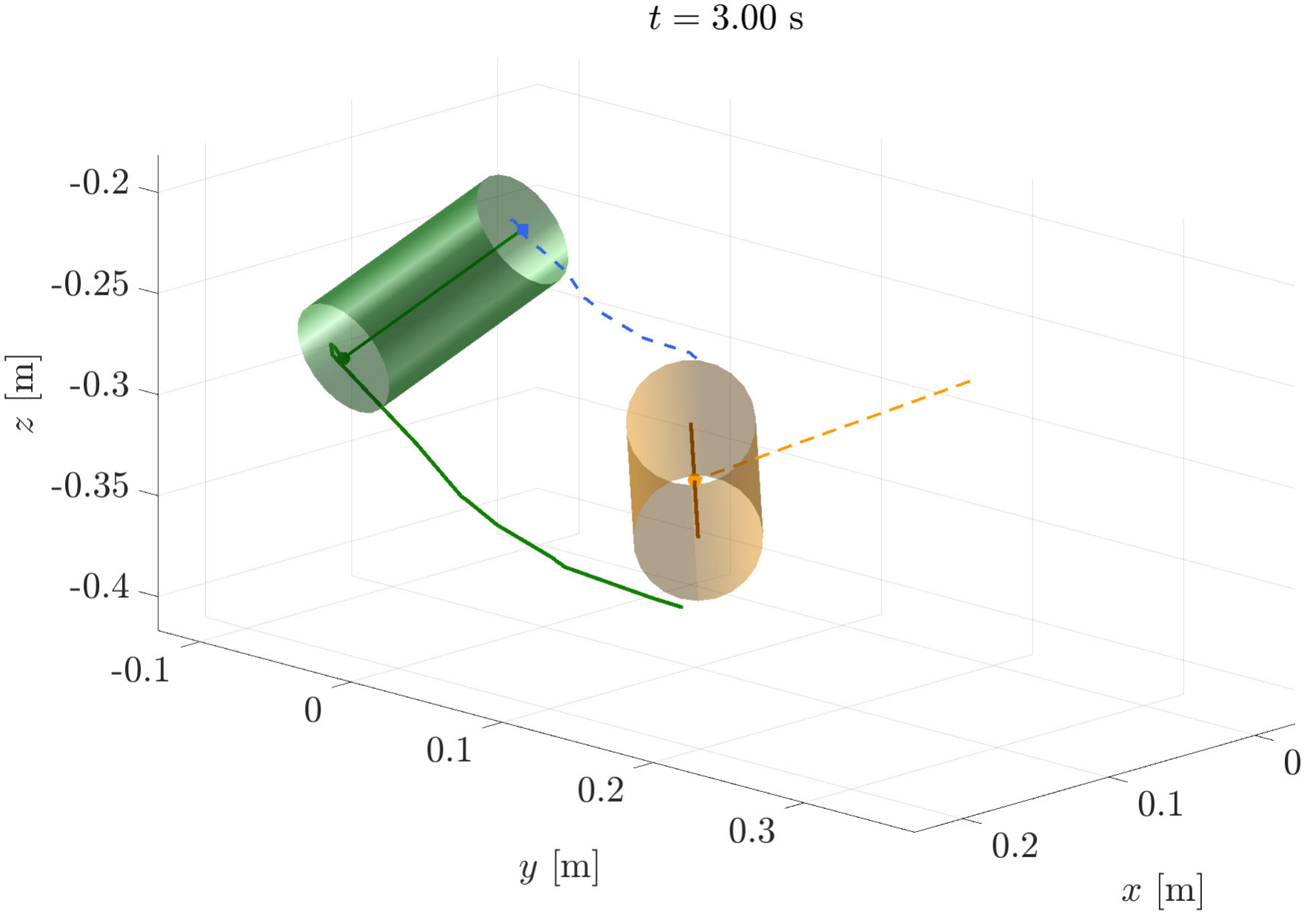}
        \caption{}
        \label{fig:fast_cyl_2}
    \end{subfigure}
    \caption{Snapshots at (a) $t=1$ [s] and (b) $t=3$ [s] for the same closed-loop avoidance task same as Fig.~(\ref{fig:constrain_3}), with a faster moving cylindrical obstacle.}
    \label{fig:constraints_4}
\end{figure}

In Fig.~\ref{fig:constrain_3}, we plot snapshots of the trunk avoiding the moving cylindrical obstacle under aSSM-based closed-loop control. We set the length of the cylindrical obstacle to $ l = 0.25$ [m] and its radius to $r = 0.031$ [m]. The centerline path $\mathbf c(t)$ moves the cylindrical obstacle inwards towards the trunk's rest position from a height of $0.2$ [m]. The cylindrical obstacle reaches the trunk's rest position in $8$ [s], after which it rotates its orientation axes $\mathbf d(t)$ clockwise in the $x_{ee}-y_{ee}$ plane for the next $10$ [s]. We show that the trunk body (plotted in green) avoids the cylindrical obstacle during dynamic variation along the centerline path (see Fig.~\ref{fig:cyl_snap1}) and dynamic changes in cylinder orientation (see Fig.~\ref{fig:cyl_snap2}).

Finally, in Fig.~\ref{fig:constraints_4}, we plot the trunk body avoiding the cylindrical obstacle for a dramatically faster case of the task described before. We see that in a span of $3$ [s], the trunk quickly moves away from the fast-approaching cylindrical object. We also describe in Supplementary Section \ref{app:cylinder_complex} a closed-loop result for a more complex centerline path.

\section{Discussion} \label{sec:discussion}

We have developed a hardware-compatible, data-driven control method for soft robotic systems based on the theory of adiabatic spectral submanifolds (aSSMs) presented in Section~\ref{subsec:aSSM_overview}. We first discussed and verified the existence of an inherent timescale separation in typical operational use of soft robots. This separation allows for reducing the soft robot dynamics onto low-dimensional attracting aSSMs. Specifically, in Section~\ref{subsec:data-driven}, we presented a methodology to extract aSSMs from limited observational data. We validated this data-driven method on a hardware soft helicoid trunk robot by learning a 5D aSSM-reduced model only from camera observation data of the trunk’s end-effector and mid positions (see Section \ref{subsec:modeling}). We then constructed a general-purpose controller using the learned aSSM-reduced model to perform dynamic target tracking and obstacle avoidance (see Section \ref{subsec:mpc}). Our modeling and control method identifies dominant dynamics of the robot without making assumptions about its underlying physics. As a consequence, our methodology transfers directly across morphologies and actuation systems.

Our aSSM-based controller was then stress-tested on a variety of trajectory tracking control tasks ranging in spatial complexity and speed. We also included systematic comparisons with other available linear baselines (DMDc and Koopman-pregain), a recent memory-based baseline (MZ-Koopman), and a routinely used reinforcement learning (RL) based control policy. In this case study, we documented that the aSSM-based controller provided consistent precision performance compared to the other baselines that failed to generalize (see Section \ref{subsec:closed-loop}). We further tested the generalizability of our aSSM-based controller for obstacle-avoidance tasks (see Section \ref{subsec:constraints}). Across both static and dynamic obstacles, our aSSM-based controller successfully avoided the obstacles while simultaneously tracking other user-defined objectives. 

The performance quality of the aSSM-based controller is achieved with only a modest training data exploration of the trunk’s workspace to learn the aSSM-reduced model. These requirements are practically realizable compared to online training of RL control policies directly on hardware. While the linear methods tested here are data-efficient, they end up extrapolating into unsafe regions, ultimately degrading precision. Thus, our aSSM-based controller meets practical trade-off requirements for hardware operation (see Table~\ref{tab:baseline_comparison}). 

On the practical side, future work will include testing the aSSM-based controller with real physical obstacles and performing interactive demos that demonstrate the accuracy of shape control for the soft robot. We also expect to demonstrate our method on different soft robotic actuation systems (beyond tendon-driven), such as embodied actuation systems with tactile sensing or pressure-driven systems. On the theory side, developing an extension that systematically learns higher-order corrections to the aSSM-reduced model from data and explores their use in control is highly desirable. Another remaining objective is to extend aSSM-reduced modeling to account for external forcing that commonly occurs during environmental interactions. Such interactions include soft robotic hands grasping a ball or a trunk end-effector carrying payloads.

\section{Methods}\label{sec:methods}

\subsection{Adiabatic spectral submanifolds}\label{subsec:setup_theory}

Soft robots are infinite-dimensional dynamical systems satisfying partial differential equations (PDEs). Available equation-based reduced-order modeling methods approximate the solutions of these PDEs using finite elements or Cosserat rods. We assume here that the controlled soft robot dynamics can be represented by a high-dimensional, first-order dynamical system, 
\begin{equation}
\label{eq:eom}
\dot{\mathbf{x}} = \mathbf{F} (\mathbf{x}, \mathbf{u}(t)),
\end{equation}
where $\mathbf{x}\in \mathbb{R}^n$, $\mathbf{u} \in \mathbb{R}^{n_u}$ are the phase- and actuation- space of the soft robot, and $\mathbf{F}$ is a smooth nonlinear function that governs the geometric and material deformation rate of the soft robot. 

For static control inputs $\mathbf{u}(t)\equiv \mathbf{u}_0$, the soft robot will deform into static configurations. We denote this family of static states by $\mathbf{x}_0(\mathbf{u}_0)$. These states are fixed points of the system \ref{eq:eom} and hence satisfy
\begin{equation}
\label{eq:cm_condition}
\mathbf{F}(\mathbf{x}_0(\mathbf{u}_0),\mathbf{u}_0) = \mathbf{0}.
\end{equation}
We further assume that for all static control inputs $\mathbf u_0$ in a compact neighborhood $\mathcal{U} \subset \mathbb{R}^{n_u}$, the corresponding static configurations are stable hyperbolic fixed points of the dynamical system (\ref{eq:eom}). This implies in that in the autonomous limit $\mathbf{u}(t)\equiv \text{const.}$, system (\ref{eq:eom}) has a normally attracting critical manifold $\mathcal{L}_0$ (see \citet{fenichel73}), 
\begin{equation}
\mathcal{L}_0 = \{(\mathbf{x},\mathbf{u}_0) \in \mathbb{R}^{n} \times \mathcal{U} \quad | \quad \mathbf{x}=\mathbf{x}_0(\mathbf{u}_0) \}.
\end{equation}
The spectrum of each stable fixed point of the critical manifold, denoted by $\text{Spec}\left(\mathbf{D}_{\mathbf{x}}\mathbf{F}(\mathbf{x}_0(\mathbf{u}_0), \mathbf{u}_0)\right) = \{\lambda_\nu(\mathbf u_0)\}_{\nu=0}^{n}$, is further assumed to have a strictly nonzero spectral gap at $\nu = d$ for all $\mathbf{u}_0 \in \mathcal{U}$, i.e.,
\begin{equation}
\text{Re}(\lambda_n(\mathbf u_0)) \leq \dots \leq \text{Re}(\lambda_{d+1}(\mathbf u_0)) < \text{Re}(\lambda_{d}(\mathbf u_0))\leq \dots \leq \text{Re}(\lambda_{1}(\mathbf u_0)) < 0.
\end{equation}
This allows us to define the slow spectral subbundle $E(\mathbf{u}_0)$ associated to the eigespaces of the $d$ slowest eigenvalues, 
\begin{equation}
\label{eq:slow_subbundle}
E(\mathbf{u}_0) = \bigoplus_{\nu = 1}^{d} E_\nu(\mathbf{u}_0).
\end{equation}
If further non-resonance conditions hold, this slow subbundle of the linearized problem survives as a slow spectral submanifold (SSM) bundle $\mathcal{A}_0 =\bigcup\limits_{u_0\in \mathcal{U}} \mathcal{W}(E(\mathbf u_0))$ (see \citet{haller24_wa}). This slow SSM bundle is normally attracting and anchored to the critical manifold. \citet{haller24_wa} further proves that for relatively slow variations of the control input $\mathbf{u} \equiv \mathbf{u}^s(\epsilon t), \quad 0<\epsilon << 1$, the slow SSM subbundle $\mathcal{A}_0$ survives as a nearby diffeomorphic invariant manifold $\mathcal{A}_\epsilon$, which we call the adiabatic spectral submanifold (aSSM). The critical manifold also survives into a nearby invariant slow manifold $\mathcal{L}_\epsilon$, with parametrization $\mathbf{x}_{\epsilon}(\mathbf{u}(\epsilon t))$. Details pertaining to using this for controls can be found in \citet{Bettini2026_muscles}.

In \citet{Kaundinya2026}, for control inputs of the form 
\begin{equation}
\label{eq:gc}
\mathbf{u}(t) \equiv \mathbf{u}^s(\epsilon t) + \mathbf{u}^d(t), \quad \|\mathbf{u}^d(t)\| < \Delta, \quad \Delta>0,
\end{equation}
the survival of the aSSM, $\mathcal{A}_\epsilon$, as a nearby invariant manifold, $\tilde{\mathcal{A}}_{\epsilon}$, was discussed. They also provided explicit formulas of the aSSM-reduced model accounting for dynamics restricted to $\tilde{\mathcal{A}}_{\epsilon}$. While their formulas accounted only for the leading-order correction from the data, here we extend their results by deriving an aSSM-reduced model with higher-order correction terms that can be learned from data.

To that end, we note in soft robotic experiments, data representing the full phase space dynamics is unavailable. In fact, most motion capture systems record only positions and orientations from rigid-body markers attached to the soft robot. In such situations, we seek embeddings of the aSSM and its dynamics using data from the soft robot's observable space, $\mathbf{y} \in \mathbb{R}^{n_y}$, $n_y\ll n$. Specifically, we use the delay embedding function eq.~(\ref{eq:delay_embedding}), with $p=\Big\lfloor \frac{2d+1}{n_y} \Big\rfloor$, to generate a delay embedded observable space by  $\mathbf{y}$ that guarantees the reconstruction of the $d$ dimensional aSSM $\mathcal{A}_{\epsilon}$. We further denote the critical manifold parametrization in the observable space as $\mathbf{y}_0(\cdot)$.

We provide in Supplementary Section \ref{app:C2} a detailed derivation for the higher-order aSSM-reduced model in the phase space coordinates, $\mathbf{x}$. The derivation in that case parametrized the aSSM as a graph over the time-varying slow spectral subbundle $E(\mathbf{u}^s(\epsilon t))$. We similarly seek the aSSM as a graph over the corresponding $E(\mathbf{u}^s(\epsilon t))$ in the delay embedding observable space. To that end, we denote a time-varying matrix $\mathbf{V}(\mathbf u^s(\epsilon t))$ whose columns comprise the vectors spanning $E(\mathbf{u}^s(\epsilon t))$  in the delay embedding observable space. Using these together with the general form of the aSSM-reduced model eq.~(\ref{eq:aSSM_final}) in the phase space, we obtain the $d$-dimensional aSSM-reduced model in the delay-embedded observable space as:
\begin{align}
\label{eq:data-driven_case}
\mathbf{r} &= \mathbf{V}^\mathrm{T}(\mathbf u)\boldsymbol \mu (\mathbf{y}-\mathbf{y}_0(\mathbf u)), \qquad \mathbf{u} \in \mathcal{U}, \nonumber \\
\mathbf{\dot{r}} &= \mathbf{R}(\mathbf{r}, \mathbf{y}_0(\mathbf{u}(t))) + \mathbf{B}\left(\mathbf{u}(t)\right)   \mathbf{D}_{\mathbf{u}}\mathbf{y}_0(\mathbf u(t)) \dot{\mathbf{u}}, \\
\mathbf{y} &=   \mathbf{W}(\mathbf{r}, \mathbf{y}_0(\mathbf{u}(t)))+ \mathbf y_0(\mathbf{u}(t)). \nonumber 
\end{align}
Here $\mathbf{r} \in \mathbb{R}^d$ is the reduced coordinate obtained via an orthogonal projection onto $E(\mathbf{u}^s(\epsilon t))$, $\mathbf{R}(\cdot)$ is the leading-order aSSM reduced dynamics, $\mathbf{B}(\cdot)$ is calibration matrix that is accounting for a higher-order term in the aSSM reduced dynamics dependent on $\dot{\mathbf u}$, $\mathbf{W}(\cdot)$ is the leading-order aSSM parametrization.

Note that in the delay-embedded observable space, the spectral subspace families of the linearized system about the critical manifold are generally non-normal (see \citet{Bettini2026}). Recent work by \citet{Bettini2026} provides methods to recover general oblique projections for SSMs from data. For aSSMs, this methodology is still missing, hence we resort to orthogonal projections but account for the projection errors by learning a calibration matrix $\mathbf{B}\left(\mathbf{u}(t)\right)$ for the reduced dynamics.

\subsection{Data-driven identification of the aSSM-reduced model}\label{subsec:data-driven}

We collect the robot's response to step control input data in the observable space. The step control inputs are generated by performing jumps held for $t_{\mathrm{decay}}$-long seconds between the $m$ most distant control inputs for each element in the static control input set $\{\mathbf{U}^i_{\mathrm{static}}\}_{i=1}^{N}$. The parameter $t_{\mathrm{decay}}$ is chosen based on the time required for the robot to settle into a static configuration, and $m$ is chosen to sample the observable space well. We set $m=4$ for our applications. The set of static control inputs is sampled using the Latin hypercube sampling (LHS) technique (see \citet{latin_hypercube}). We denote this dataset as $\{\mathbf Y_{\mathrm{step}},\mathbf U_{\mathrm{step}}\}$. 

We further split this dataset into static
\begin{equation}
\{\mathbf{Y}^i_{\mathrm{static}},\mathbf U^i_{\mathrm{static}}\}_{i=1}^{N},
\end{equation} and decaying 
\begin{equation}
    \Big\{ \quad \mathbf{Y}^i_{\mathrm{decay}} \in \mathbb{R}^{n_y \times T}, \quad \mathbf U^i_{\mathrm{decay}}\in \mathbb{R}^{n_u \times T} \quad \Big\}_{i=1}^{mN},
\end{equation}
datasets, where $T$ is the temporal length of the decaying trajectory. We also collect the robot's response to random smooth control data generated with Perlin noise (\citet{perlin85}). We denote this response by $\{\mathbf Y_{\mathrm{calib}},\mathbf U_{\mathrm{calib}}\}$.

\subsubsection{Learning the critical manifold}
\label{subsec:learn_cm}
We learn the critical manifold parametrization $\mathbf{y}_0(\mathbf{u}_0)$ in the observable space as a nonlinear static mapping fit to the static dataset. The corresponding nonlinear regression problem takes the form
\begin{align}
\mathbf{y}_0^{\star}(\mathbf{u}) = \operatorname*{arg\,min}_{\mathbf{y}_0 : \mathcal{U}\subset\mathbb{R}^{n_u}\to\mathbb{R}^{n_y}} \sum_i^{N} \big\| \mathbf{Y}^i_{\mathrm{static}} - \mathbf{y}_0(\mathbf{U}^i_{\mathrm{static}}) \big\|^2.
\end{align}
We may seek the critical manifold $\mathcal{L}_0$ in the observable space  in the following three different ways:

(1) A polynomial expansion with polynomial order $k_u$,
    \begin{align}
        \mathbf{y}_0(\mathbf{u}) = \sum_{0\leq|\boldsymbol{\alpha}| \leq k_u} \mathbf{c}_{\boldsymbol{\alpha}} \, \mathbf{u}^{\boldsymbol{\alpha}},
    \end{align}
    where $\boldsymbol{\alpha}$ is a multi-index and $\mathbf{c}_{\boldsymbol{\alpha}}$ are the polynomial coefficients.
    
(2) A radial basis function expansion,
    \begin{align}
        \mathbf{y}_0(\mathbf{u}) = \sum_{j=1}^{n_c} \mathbf{w}_j \, \phi\big(r_j\big), \qquad r_j = \sqrt{\lVert \mathbf{u} - \mathbf{c}_u^{j} \rVert^2 + \epsilon},
    \end{align}
    where $\mathbf{c}_u^{j}$, $j = 1, \dots, n_c$, are the centers, $\mathbf{w}_j$ are the corresponding weights, and $\epsilon = 10^{-7}$ is a tolerance added to regularize against the singularity at $r_j = 0$. The kernel of $\phi(\cdot)$ is chosen as either a linear kernel,
    \begin{align}
        \phi(r) = r,
    \end{align}
    or a Gaussian kernel,
    \begin{align}
        \phi(r) = \exp\!\left(-\frac{r^2}{2\sigma^2}\right),
    \end{align}
    with parameter $\sigma$ controlling the width of each Gaussian basis function.

(3) A neural network,
    \begin{align}
        \mathbf{y}_0(\mathbf{u}) = \mathcal{N}_{\boldsymbol{\theta}}(\mathbf{u}),
    \end{align}
    where $\mathcal{N}_{\boldsymbol{\theta}}$ denotes the network with a trainable parameter vector $\boldsymbol{\theta}$.

For the first two parameterizations, the expansion is linear in the unknown coefficients ($\mathbf{c}_{\boldsymbol{\alpha}}$ or $\mathbf{w}_j$). Hence, the regression problem reduces to a linear least-squares problem that admits a quick, closed-form solution. The neural network parameterization, in contrast, is optimized using stochastic gradient descent due to its nonlinear dependence on $\boldsymbol{\theta}$.

To fit the critical manifold outlined above, we are not limited to static datasets generated solely by step inputs. We can also use slowly varying control inputs that span the robot's observable space, in a quasi-static sense, to enrich the static datasets.

\subsubsection{Learning the aSSM paramterization and reduced dynamics}
\label{subsec:learn_aSSM}
We center each decaying observable data about the static configuration they converge to we also truncate the initial part of the decaying signal at time time $t_0$, 
\begin{equation}
\mathbf{Y}^i_{\mathrm{centered}} =  (\mathbf Y^i_{\mathrm{decay}})_{:,t_0:T} -  (\mathbf Y^i_{\mathrm{decay}})_{:,T}.
\end{equation}
The truncation parameter $t_0$ is set to remove the initial transients that are still far from the dominant slow SSM. We then infer the minimal dimension $d$ of the slow SSM slices that capture the centered decaying data, based on possible symmetries of the robot and a spectrogram analysis of the decaying data for underdamped modes (see \citet{Kaundinya2026} for details). Using the inferred dimension $d$, we delay embed the centered data $p$ times, where $(p+1)n_y \geq 2d+1$ to guarantee a delay embedded observable to embed the slow SSM slices (see \citet{takens81}). We construct the delay-embedded dataset as follows, 
\begin{equation}
\mathbf Y^i_{\mathrm{delay}} = \begin{bmatrix}
(\mathbf{Y}^i_{\mathrm{centered}})_{:,p+1:T} \\
(\mathbf{Y}^i_{\mathrm{centered}})_{:,p:T-1} \\
\vdots \\
(\mathbf{Y}^i_{\mathrm{centered}})_{:,1:T-p}
\end{bmatrix} \in \mathbb{R}^{(p+1)n_y\times (T-p)}.
\end{equation}

Equipped with the delay embedded dataset $\{\mathbf Y^i_{\mathrm{delay}}\}_{i=1}^{mN}$, and grouping $m$ of the decays corresponding to the same static configuration, we perform an economy singular value decomposition on the grouped dataset to obtain a parameter set of chart maps $\Big\{\mathbf{V}^i \in \mathbb{R}^{(p+1)n_y \times d} \Big\}_{i=1}^{N}$. We learn interpolants $\mathbf{V}\left(\cdot\right)$ for the chart maps over the static set of inputs $\mathbf U_{\mathrm{static}}$. In some robotics applications, the charts for the near configurations around each static configuration can be very similar. In that case, we can learn a global chart map $\mathbf{V}_g \in \mathbb{R}^{(p+1)n_y \times d}$  by performing the singular value decomposition on the entire dataset. 
Using the learned chart maps, we construct data lying on the reduced space of the aSSM, 
\begin{equation}
\boldsymbol{\mathcal{R}}^i = (\mathbf{V}^i)^{\mathrm{T}} \mathbf{Y}^i_{\mathrm{delay}},
\end{equation}
and via finite differencing on $\boldsymbol{\mathcal{R}}^i$ we obtain reduced dynamics data on the aSSM, which we label as $\boldsymbol{\mathcal{\dot{R}}}^i$. 

We seek the leading-order aSSM parametrization and the aSSM-reduced dynamics as polynomial expressions, given by
\begin{align}
\label{eq:aSSMLearn_1}
    \mathbf{W}(\mathbf{r}, \mathbf{y}) &=  \sum_{1 \leq |\boldsymbol{\alpha}| \leq k^W_r} \sum_{0 \leq |\boldsymbol{\beta}| \leq k^W_y} \mathbf{W}_{\boldsymbol{\alpha}\boldsymbol{\beta}} \, \mathbf{r}^{\boldsymbol{\alpha}} \, \mathbf{y}^{\boldsymbol{\beta}},
\end{align}
and the aSSM-reduced dynamics as
\begin{align}
    \mathbf{R}(\mathbf{r}, \mathbf{y}) &= \sum_{1 \leq |\boldsymbol{\alpha}| \leq k^R_r} \sum_{0 \leq |\boldsymbol{\beta}| \leq k^R_y} \mathbf{R}_{\boldsymbol{\alpha}\boldsymbol{\beta}} \, \mathbf{r}^{\boldsymbol{\alpha}} \, \mathbf{y}^{\boldsymbol{\beta}},
\end{align}
where $\boldsymbol{\alpha}$ and $\boldsymbol{\beta}$ are multi-indices over $\mathbf{r}$ and $\mathbf{y}$ respectively, $k^W_r, k^W_y$ and $k^R_r, k^R_y$ denote the corresponding polynomial orders, and $\mathbf{w}_{\boldsymbol{\alpha}\boldsymbol{\beta}}$, $\mathbf{r}_{\boldsymbol{\alpha}\boldsymbol{\beta}}$ are the unknown polynomial coefficients.

We solve for the polynomial coefficients by minimizing the reconstruction and dynamics regression losses
\begin{align}
\mathbf{W}^{\star}(\mathbf{r},\mathbf{y}) &= \operatorname*{arg\,min}_{\mathbf{W}} \sum_i \big\| \mathbf{Y}^i_{\mathrm{centered}} - \mathbf{W}\big(\boldsymbol{\mathcal{R}}^i, \left(\mathbf{Y}^i_{\mathrm{decay}}\right)_{:,T}\big) \big\|^2, \\
\mathbf{R}^{\star}(\mathbf{r},\mathbf{y}) &= \operatorname*{arg\,min}_{{\mathbf{R}}} \sum_i \big\| \dot{\boldsymbol{\mathcal{R}}}^i - \mathbf{R}\big(\boldsymbol{\mathcal{R}}^i, \left(\mathbf{Y}^i_{\mathrm{decay}}\right)_{:,T}\big) \big\|^2.
\end{align}

Finally, using the learned leading-order aSSM parametrization $\mathbf{W}^{\star}$, the aSSM-reduced dynamics $\mathbf{R}^{\star}$, and the critical manifold parametrization $\mathbf{y}_0(\mathbf{u})$, we infer the calibration matrix $\mathbf{B}(\mathbf{u})$ as a polynomial expression, 
\begin{align}
    \mathbf{B}(\mathbf{u}) &=  \sum_{0 \leq |\boldsymbol{\alpha}| \leq k^B_u} \mathbf{B}_{\boldsymbol{\alpha}} \, \mathbf{u}^{\boldsymbol{\alpha}} \, ,
\end{align}
by minimizing the following regression problem using the calibration dataset:
\begin{align}
\label{eq:aSSMLearn_fin}
    \mathbf{B}^{\star}(\mathbf{u}) = \operatorname*{arg\,min}_{\mathbf{B}(\mathbf{u})} \Big\| \dot{\boldsymbol{\mathcal{R}}}_{\mathrm{calib}} - \mathbf{R}^{\star}\big(\boldsymbol{\mathcal{R}}_{\mathrm{calib}}, \mathbf{y}^\star_0(\mathbf{U}_{\mathrm{calib}})\big) - \\\mathbf{B}\big(\mathbf{U}_{\mathrm{calib}}\big)\, \mathbf{D}_{\mathbf{u}}\mathbf{y}^\star_0\big(\mathbf{U}_{\mathrm{calib}}\big)\, \dot{\mathbf{U}}_{\mathrm{calib}} \Big\|^2,
\end{align}
Here $\boldsymbol{\mathcal{R}}_{\mathrm{calib}} = \mathbf{V}^{\mathrm{T}}(\mathbf{U}_{\mathrm{calib}})\mathbf{Y}_{\mathrm{calib}}$ and $\dot{\boldsymbol{{\mathcal{R}}}}_{\mathrm{calib}}$ is obtained by finite-differencing $\boldsymbol{\mathcal{R}}_{\mathrm{calib}}$ . The choice of all parameters involved in deducing these terms is optimized to reduce the normalized mean trajectory error on test data. We also test the predictive power of the learned aSSM-reduced model on unseen control inputs $\mathbf{u}(t)$ before deploying it for closed-loop control.

\subsection{Model predictive control using aSSM-reduced model}\label{subsec:mpc}

We use the aSSM-reduced model in eq.~(\ref{eq:data-driven_case}) in a model predictive control (MPC) scheme to track a target track $\boldsymbol{\Gamma}(t)$. The MPC is formulated as a finite-horizon optimization problem. The finite horizon is set up from time $t_j$ recorded on the camera sensors monitoring the robot, with a user-defined window uniformly spaced in the space window width $M \Delta t$, with time spacing $\Delta t$. We also take into account motion capture $\mathbf{y}(t_j)$. We then solve for the optimal control inputs that will follow the target track in this window,  
\begin{align}
    \label{eq:aSSM_mpc0}
   \text{minimize}_{\mathbf{u}(\cdot)}~ \quad
        &\int_{t_j}^{t_{j}+M\Delta t} \Big(||\mathbf{z}(s)-\mathbf{\Gamma}(s)||^2_\mathbf{Q_z} + ||\mathbf{u}(s)||^2_\mathbf{R_u} + ||\mathbf{\dot{u}}(s)||^2_\mathbf{R_{\dot{u}}}\Big) ds, \nonumber \\
        \mathrm{subject~to}~ \quad
            & \mathbf{r}(t_j) = \mathbf{V}^\mathrm{T} (\mathbf{u}(t_j)) \boldsymbol{\mu}(\mathbf{y}(t_j)-\mathbf{y}_0(\mathbf u (t_j))), \nonumber\\ 
            & \dot{\mathbf{r}} = \mathbf{R}(\mathbf{r},\mathbf{u}(t)) + \mathbf{B}(\mathbf{u}(t)) \mathbf{D}_{\mathbf u} \mathbf y_0(\mathbf{u}(t))\dot{\mathbf{u}}(t),  \\ 
            & \mathbf{y}(t) = \mathbf{W}(\mathbf{r}(t),\mathbf{u}(t)) + \mathbf{y}_0(\mathbf{u}(t)),  \nonumber \\ 
            &\mathbf{y}(t) \in \mathcal{Y}, \quad  \mathbf{u}(t) \in \mathcal{U}, \nonumber \\
              & \mathbf{z}(t) = \mathbf{C}(\mathbf{y}(t)). \nonumber \\
\end{align}

The non-convex optimization problem (\ref{eq:aSSM_mpc0}) is solved using the \textit{GuSTO} algorithm of \citet{pavone192}, which recasts the problem as a sequence of convex local optimal control problems (LOCPs). At each iteration, the resulting LOCP is a quadratic program, and we solve it using the high-performance interior-point solver (HPIPM) by \citet{frison2020hpipm}. To enable real-time operation, only the first two control inputs of the optimal sequence are applied to the robot, after which the prediction horizon is shifted forward by the same amount and the LOCP is resolved, yielding a receding-horizon implementation.

\backmatter


\subsection*{Acknowledgments}
This work has been partially supported by grant $200021\_214908$ from the Swiss National Science
Foundation (SNF).

\subsection*{Author Contributions}
\ja
{
G.H., M.P., J.A. and R.K. designed the research. R.K. and M.L carried out the research. M.L. and R.S. designed the hardware. R.K., M.L., H.B. and L.P. developed the software. R.K. wrote the paper. G.H., M.P. and R.K. led the research team. M.P. provided funding for hardware resources. 
}

\subsection*{Competing Interests}
\ja{
All authors declare no financial or non-financial competing interests. 
}


\bigskip

\clearpage
\setcounter{section}{0}
\setcounter{figure}{0}
\setcounter{table}{0}
\setcounter{equation}{0}
\renewcommand{\theHsection}{S\arabic{section}}
\renewcommand{\theHsubsection}{S\arabic{section}.\arabic{subsection}}
\renewcommand{\theHfigure}{S\arabic{figure}}
\renewcommand{\theHtable}{S\arabic{table}}
\renewcommand{\theHequation}{S\arabic{equation}}

\begin{center}
  {\Titlefont \textit{Supplementary Material for}\\[4pt]
   Data-Driven Precision Control of Soft Helicoid Robots\par}
\end{center}
\vskip 20pt


\section{Soft helicoid design: Morphology and fabrication trade-offs}\label{subsec:design_supp}

\subsection{Morphology comparison}
Once buckling initiates in a homogeneous hollow-cylinder morphology, the monotonic relationship between tendon displacement and backbone curvature is broken, causing the trunk to collapse rather than bend smoothly. A structural bifurcation occurs, rendering the kinematic mapping invalid and precluding accurate control. This motivated the choice of the Trimmed Helicoid (TH) architecture (Section~\ref{subsec:design}), which independently tunes axial and bending stiffness to avoid this failure mode.

\subsection{Fabrication method comparison}

Manual casting was discarded due to the complexity of the multi-stage molding process required for the internal topology, which introduces alignment errors and limits the geometric intricacy that can be realized. Extrusion-based methods were rejected because their layer-by-layer anisotropy creates stress concentrations that are prone to delamination under high-frequency cyclic loading (see \citet{S:yap2020review} for a review). Digital Light Synthesis (DLS\texttrademark), driven by Continuous Liquid Interface Production (CLIP\texttrademark), avoids both issues: an oxygen-permeable window inhibits polymerization at the interface (dead zone), enabling continuous part generation and monolithic, isotropic structures. EPU~46 was selected for this process for its high resilience and tear strength alongside its sustainability (40\% bio-based).

\subsection{Design-space exploration}
Existing theoretical models for TH structures are restricted to small-deformation regimes and do not capture the viscoelastic behavior of the printed elastomer. Also, the lack of prior characterization of the TH--DLS--EPU~46 hardware triad made predictive simulation unreliable. We therefore iterated through sixteen morphological variations (Fig.~\ref{fig:design_full}a) to empirically converge on parameters $P1$--$P4$. Microstructures ($P4$) reduced weight but introduced stochastic friction, degrading control transparency, so a solid-wall topology was selected instead.

\section{Closed-loop results for a faster varying 3D Pacman track}
\label{app:pacman_fast}

\begin{figure}[t!]
    \centering
    \begin{subfigure}[b]{0.48\linewidth}
        \centering
        \includegraphics[width=\linewidth]{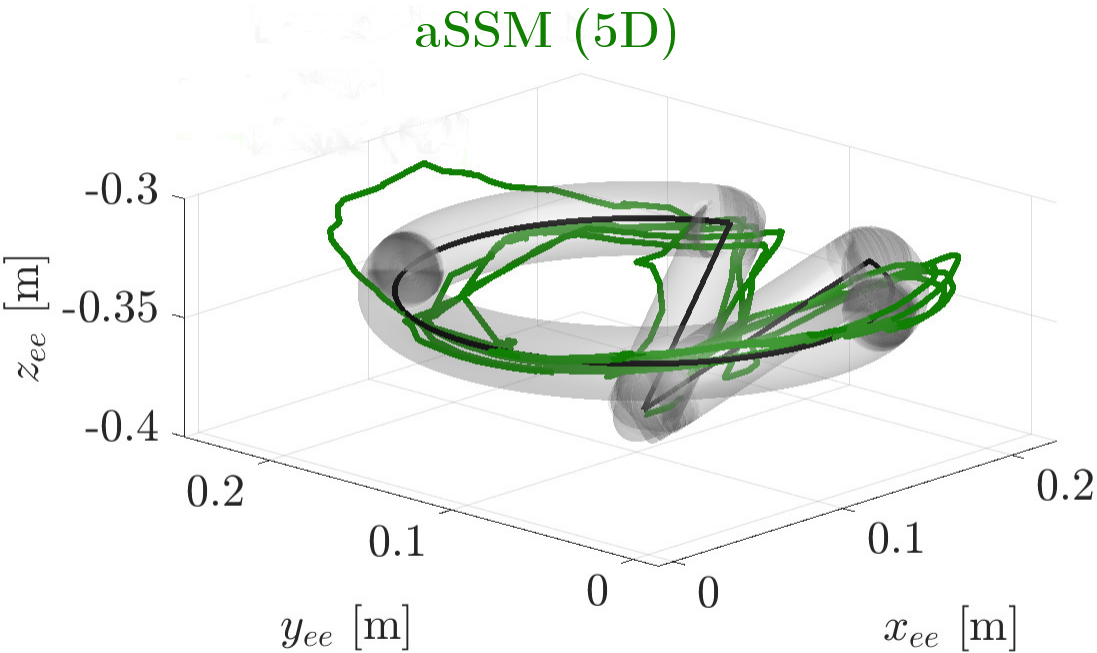}
        \label{fig:aSS_fast_pac}
    \end{subfigure}
    \hfill
    \begin{subfigure}[b]{0.48\linewidth}
        \centering
        \includegraphics[width=\linewidth]{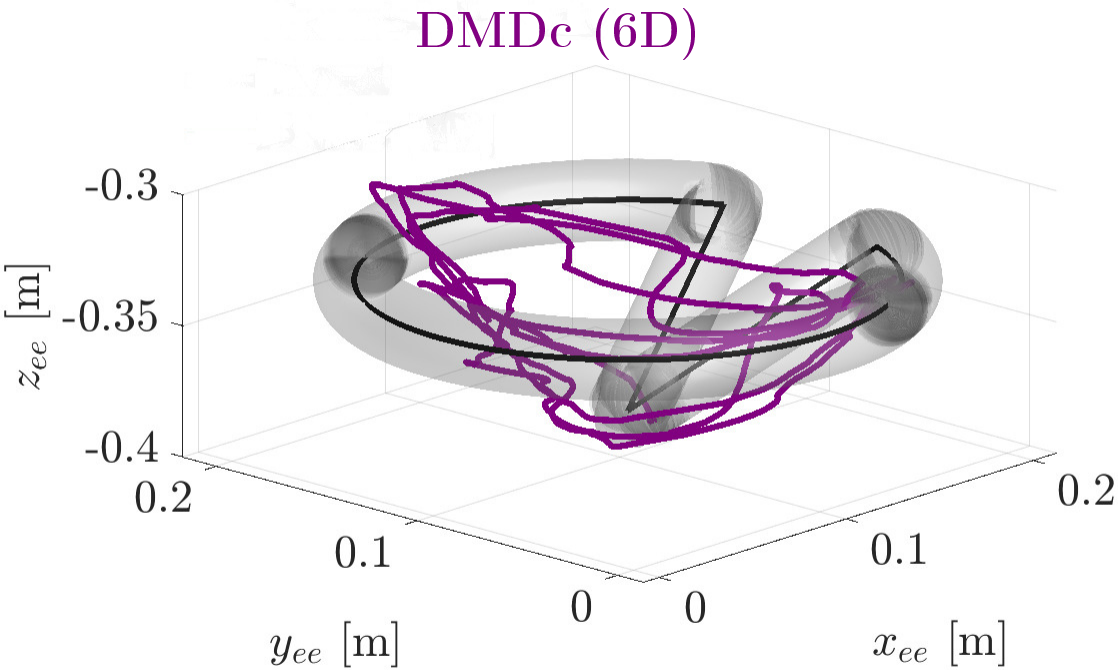}
        \label{fig:DMDc_fast_pac}
    \end{subfigure}

    \begin{subfigure}[b]{0.48\linewidth}
        \centering
        \includegraphics[width=\linewidth]{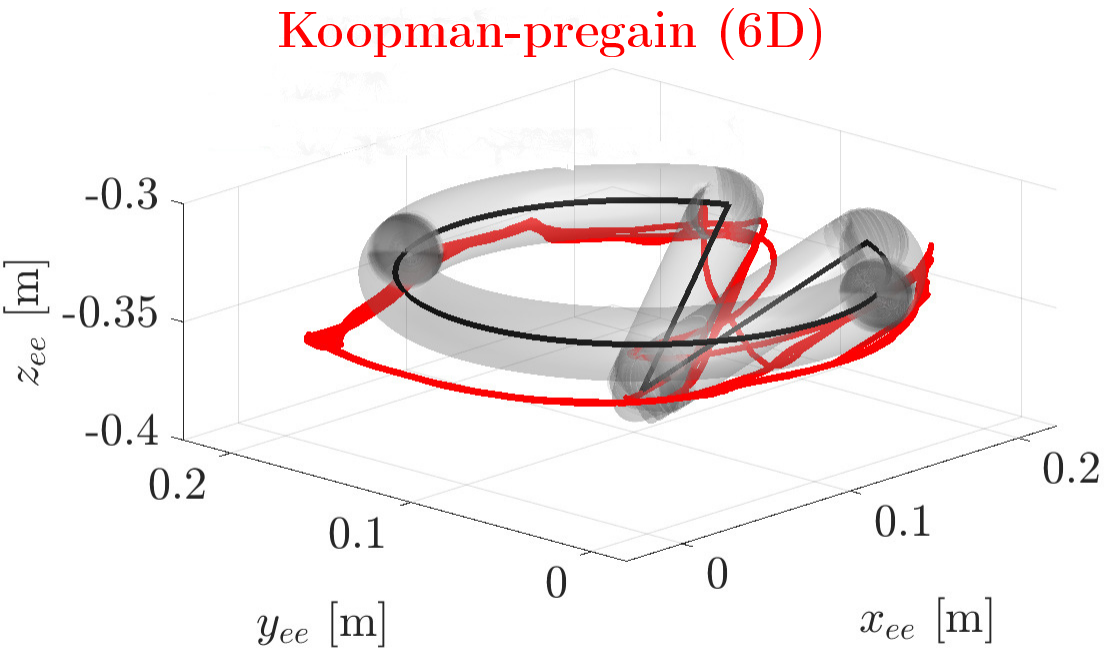}
        \label{fig:Koop_rand_fast_pac}
    \end{subfigure}
    \hfill
    \begin{subfigure}[b]{0.48\linewidth}
        \centering
        \includegraphics[width=\linewidth]{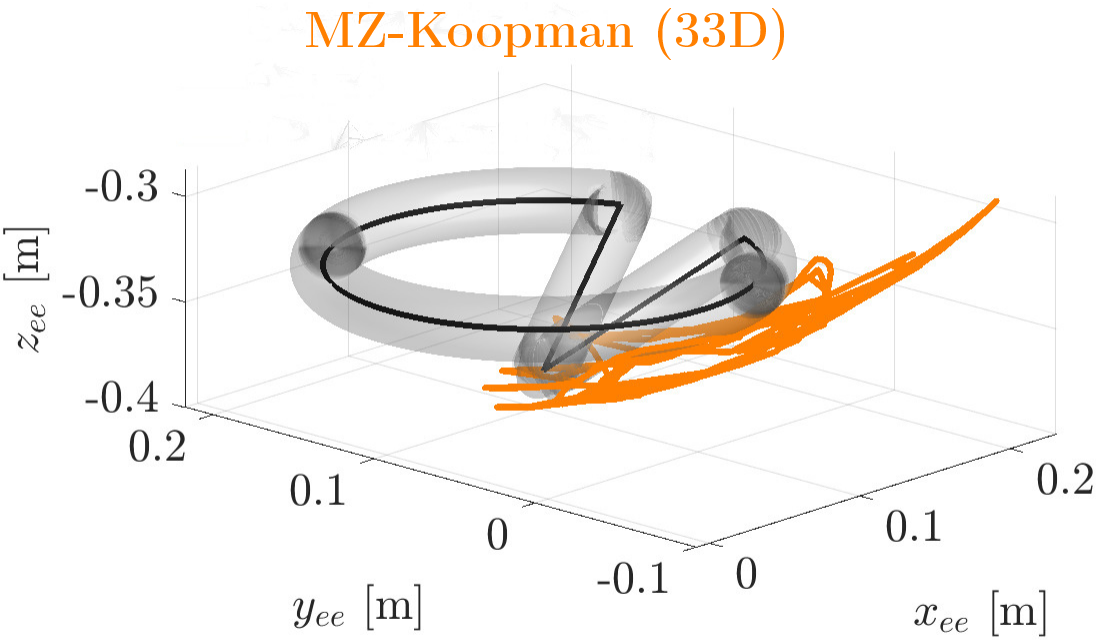}
        \label{fig:MZ_Koop_fast_pac}
    \end{subfigure}
    
    \begin{subfigure}[b]{0.48\linewidth}
        \centering
        \includegraphics[width=\linewidth]{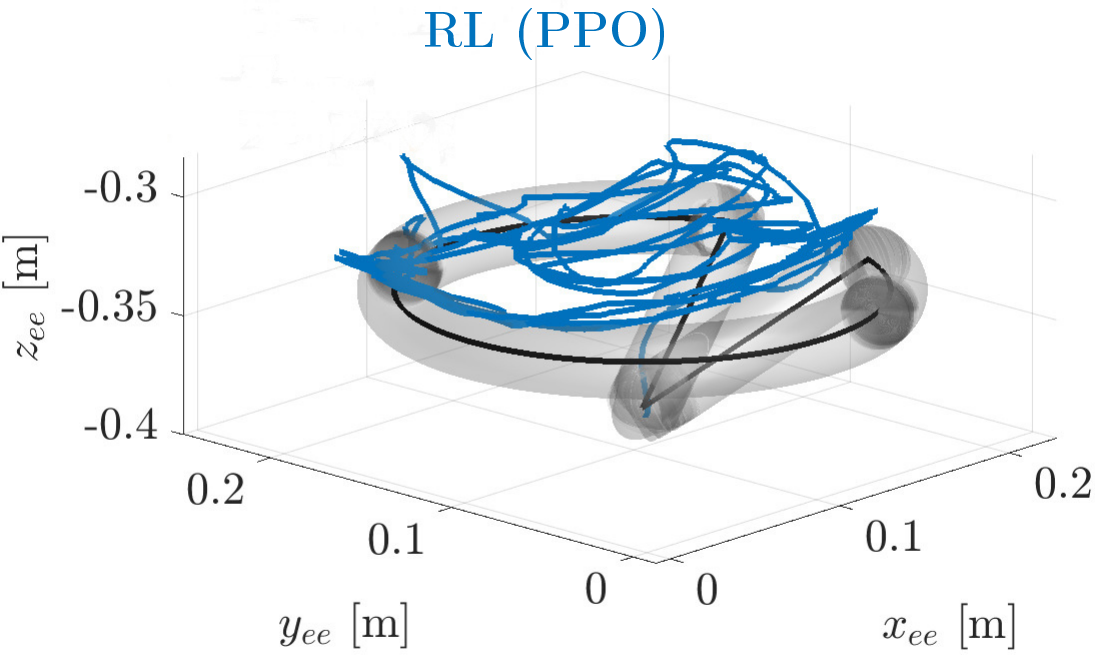}
        \label{fig:RL_fast_pac}
    \end{subfigure}

    \caption{Closed-loop tracking results for an extremely fast 3D Pacman track (black) with the uncertainty tube of radius $3$ [cm] plotted in gray. The average instantaneous speed along the target track is $12$ [cm/s].}
    \label{fig:pacman_fast_fast}
\end{figure}

\begin{figure}[t!]
    \centering
    
        \includegraphics[width=0.6\linewidth]{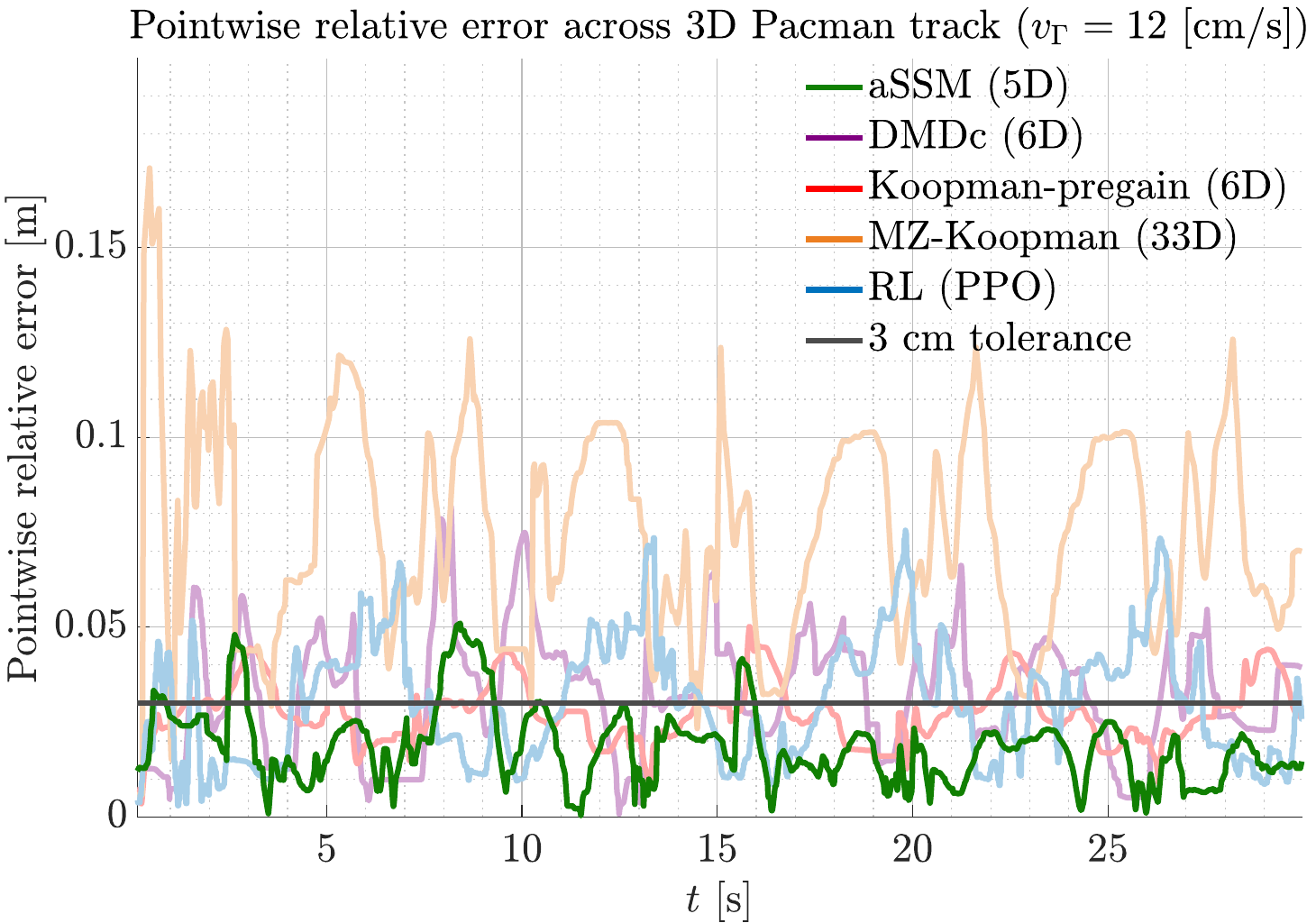}
        
    \caption{Pointwise relative error plots for closed-loop tracking for a 3D Pacman track with speed $12$ [cm/s]}
    \label{app:fig::p3D_fast_error}
\end{figure}

We present here closed-loop experimental results for the 3D Pacman track (appearing in the main text in Section \ref{subsec:closed-loop}) in Fig.~\ref{fig:pacman_fast_fast}. This track is speeded up to have a mean instantaneous speed of $v_{\boldsymbol \Gamma} = 12$ [cm/s] with slowness measure $r_s = 0.2$. This is a challenging 3D track, as within a span of a second, the trunk must make a sharp lift-off. We find that, despite delays and motor hardware uncertainty, the 5D aSSM-based controller provides the most consistent performance among all baselines considered. The closed-loop errors start to compound for the RL policy and the MZ-Koopman controller, leading to highly inaccurate tracking performance. The linear methods performs without delay because it is fast, but ultimately fail to lift the trunk and maintain the height. We support these results with pointwise relative error plots in Fig.~(\ref{app:fig::p3D_fast_error}).

\section{Relative error plots for spherical obstacle avoidance}
\label{app:sphere_appendix}

\begin{figure}[t!]
    \centering
    \begin{subfigure}[b]{0.6\linewidth}
        \centering
        \includegraphics[width=\linewidth]{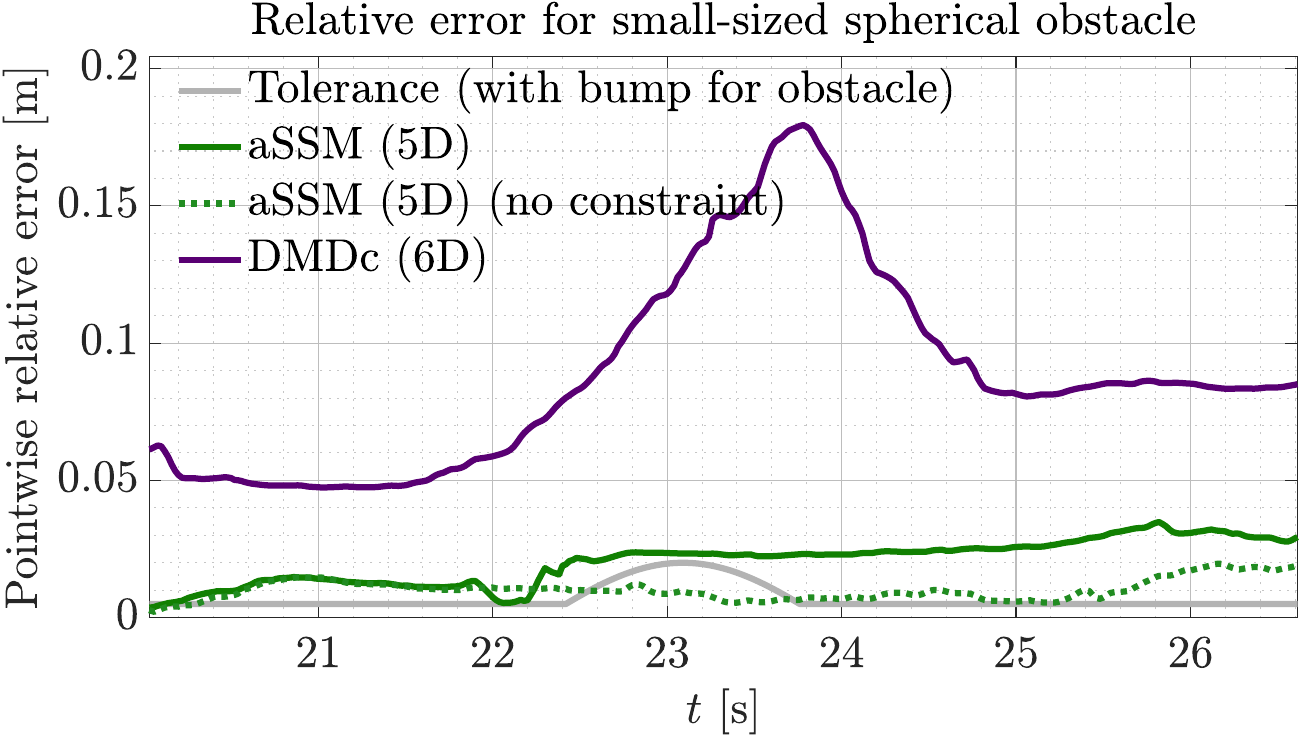}
        \caption{}
        \label{fig:app:sph1}
    \end{subfigure}
    \hfill
    \begin{subfigure}[b]{0.48\linewidth}
        \centering
        \includegraphics[width=\linewidth]{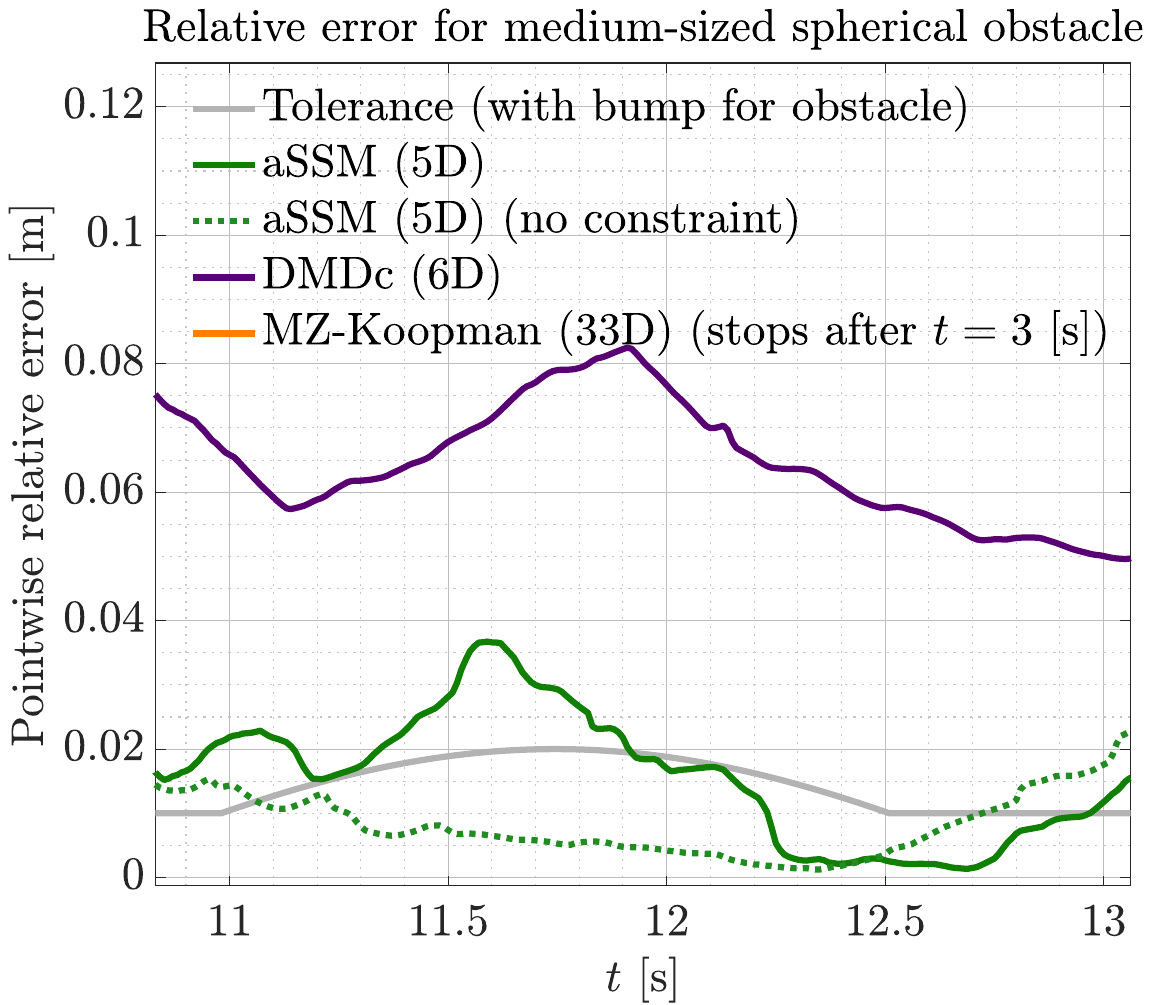}
        \caption{}
        \label{fig:app:sph2}
    \end{subfigure}
    \caption{Pointwise relative error plots for (a) the small-sized spherical obstacle and (b) the medium-sized spherical obstacle avoidance task. The tolerance line accounts for the sphere's location in real time as the target track is traversed.}
    \label{fig:app:rel_sphere}
\end{figure}

For completeness, we present here the pointwise relative error plots for the small-sized (see Fig.~(\ref{fig:app:sph1})) and medium-sized (see Fig.~(\ref{fig:app:sph2})) spherical obstacle avoidance tasks discussed in Section \ref{subsec:constraints}. Both plots show that the aSSM-based controller with constraints activated grazes past the spherical obstacles, whereas the one without constraints minimizes the tracking error to the target. The DMDc controller does not track the target closely and avoids the spherical obstacle by a large distance. The DMDc model lacks precision information to avoid the spheres delicately and brute-forces its way out of the region. The MZ-Koopman model fails to solve the optimization problem in real time. As a result, it stops abruptly at $t=3$ [s] for the medium-sized spherical obstacle task and at $t=10$ [s] for the small-sized spherical obstacle task.

\section{Soft trunk avoiding a moving cylindrical obstacle along a complex path}
\label{app:cylinder_complex}
\begin{figure}[t!]
    \centering
    \begin{subfigure}[b]{0.6\linewidth}
        \centering
        \includegraphics[width=\linewidth]{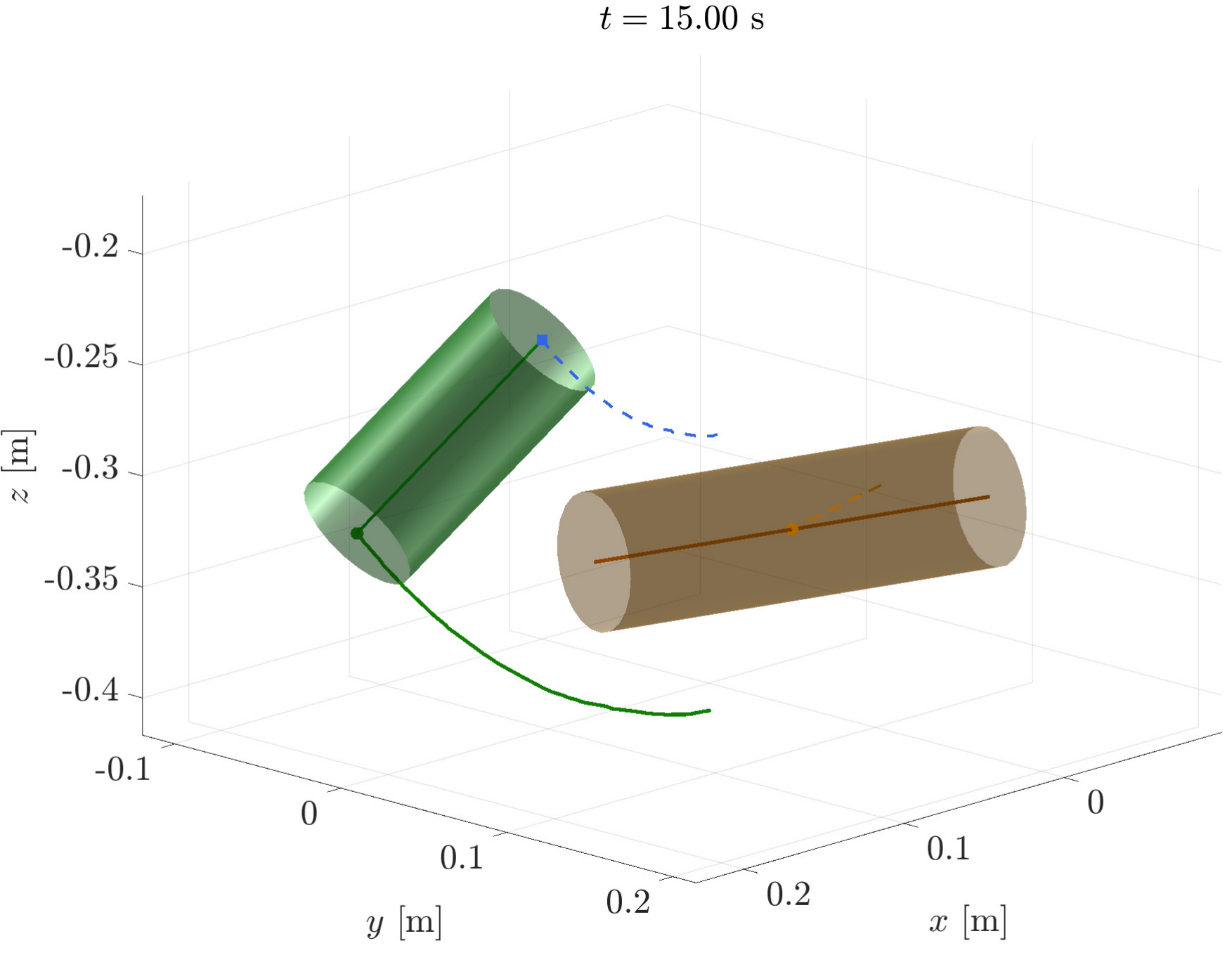}
        \caption{}
        \label{fig:app:cyl1}
    \end{subfigure}
    \hfill
    \begin{subfigure}[b]{0.6\linewidth}
        \centering
        \includegraphics[width=\linewidth]{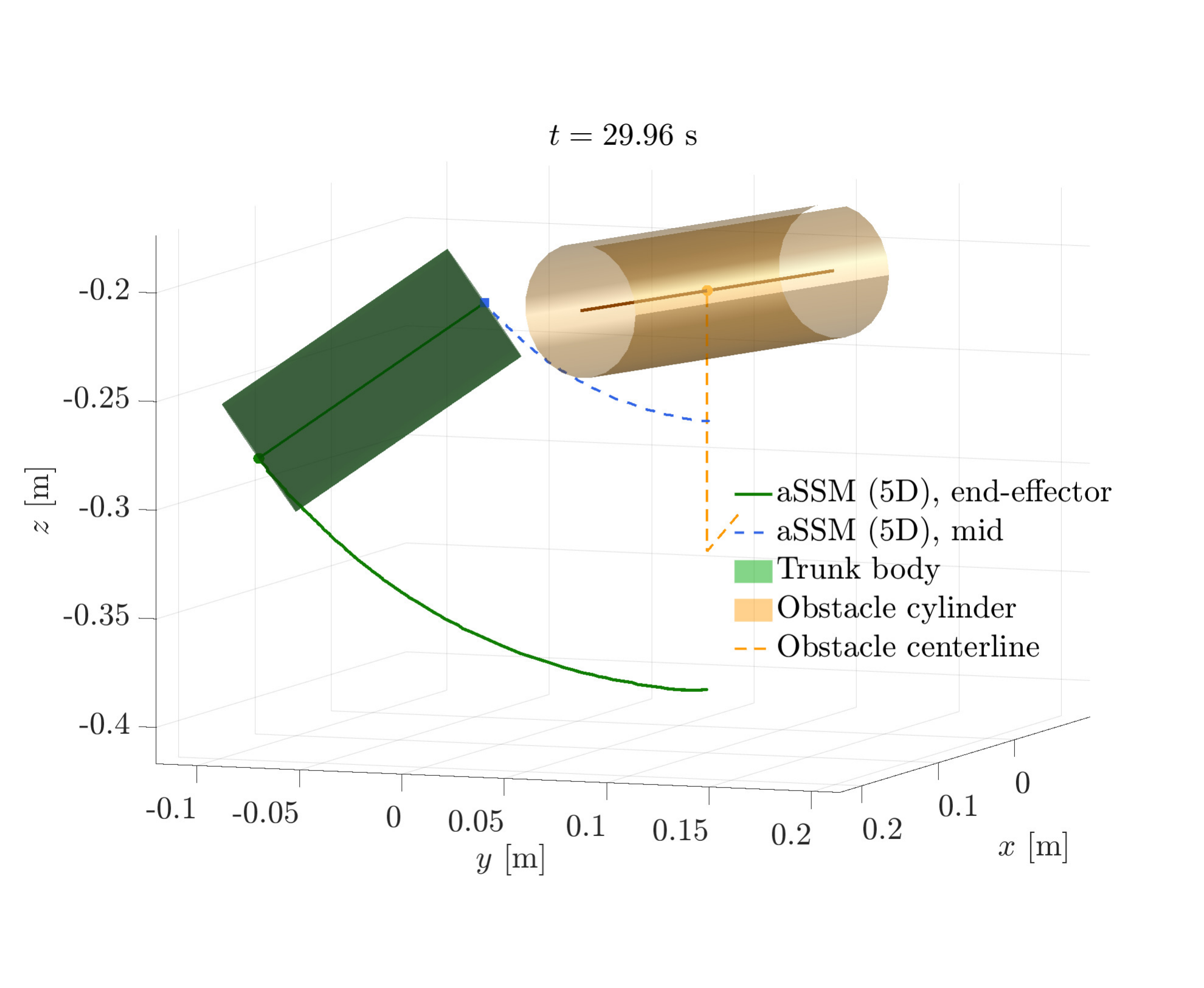}
        \caption{}
        \label{fig:app:cyl2}
    \end{subfigure}
    \caption{Snapshots of the trunk main body avoiding a moving cylindrical obstacle at (a) $t=15$ [s] as it approaches the rest state and (b) $t=30$ [s] as the obstacle moves up in the vertical direction. The trunk body is green, and the cylindrical obstacle is yellow.}
    \label{fig:app:cylfig}
\end{figure}

In this section, we provide closed-loop results for the soft trunk avoiding a moving cylinder. This is a continuation of Section \ref{subsec:cylinder} for a complex centerline trajectory path of the cylindrical obstacle. The path starts similarly to the ones described in the Section \ref{subsec:cylinder}, but after performing the rotation, it rises vertically in the $z$ direction. We expect the trunk body to react by avoiding the moving obstacle by conforming to a nonlinear bending shape. We precisely see these in the snapshot plots in Fig.~\ref{fig:app:cyl1} and Fig.~\ref{fig:app:cyl2}. The trunk body rises to a height of $0.25$ [m] to avoid the moving cylindrical obstacle.

\section{Derivation of the higher-order aSSM-reduced model}
\label{app:C2}
Here, we derive the aSSM-reduced model and 
Using the coordinate change
\begin{equation}
    \label{eq:flatten_cm}
    \boldsymbol{\zeta}  = \mathbf{x} -\mathbf{x}_0(\mathbf u),
\end{equation}
we monitor the error between the phase space coordinate and the critical manifold's prediction for general control inputs eq.~(\ref{eq:gc}). Differentiating eq. (\ref{eq:flatten_cm}), we obtain
\begin{equation}
    \dot{\boldsymbol{\zeta}}(t)  = \dot{\mathbf{x}}(t) -\mathbf{D}_{\mathbf{u}}\mathbf{x}_0(\mathbf u(t)) \dot{\mathbf{u}},
\end{equation}
\begin{equation}
\label{eq:xi_x0}
    \dot{\boldsymbol{\zeta}}(t)  = \mathbf{F}\left(\boldsymbol{\zeta}(t) + \mathbf{x}_0(\mathbf u(t)),\mathbf u(t) \right) -\mathbf{D}_{\mathbf{u}}\mathbf{x}_0(\mathbf u(t)) \dot{\mathbf{u}},
\end{equation}
Taylor expanding eq. (\ref{eq:xi_x0}) about $\mathbf x_0(\mathbf u(t))$ then gives, 
\begin{align}
    \dot{\boldsymbol{\zeta}}(t)  = \mathbf{F}&\left(\mathbf{x}_0( \mathbf  u (t)),\mathbf u(t) \right) + \mathbf{D}_{\boldsymbol{\zeta}}\mathbf{F}(\mathbf{x}_0(\mathbf u(t)),\mathbf{u}(t))   \boldsymbol{\zeta} + \dots \\ &+ \mathbf{F}_{0}(\boldsymbol \zeta, \mathbf{x}_0(\mathbf u(t)), \mathbf{u}(t))-\mathbf{D}_{u}\mathbf{x}_0(\mathbf u(t)) \dot{\mathbf{u}}, \quad \mathbf{F}_{0} = o(|\boldsymbol \zeta|). \nonumber
\end{align}
Using the critical manifold definition from eq.~(\ref{eq:cm_condition}) of the main text, we find that the above equation simplifies to, 
\begin{equation}
\label{eq:slow-shift}
    \dot{\boldsymbol{\zeta}}(t)  = \mathbf{D}_{\boldsymbol \zeta}\mathbf{F}(\mathbf{x}_0(\mathbf u(t)),\mathbf{u}(t))  \boldsymbol{\zeta} + \mathbf{F}_{0}(\boldsymbol \zeta, \mathbf{x}_0(\mathbf u(t)), \mathbf{u}(t))-\mathbf{D}_{\mathbf{u}}\mathbf{x}_0(\mathbf u(t)) \dot{\mathbf{u}}.
\end{equation}

We substitute the form in eq.~(\ref{eq:gc}) for the general control inputs $\mathbf{u}(t)$ and expand eq.~(\ref{eq:slow-shift}) about the trajectory $\left(\mathbf{x}_0(\mathbf{u^s}(\epsilon t)),\mathbf{u^s}(\epsilon t)\right)$ in the phase space. This results in the following set of equations (with  higher-order terms written out):
\begin{align}
\label{eq:higher-aSSM}
    \dot{\boldsymbol{\zeta}}(t)  = \mathbf{D}_{\boldsymbol \zeta}\mathbf{F}(\mathbf{x}_0(\mathbf u^s(\epsilon t)),\mathbf{u}^s(\epsilon t))  \boldsymbol{\zeta} + \mathbf{F}_{0}(\boldsymbol \zeta, \mathbf{x}_0(\mathbf u^s(\epsilon t)), \mathbf{u}^s(\epsilon t))\\
    -\mathbf{D}_{\mathbf{u}}\mathbf{x}_0(\mathbf u^s(\epsilon t)) \dot{\mathbf{u}} + O\left((\epsilon+\delta)^2,\epsilon |\boldsymbol \zeta|,\delta |\boldsymbol \zeta|\right).
\end{align}
We further make the assumed time-scale separation explicit in eq.~(\ref{eq:higher-aSSM}) by introducing a relatively slowly varying parameter $\alpha$, 
\begin{align}
\label{eq:time-scale}
    \dot{\boldsymbol{\zeta}}(t)  &= \mathbf{D}_{\boldsymbol \zeta}\mathbf{F}(\mathbf{x}_0(\mathbf u^s(\alpha)),\mathbf{u}^s(\alpha))  \boldsymbol{\zeta} + \mathbf{F}_{0}(\boldsymbol \zeta, \mathbf{x}_0(\mathbf u^s(\alpha)), \mathbf{u}^s(\alpha))\nonumber \\ 
    &\quad \quad \quad -\mathbf{D}_{\mathbf{u}}\mathbf{x}_0(\mathbf u^s(\alpha)) \dot{\mathbf{u}} + O\left((\epsilon+\delta)^2, \epsilon |\boldsymbol \zeta|,\delta |\boldsymbol \zeta|\right), \\ \nonumber
    \dot{\alpha}&= \epsilon.
\end{align}

In the frozen-time limit, $\epsilon =0$ and $\dot{\mathbf{u}}=0$, the decomposition in eq. (\ref{eq:slow_subbundle}) of the main paper applies, which allows us to represent the family of the $d$ slowest right eigenvectors of the linear operator $\mathbf{D}_{\boldsymbol \zeta}\mathbf{F}(\mathbf{x}_0(\mathbf u^s(\alpha)),\mathbf{u}^s(\alpha))$ as columns of the matrix $\mathbf{V}(\mathbf{u}^s(\alpha)) \in \mathbb{R}^{n \times d}$. This enables us to define a local coordinate along the slow spectral subbundle $\mathbf{E}(\mathbf{u}^s(\alpha))$, 
\begin{equation}
    \mathbf{r} = \mathbf{V}^{\mathrm T}(\mathbf{u}^s(\alpha)) \boldsymbol{\zeta} , \quad \mathbf{r} \in \mathbb{R}^d.
\end{equation}
Using this, we can define the aSSM fiber parametrization, 
\begin{equation}
\boldsymbol{\zeta}(t) = \mathbf{W}(\mathbf{r}, \mathbf{x}_0(\mathbf{u}^s(\epsilon t))) + \mathbf{\tilde{W}}(\mathbf{x}_0(\mathbf{u}^s(\epsilon t)),t)+ O\left((\epsilon+\delta)^2,\epsilon|\mathbf{r}|, \delta |\mathbf{r}|\right),
\end{equation}
and the aSSM reduced dynamics,   
\begin{equation}
\dot{\mathbf{r}} = \mathbf{R}(\mathbf{r}, \mathbf{x}_0(\mathbf{u}^s(\epsilon t)))  + \tilde{\mathbf{R}}(\mathbf{x}_0(\mathbf{u}^s(\epsilon t)),t) + O\left((\epsilon+\delta)^2,\epsilon|\mathbf{r}|,\delta |\mathbf{r}|\right).
\end{equation}
where we seek them as Taylor expansions in $\epsilon$, $\delta$, and $\mathbf{r}$, and the terms $\tilde{\mathbf W}$ and $\tilde{\mathbf R}$ represent the terms at $O(\epsilon +\delta)$.  

The aSSM parameterization and the reduced dynamics satisfy the following invariance equations, 
\begin{align}
\label{eq:ie-leading}
&\mathbf{D}_{\boldsymbol \zeta}\mathbf{F}\big(\mathbf{x}_0(\mathbf u^s(\alpha)),\mathbf{u}^s(\alpha)\big)\, \mathbf{W}\big(\mathbf{r}, \mathbf{x}_0(\mathbf{u}^s(\epsilon t))\big) \nonumber \\
&\quad + \mathbf{F}_{0}\big(\mathbf{W}(\mathbf{r}, \mathbf{x}_0(\mathbf{u}^s(\epsilon t))) , \mathbf{x}_0(\mathbf u^s(\alpha)), \mathbf{u}^s(\alpha)\big) \nonumber \\
&= \mathbf{D}_{\mathbf r}\mathbf{W}\big(\mathbf{r}, \mathbf{x}_0(\mathbf{u}^s(\epsilon t))\big)\,\mathbf{R}\big(\mathbf{r}, \mathbf{x}_0(\mathbf{u}^s(\epsilon t))\big) ,
\end{align}
at leading order. At higher order, we obtain
\begin{align}
\label{eq:ie-higher-}
&\mathbf{D}_{\boldsymbol \zeta}\mathbf{F}\big(\mathbf{x}_0(\mathbf u^s(\alpha)),\mathbf{u}^s(\alpha)\big)\, \tilde{\mathbf{W}}\big(\mathbf{x}_0(\mathbf{u}^s(\epsilon t)),t\big) - \mathbf{D}_{\mathbf{u}}\mathbf{x}_0(\mathbf u^s(\alpha))\, \dot{\mathbf{u}} \nonumber \\
&= \mathbf{D}_{\mathbf r}\mathbf{W}\big(\mathbf{0}, \mathbf{x}_0(\mathbf{u}^s(\epsilon t))\big)\, \tilde{\mathbf{R}}\big(\mathbf{x}_0(\mathbf{u}^s(\epsilon t)),t\big) + \dot{\tilde{\mathbf{W}}}\big(\mathbf{x}_0(\mathbf{u}^s(\epsilon t)),t\big).
\end{align}
Eq.~(\ref{eq:ie-leading}) is solved by seeking Taylor expansions for the aSSM parameterization and the aSSM-reduced dynamics in the reduced coordinates. \citet{S:haller24_wa} derive recursive algebraic solutions for these Taylor coefficients with initial condition $\mathbf{D}_{\mathbf r}\mathbf{W}(\mathbf{0}, \mathbf{x}_0(\mathbf{u}^s(\epsilon t))) = \mathbf{V}(\mathbf{u}^s(\epsilon t))$. We adopt the same choices to derive the higher-order terms. 

Note that we still need to choose either of the unknowns in eq.~(\ref{eq:ie-higher-}) to solve it. We choose that the higher-order reduced dynamics has the form $\tilde{\mathbf{R}}(\mathbf{x}_0(\mathbf{u}^s(\epsilon t)),t) = -\mathbf{V}^\mathrm{T}(\mathbf u^s(\epsilon t))   \mathbf{D}_{\mathbf{u}}\mathbf{x}_0(\mathbf u^s(\epsilon t)) \dot{\mathbf{u}}  $, this is consistent with existing choices (see \citet{S:breunung2018}, \citet{S:jain2022} and) where the higher-reduced dynamics term appears as the projection of the purely time-dependent term in eq.~(\ref{eq:higher-aSSM}) onto $\mathbf{E}(\mathbf u^s(\epsilon t))$. 

We further assume that control term $\mathbf{D}_{\mathbf{u}}\mathbf{x}_0(\mathbf u^s(\epsilon t)) \dot{\mathbf{u}} $ is nearly tangential to the aSSM. We have reason to expect this to be true, as the aSSM captures the most dominant slowest directions near the critical manifold, and the term of interest is a measure of the deviation from the critical manifold, which should lie close to the aSSM. Hence, we can safely ignore the contribution of the higher-order term to the aSSM parameterization. Accounting for these approximations, the $d$-dimensional aSSM-reduced model takes the final form, 
\begin{align}
\label{eq:aSSM_split}
\mathbf{r} &= \mathbf{V}^\mathrm{T}(\mathbf u^s(\epsilon t))(\mathbf{x}-\mathbf{x}_0(\mathbf{u}(t))), \nonumber \\
\mathbf{\dot{r}} &= \mathbf{R}(\mathbf{r}, \mathbf{x}_0(\mathbf{u}^s(\epsilon t))) - \mathbf{V}^\mathrm{T}(\mathbf u^s(\epsilon t))   \mathbf{D}_{\mathbf{u}}\mathbf{x}_0(\mathbf u^s(\epsilon t)) \dot{\mathbf{u}} + O\left((\epsilon+\delta)^2,\epsilon|\mathbf{r}|,\delta |\mathbf{r}|\right), \\
\mathbf{x} &=   \mathbf{W}(\mathbf{r}, \mathbf{x}_0(\mathbf{u}^s(\epsilon t)))+ \mathbf x_0(\mathbf{u}^s(\epsilon t))  + O\left((\epsilon+\delta),\epsilon|\mathbf{r}|,\delta |\mathbf{r}|\right). \nonumber 
\end{align}

\begin{remark}
\textbf{[aSSM-reduced model without control splitting]}
In trajectory tracking applications, the slowly varying control input $\mathbf{u}^s(\epsilon t)$ is typically dominant in magnitude over the faster deviation control forces $\mathbf{u}^d(t)$ at all times. Under this assumption, the critical manifold $\mathcal{L}_0$ and the chart mappings to the aSSM, evaluated along $\mathbf{u}^s(\epsilon t)$, remain consistent with those evaluated on the general control input $\mathbf{u}(t) = \mathbf{u}^s(\epsilon t) + \mathbf{u}^d(t)$. This justifies expressing the aSSM-reduced model (see eq.~(\ref{eq:aSSM_split})) with only the general control input $\mathbf{u}(t)$:
\begin{align}
\label{eq:aSSM_final}
\mathbf{r} &= \mathbf{V}^\mathrm{T}(\mathbf u(t))(\mathbf{x}-\mathbf{x}_0(\mathbf{u}(t))), \nonumber \\
\mathbf{\dot{r}} &= \mathbf{R}(\mathbf{r}, \mathbf{x}_0(\mathbf{u}(t))) - \mathbf{V}^\mathrm{T}(\mathbf u(t))   \mathbf{D}_{\mathbf{u}}\mathbf{x}_0(\mathbf u(t)) \dot{\mathbf{u}}, \\
\mathbf{x} &=   \mathbf{W}(\mathbf{r}, \mathbf{x}_0(\mathbf{u}(t)))+ \mathbf x_0(\mathbf{u}(t)). \nonumber 
\end{align}
\end{remark}

\begin{remark}
\textbf{[Related aSSM-reduced model derivations]}
\citet{S:haller24_wa} provides proofs for the existence of aSSMs and derives the aSSM-reduced model for the same setting as in eq. (\ref{eq:eom}) with $n_u =1$ and $\mathbf{u}(t)\equiv \mathbf{u}^s(\epsilon t)$. \citet{S:Kaundinya2026} presented this derivation for general control inputs assuming the survival of the aSSM defined by \citet{S:haller24_wa} under additional fast uniformly bounded deviation control inputs. 

Both their derivations differ in the choice of coordinate change $\boldsymbol \zeta$ in eq.~(\ref{eq:flatten_cm}). Translating the choice of  \citet{S:haller24_wa} to our setting results in $\boldsymbol \zeta(t) = \mathbf{x}(t)-\mathbf{x}_{\epsilon}(\mathbf{u}^s(\epsilon t))$, which measures the distance of the robot's response from the slow manifold $\mathcal{L}_{\epsilon}$. From the data, one does not readily have access to the slow manifold $\mathcal{L}_{\epsilon}$ but will have access to the critical manifold $\mathcal{L}_{0}$. This observation prompted \citet{S:Kaundinya2026} to choose $\boldsymbol \zeta(t) = \mathbf{x}(t)-\mathbf{x}_{0}(\mathbf{u}^s(\epsilon t))$ as the coordinate for their derivation. Our choice of derivation, eq.~(\ref{eq:flatten_cm}), takes this a step further by evaluating the critical manifold for general control inputs, as would be done in practice.
\end{remark}

\section{Details for the linear and RL baselines used in our comparisons}
\label{app:baselines}

\subsection{DMDc}
We identify a linear time-invariant model directly in the delay-embedded observable space $\mathbf{y}$, using the same 2 hours of static $\{\mathbf{Y}^i_{\mathrm{step}}, \mathbf{U}^i_{\mathrm{step}}\}$ and random smooth data $\{\mathbf{Y}_{\mathrm{calib}}, \mathbf{U}_{\mathrm{calib}}\}$  collected to learn the aSSM-reduced model. We denote these as snapshot matrices $\mathbf{Y}, \mathbf{U}$ and their one-step shifted snapshot matrix as $\overline{\mathbf{Y}}$, we solve
\begin{align}
\{\mathbf{A}^\star, \mathbf{B}^\star\} = \operatorname*{arg\,min}_{\mathbf{A}, \mathbf{B}} \big\| \overline{\mathbf{Y}} - \mathbf{A}\mathbf{Y} - \mathbf{B}\mathbf{U} \big\|^2
\end{align}
via least-squares regression (see \citet{S:Proctor2016}), and obtain the linear time-invariant model  
\begin{equation}
\label{eq:dmdc}
\mathbf{y}_{k+1} = \mathbf{A}^*\mathbf{y}_k + \mathbf{B}^*\mathbf{u}_k.
\end{equation}
We then plug eq.~(\ref{eq:dmdc}) as the dynamics constraint in the MPC formulation eq.~(\ref{eq:aSSM_mpc0}) to obtain a DMDc-based controller. Specifically, we tuned the controller to obtain decent closed-loop performance with cost matrices given by $\mathbf{R_{u}} = 0.001 \cdot \mathbb{I}_{6}$, $\mathbf{R_{\dot{u}}} = 0.001 \cdot \mathbb{I}_{6}$, and $\mathbf{Q_{z}} = \text{diag}(10,10,10)$. We used horizon lengths $M=10$ for target-tracking tasks and $M=20$ for obstacle-avoidance tasks. The model solve times were very fast, hence we could set the controller frequency to 100 Hz.

\subsection{Koopman-pregain}
 We fit a Koopman-pregain model (see \citet{S:haggerty23}) to the workspace coordinates $\mathbf{z}$. We collect only 30 minutes of random step-response data $\{\mathbf{Z}, \mathbf{U}\}$ for this baseline. We then delay-embed this once to obtain a delayed embedded dataset $\mathbf{Q}$.  We first fit a dynamic Koopman operator to $\mathbf{Q}$,
\begin{align}
\{\mathbf{A}^\star, \mathbf{B}^\star\} = \operatorname*{arg\,min}_{\mathbf{A}, \mathbf{B}} \big\| \overline{\mathbf{Q}} - \mathbf{A}\mathbf{Q} - \mathbf{B}\mathbf{U} \big\|^2.
\end{align}
This dynamic part alone underperformed, so \citet{S:haggerty23} added a static correction using the static equilibria $\{\mathbf{Z}^i_{\mathrm{static}}, \mathbf{U}^i_{\mathrm{static}}\}$ obatiend from the sampled dataset. We fit a linear map $\mathbf{G}$ from steady-state workspace positions to the static input that produced them,
\begin{align}
\mathbf{G}^\star = \operatorname*{arg\,min}_{\mathbf{G}} \sum_i \big\| \mathbf{U}^i_{\mathrm{static}} - \mathbf{G}\, \mathbf{Z}^i_{\mathrm{static}} \big\|^2.
\end{align}

They reformulate the optimization problem in eq.~(\ref{eq:aSSM_mpc0}) as a linear quadratic regulator (LQR) problem subject to a linear time-invariant dynamical system. They then approximate the finite-horizon problem by treating it as an infinite-horizon one, and obtain the optimal control inputs needed to track the target trajectory $\boldsymbol{\Gamma}(t)$
 by solving the algebraic Riccati equation for the pregain matrix $\mathbf{K}$
. This yields the control law
\begin{equation}
    \mathbf{u}^{*}(t) = -\mathbf{K} (\mathbf{z}-\boldsymbol{\Gamma}(t)).
\end{equation}

Because this version produced inaccurate tracking, they introduced a static feedforward correction term, resulting in
\begin{equation}
    \mathbf{u}^{*}(t) = -\mathbf{K} (\mathbf{z}-\boldsymbol{\Gamma}(t)) + \mathbf{G} \boldsymbol{\Gamma}(t).
\end{equation}

 \citet{S:haggerty23} observed accurate tracking with the added static term, and their results showed that the dynamic component contributed very little. This approach can reasonably be expected to work only when the target trajectory varies slowly, and the robot’s dynamics are purely linear. In practice, they rely on these assumptions by first driving the robot along slowly varying circular paths, then gradually increasing speed once it has latched onto the target. This method would fail if the Koopman-based LQR controller were required to track a rapidly accelerating target from the beginning.

\subsection{Mori-Zwanzig (MZ) Koopman approach}
The MZ-Koopman baseline by \citet{S:mori_soft_26} is trained on the same 2 hours of static and random control data used for learning the aSSMs and DMDc models. Closely following the procedure by \citet{S:mori_soft_26}, we lift the observable state $\mathbf{y}$ using Gaussian radial basis functions (RBF) and include a one-step memory term. This results in the following MZ-Koopman dynamic map, 
\begin{align}
\mathbf{q}_{k+1} = \mathbf{A}_0 \mathbf{q}_k + \mathbf{A}_1 \mathbf{q}_{k-1} + \mathbf{B}_0 \mathbf{u}_k + \mathbf{B}_1 \mathbf{u}_{k-1}, \qquad \mathbf{q} = \phi(\mathbf{y}).
\end{align}
The RBF map $\phi$ is built from $N_c = 33$ unique centers $\{\mathbf{c}_j\}_{j=1}^{33}$, spanning the observable space of trunk uniformly, and chosen from the training data. The map $\phi$ is given by 
\begin{align}
\phi_j(\mathbf{y}) = \exp\!\left(-\frac{\|\mathbf{y} - \mathbf{c}_j\|^2}{2\sigma^2}\right), \qquad j = 1, \dots, N_c.
\end{align}
We used the same choice for the number of centers as in \citet{S:mori_soft_26}, as a large number of centers led to larger linear dynamic models that slow down MPC. We also observed that the model's performance did not quantitatively improve with 33 centers vs. 50 centers. The standard deviation $\sigma$ is kept fixed as in \citet{S:mori_soft_26}. The authors do not provide a systematic method to compute this. Hence, we set the parameter $\sigma$ by sweeping over values and selecting the one that yielded the lowest prediction error on an unseen test dataset. Similar to DMDc, the set $\{\mathbf{A}_0, \mathbf{A}_1, \mathbf{B}_0, \mathbf{B}_1\}$ is solved by least-squares regression on the lifted snapshots $\mathbf{Q}$ and control inputs $\mathbf U$. 

For MPC deployment, we absorbed the memory term into an augmented state $\boldsymbol{\xi}_k = \{\mathbf{q}_k, \mathbf{q}_{k-1},\mathbf{u}_{k-1} \} $, and recovered a linear time invariant system that can readily be used with standard MPC schemes. We found that tuning $\mathbf{R_{u}} = 0.0001 \cdot \mathbb{I}_{6}$, $\mathbf{R_{\dot{u}}} = 0.001 \cdot \mathbb{I}_{6}$, and $\mathbf{Q_{z}} = \text{diag}(10,10,10)$ led to reasonable closed-loop performance. We used horizon lengths $M=10$ for target tracking tasks, and for obstacle avoidance tasks, sometimes $M=20$. The model solve times were very slow, hence we could only set the controller frequency to 20 Hz to achieve good tracking performance.

\subsection{Reinforcement Learning}

We train a model-free RL baseline using proximal policy optimization (PPO) as discussed by \citet{S:schulman2017proximalpolicyoptimizationalgorithms}. We adopt the neural network architecture from their work and train it to update its parameters directly on hardware. The policy is trained to minimize the same cost structure as in the aSSM-reduced MPC objective, combining target-tracking error with a control-rate penalty. Part of this cost structure is the reward,
\begin{align}
\text{Reward}(t) = -\Big(\|\mathbf{z}(t) - \boldsymbol{\Gamma}(t)\|^2 \Big),
\end{align}
used as the negative reward at each control step. In addition to penalizing the control rate, we ensured that the guessed control inputs were generated continuously to prevent abrupt movements that could cause the trunk robot to break. The target trajectory $\Gamma(t)$ is chosen in the trunk's workspace from randomly sampling smooth 10-second-long trajectories from the collected smooth random training data $\mathbf Y_{\text{calib}}$ and $\mathbf{U}_{\text{calib}}$ for the other models. We did this to avoid forcing policy to conform to physical targets that might be unfeasible to reach physically.

\begin{figure}[t!]
    \centering
   
        \includegraphics[width=0.7\linewidth]{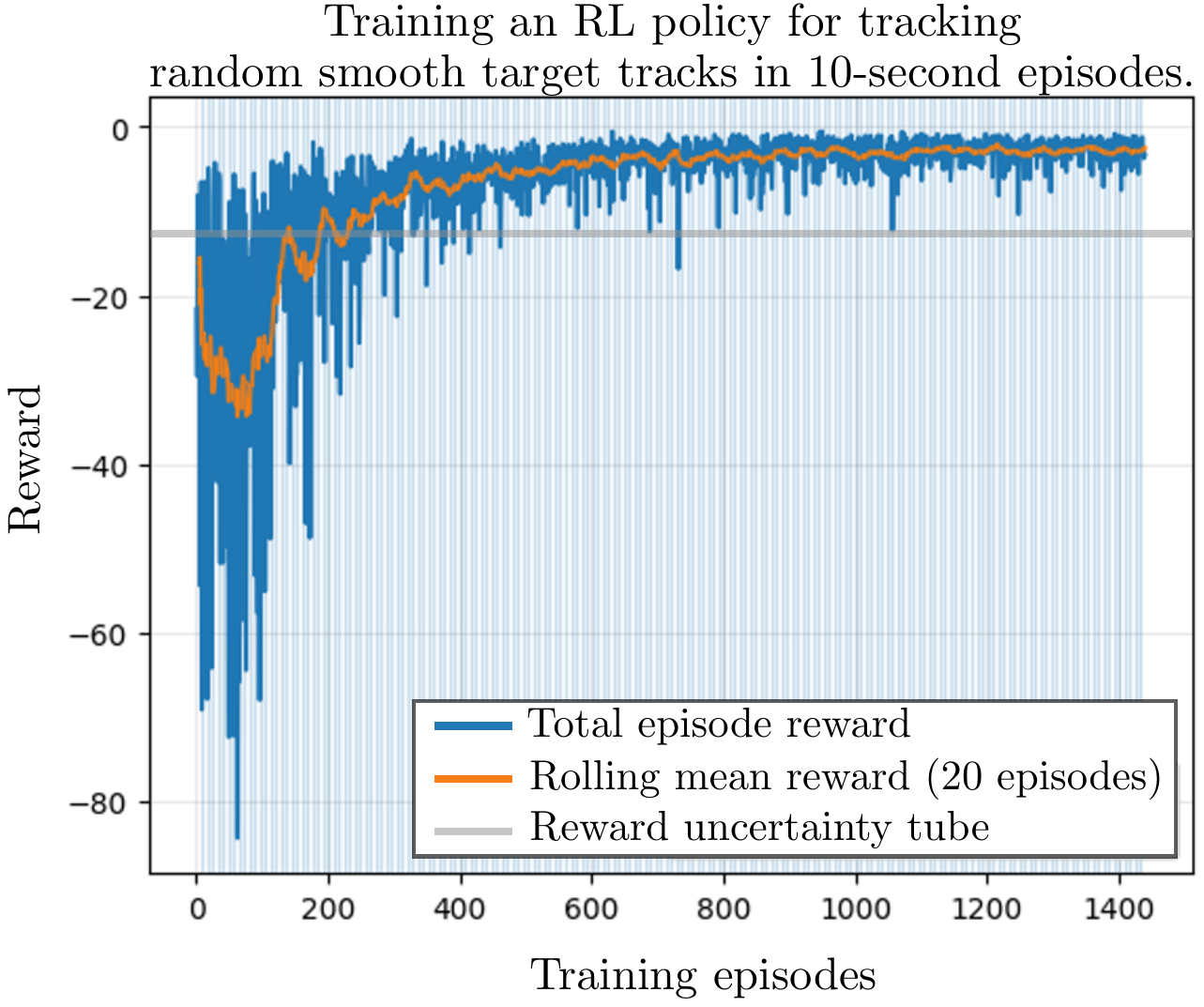}
   
    \caption{Training reward curve for proximal policy optimization (PPO) on hardware. The solid gray line indicates the allowed uncertainty of $-17.5$, computed by summing the maximum allowed positional uncertainty (3~cm tube radius, measured on hardware) across the 10-second episode.}
    \label{app:fig:RL_reward}

\end{figure}

In Fig.~(\ref{app:fig:RL_reward}), we show the resulting training reward curve. The solid gray line marks the maximum allowed uncertainty of $-17.5$ in the reward. We obtain it by summing the maximum allowed positional uncertainty due to the motors (the 3~cm uncertainty tube radius measured in hardware tests) over the 10-second episode. We observe that the RL policy converges below this uncertainty threshold after 2 hours of training, and further reduction in jittery behavior was observed after training for 2 hours more. We selected the best-performing converged policy during this period for closed-loop deployment. The execution of the RL policy was extremely fast, and we operated it at 100 Hz.

\renewcommand{\refname}{Supplementary References}


\end{document}